%% file: main.tex
\documentclass[11pt, a4paper, logo, copyright]{googlecloud}

\pdftrailerid{redacted}

\makeatletter
\renewcommand\bibentry[1]{\nocite{#1}{\frenchspacing\@nameuse{BR@r@#1\@extra@b@citeb}}}
\makeatother

\usepackage{kantlipsum, lipsum}
\usepackage{dsfont}

\usepackage[authoryear, sort&compress, round]{natbib}

\input{math_commands.tex}

\usepackage[utf8]{inputenc} 
\usepackage[T1]{fontenc}    
\usepackage{url}            
\usepackage{booktabs}       
\usepackage{nicefrac}       
\usepackage{microtype}      
\usepackage{amsmath}
\usepackage{graphicx}
\usepackage{multicol}
\usepackage{hyperref}       
\usepackage[nameinlink]{cleveref}
\usepackage{bbm}
\usepackage{multirow}
\usepackage{soul}
\usepackage{floatrow}
\usepackage{float}
\usepackage{blindtext}
\usepackage{tablefootnote}
\usepackage{amsfonts}
\usepackage[flushleft]{threeparttable}
\usepackage{bbding}
\usepackage{xcolor}
\usepackage{xspace}
\usepackage{bm}
\usepackage{arydshln}
\usepackage{subfigure}
\usepackage{enumitem}
\usepackage{setspace}
\usepackage{color}
\usepackage{algorithm}
\usepackage{algpseudocode}
\usepackage{longtable}

\usepackage[normalem]{ulem}
\usepackage{ulem}
\usepackage[nomargin,inline,marginclue,draft]{fixme}
\usepackage{balance}
\usepackage{verbatim}
\usepackage{diagbox}
\usepackage{changepage}
\usepackage{amssymb}
\usepackage{pifont}
\usepackage{array}   
\usepackage{wrapfig}

\newcommand{\ours}{\textsc{Learn-by-interact}\xspace}
\definecolor{rebuttal}{RGB}{0,0,0}

\title{Learn-by-interact: A Data-Centric Framework for Self-Adaptive Agents in Realistic Environments}

\correspondingauthor{hjsu@cs.hku.hk}

\renewcommand{\today}{}

\author[1 2 *]{Hongjin Su}
\author[1]{Ruoxi Sun}
\author[1]{Jinsung Yoon}
\author[1]{Pengcheng Yin}
\author[2]{Tao Yu}
\author[1]{Sercan Ö. Arık\hspace{-0.4ex}}

\affil[1]{Google}
\affil[2]{The University of Hong Kong}

\input{texts/abstract}

\begin{document}

\maketitle

\input{texts/introduction}

\input{texts/learn_by_interact}

\input{texts/experiments}

\input{texts/analysis}

\input{texts/related_work}
\input{texts/conclusion}

\input{texts/limitations}


\bibliographystyle{abbrvnat}
\bibliography{bibtex}

\clearpage

\appendix
\input{texts/appendix}

\end{document}

%% file: math_commands.tex
\usepackage{amsmath,amsfonts,bm}

\def\eqref#1{equation~\ref{#1}}

\def\1{\bm{1}}

\DeclareMathAlphabet{\mathsfit}{\encodingdefault}{\sfdefault}{m}{sl}
\SetMathAlphabet{\mathsfit}{bold}{\encodingdefault}{\sfdefault}{bx}{n}



%% file: texts/abstract.tex
\begin{abstract}
Autonomous agents powered by large language models (LLMs) have the potential to enhance human capabilities, assisting with digital tasks from sending emails to performing data analysis. The abilities of existing LLMs at such tasks are often hindered by the lack of high-quality agent data from the corresponding environments they interact with. 
We propose \ours, a data-centric framework to adapt LLM agents to any given environments without human annotations. 
\ours synthesizes trajectories of agent-environment interactions based on documentations, and constructs instructions by summarizing or abstracting the interaction histories, a process called \textit{backward construction}.  
We assess the quality of our synthetic data by using them in both training-based scenarios and training-free in-context learning (ICL), where we craft innovative retrieval approaches optimized for agents.
Extensive experiments on SWE-bench, WebArena, OSWorld and Spider2-V spanning across realistic coding, web, and desktop environments show the effectiveness of \ours in various downstream agentic tasks --- baseline results are improved by up to 12.2\% for ICL with Claude-3.5 and 19.5\% for training with Codestral-22B.
We further demonstrate the critical role of backward construction, which provides up to 14.0\% improvement for training.
Our ablation studies demonstrate the efficiency provided by our synthesized data in ICL and the superiority of our retrieval pipeline over alternative approaches like conventional retrieval-augmented generation (RAG).
We expect that \ours will serve as a foundation for agent data synthesis as LLMs are increasingly deployed at real-world environments. 
\end{abstract}

%% file: texts/introduction.tex
\section{Introduction}

\begin{figure}[t!]
\centering
    \includegraphics[width=\textwidth]{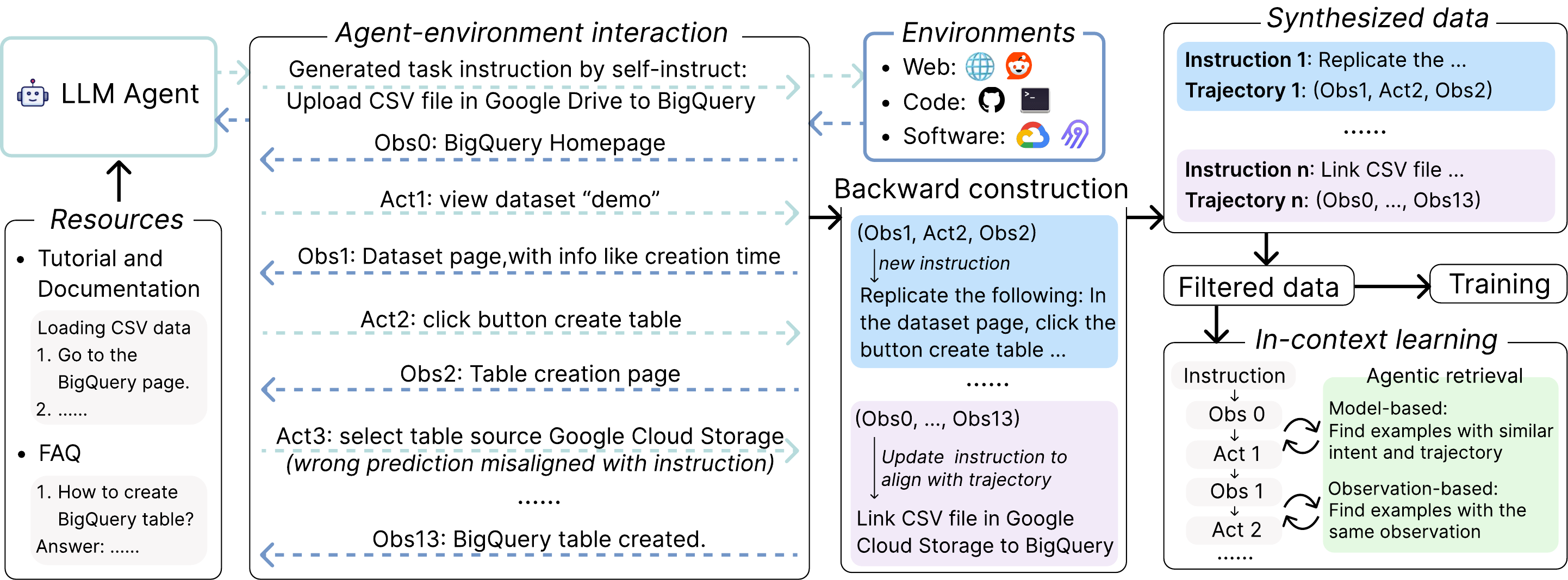}
\caption{
Overview of the data synthesis and adaptation processes.
Given an environment and standard resources, we first leverage self-instruct to create a diverse set of instructions. 
LLMs are then employed to complete these tasks, resulting in long trajectories of agent-environment interactions.
We construct task instructions using LLMs for each sub-trajectory, a process called \textit{backward construction}.
The synthesized data are then filtered and used for both training and in-context learning, where we design agentic retrieval to retrieve demonstration examples based on information at each step, using both model-based and observation-based approaches.
See Appendix~\ref{app:data_synthesis_example} for the complete data synthesis example and Algorithm~\ref{alg:adaptation} for more details on agentic retrieval.
}
\label{fig:figure1}
\end{figure}

Pre-trained large language models (LLMs) offer great potential for assisting humans with various tasks in digital settings, such as editing images, performing data analysis, resolving software engineering issues, and navigating commercial platforms ~\citep{xie2023openagents,xie2024osworld,yao2022webshop,jimenez2023swe}.
By streamlining these, LLM agents can greatly enhance human efficiency and productivity, allowing users to shift their focus toward higher-level, creative, and strategic endeavors.
To explore this potential, many benchmarks~\citep{jimenez2023swe,zhou2023webarena,xie2024osworld,cao2024spider2,koh2024visualwebarena} and agentic frameworks~\citep{yang2024swe,zhan2023you,yang2023appagent,gur2023real,chen2024coder} 
have been established based on realistic digital environments, spanning web applications, code development, \textcolor{rebuttal}{desktop computing, etc.}
However, LLMs often fall short of expected performance in these tasks, consistently displaying a significant gap compared to human capabilities. As a result, they remain less practical and reliable for real-world applications.

Efficient adaptation to new environments can be a key part of the performance improvements.
Prior works have explored various prompt-based approaches~\citep{yao2022react,yang2024swe,gur2023real,zhan2023you}, that are constrained by the capabilities of underlying foundation models.
Other studies on training LLMs with human-labeled examples~\citep{chen2023fireact,chen2024agent,li2020mapping} on the other hand,
come with the fundamental limitation of high annotation costs when new environments are considered.
In particular, annotating agentic data can be quite difficult and expensive due to long-trajectory interactions with environments and specific domain expertise required.
Few works have explored fully-autonomous data construction pipelines towards self-adaptive agents that can efficiently learn new environments~\citep{gulcehre2023reinforced,aksitov2023rest}.

In this paper, we introduce \ours, a data-centric framework for LLMs to self-adapt to new environments, utilizing agent data synthesis via interactions.
Intuitively, the effects of actions executed 
in environments (e.g., the next webpage after clicking a button) serve as informative demonstrations that help LLMs in future navigation.
Inspired by this, we design \ours that first uses self-instruct~\citep{wang2022self} to develop a variety of task instructions, referring to standard resources such as documentations and tutorials for a given environment. 
\textcolor{rebuttal}{This covers most important scenarios that human users are interested in and avoids intensive prompt engineering to control the distribution and diversity of the generated data.}
We then collect diverse trajectories from interactions between LLMs and environments, as illustrated in Fig.~\ref{fig:figure1}.
However, given the low performance of LLMs in existing agentic benchmarks~\citep{xie2024osworld,cao2024spider2}, it is likely that a large percentage of synthesized trajectories would not match with the instructions.
To tackle this challenge, we construct new instructions by summarizing or abstracting each sub-trajectory, leveraging the strong summarization capabilities of LLMs~\citep{pu2023summarization,liu2023learning}.
We call this process \textit{backward construction}.
After obtaining synthesized instruction-trajectory pairs and filtering low-quality ones, we apply it to both training and ICL, where we craft innovative retrieval pipelines optimized for agents.
Specifically, the approach comprises two components: (1) a model-based approach where LLMs generate queries guided by instructions, interaction histories, and current observations, followed by retrieval models selecting demonstration examples from synthesized data; and (2) an observation-based approach that identifies examples in which the current observation appears in trajectories, signaling that the current state was encountered during the data synthesis process.

Our comprehensive evaluations across four challenging benchmarks: SWE-bench~\citep{jimenez2023swe}, 
WebArena~\citep{zhou2023webarena}, 
OSWorld~\citep{xie2024osworld}, 
and Spider2-V~\citep{cao2024spider2}, 
highlight the efficacy of the data generated by \ours. 
With ICL, both Gemini-1.5-pro~\citep{reid2024gemini} and Claude-3.5-sonnet~\citep{claude35anthropic} show consistent and remarkable improvements -- for OSWorld~\citep{xie2024osworld}, our generated data nearly doubles Claude-3.5-sonnet's baseline performance, increasing it from 12.4\% to 22.5\%. 
This enables \ours to achieve the best-class performance in all of the four leaderboards.
Furthermore, substantial improvements are observed by training models of varying sizes and architectures with our synthesized data. 
As an example, Codestral-22B's~\citep{codestralblog} performance on WebArena significantly increases from 4.7\% to 24.2\% after training. 
These results underscore the high quality of the generated agentic data and the broad applicability across diverse environments.

Our extensive ablation studies reveal that backward construction not only increases the quantity of the synthesized data, but also improves its overall quality (\S\ref{sec:results}). 
With data synthesized by \ours, we observe significant improvements in both performance and efficiency during LLM inference (\S\ref{sec:efficiency}).
Our empirical results demonstrate the superiority of the agentic retrieval in ICL (\S\ref{subsec:retrieval}).
We anticipate that this research will spark innovative developments in enhancing agentic performance using LLMs and contribute to its wider-spread adoption in real-world application scenarios.

%% file: texts/learn_by_interact.tex
\section{Learn-by-interact}

We introduce the proposed \ours framework to synthesize agent data in an autonomous way by leveraging interactions between LLMs and environments.
We first formalize the canonical agentic tasks (\S\ref{sec:task_formulation}), and introduce the detailed synthesis (\S\ref{sec:data_synthesis}) and filtering (\S\ref{sec:filtering}) procedures.
We then describe the application of the synthesized data in adapting LLMs in both training-free and training-based settings (\S\ref{sec:adaptation}).

\subsection{Task formulation}
\label{sec:task_formulation}
Given an environment $E$ and a task instruction $I$, the objective of an agent $A$ is to achieve the target $G$ through multi-step interactions with $E$.
At each step $i$, $A$ predicts the next action $a_{i}$ based on the instruction $I$ and the previous history $H=(o_0, a_1, o_1, a_2, ..., o_{i-1})$, which is then executed in the environment $E$ to get a new observation $o_i$.
The interactions terminated until $A$ predicts the action $stop$ or the maximum number of steps $m$ is reached.

\subsection{Agentic data synthesis}
\label{sec:data_synthesis}

The essential idea of \ours is manifested in synthesizing environment-specific agent data with zero human effort. 
In Algorithm~\ref{alg:data_synthesis}, we show the overall process with pseudo-code.
Given an environment for a downstream application (e.g., Visual Studio Code), we first leverage commonly-accessible resources such as documentation to generate diverse task instructions using self-instruct~\citep{wang2022self} (line 5).
These resources are usually created by human experts to address common concerns and provide usage suggestions, e.g., how to navigate a website or operate a software.
Intuitively, such references often cover representative use cases of an application. 
Therefore, the task instructions generated conditioned on them could cover most popular scenarios in the domain and avoid potentially unreasonable cases that may be of less value.

\input{texts/algorithms/data_synthesis}

For each generated task, LLMs then aim to solve it, which results in a long trajectory $T=(o_0, a_1, o_1, ..., a_n, o_n)$ (line 9-14 in Algorithm~\ref{alg:data_synthesis}). 
To address the potential misalignment between the instruction $I$ and the generated trajectories $T$, we introduce a novel mechanism, backward construction, to construct instructions based on trajectories (lines 15-22 in Algorithm~\ref{alg:data_synthesis}).
Specifically, for each sub-trajectory $T'=(o_i, a_{i+1}, o_{i+1}, ..., a_j, o_j), 0 \leq i<j \leq n$, we obtain two types of new instructions: (1) summaries of trajectory steps; and (2) abstractions of the trajectory purpose. 
In Fig. \ref{fig:figure1}, the sub-trajectory $(Obs 1, Act 2, Obs 2)$ is summarized into a new task instruction that requires to replicate the $Act 2$.
The abstraction of the full trajectory updates the original task objective, which is no longer aligned with the generated trajectory due to the wrong prediction in the action 3.
Overall, the \ours pipeline offers two notable advantages:
(1). It corrects the potential misalignment between instructions and predicted trajectories by updating task objectives, which enhances the data quality as verified by the experimental results in \S\ref{sec:results}.
(2). It maximizes the utility of each generated trajectory by crafting new instructions for each sub-trajectory. This results in a quadratic increase in the number of synthesized examples with respect to the steps in the sequence per generated trajectory. For a given target dataset size, backward construction substantially decreases the necessary interactions, which is particularly valuable in scenarios where such interactions are challenging and costly to obtain such as Robotics~\citep{keipour2022physical}.


\subsection{Filtering}
\label{sec:filtering}
To further enhance the data quality, we design the following criteria to filter inferior synthesized data: 
(1). Remove duplicate states: We remove duplicate $(a_i,o_i)$ from $T'$ if $(a_i,o_i)$=$(a_{i-1},o_{i-1})$, which is potentially introduced by the invalid action or the environment error (inactivity).
(2). LLM committee check: We feed the generated instruction-trajectory pair ($I', T')$ into a committee of LLMs, and only classify it of high-quality if all LLMs consider the trajectory coherent, natural, reasonable and aligned with the instruction.
The listed criteria are all fully-autonomous and canonically-applicable for filtering data synthesized in general agent scenarios.
Additionally, we employ iterative prompting to augment LLMs with high-quality examples to enhance their capabilities in data generation.
See Table~\ref{tab:filter} for our prompts used in LLM committee check.

\input{texts/algorithms/adaptation}

\newpage
\subsection{Adaptation}
\label{sec:adaptation}
After obtaining the synthesized data $D$, we apply it to both ICL and training.
Given the unique characteristics of multi-round interactions with environments in agent settings, we design agentic retrieval (pseudo-code in Algorithm~\ref{alg:adaptation}) to maximize the effectiveness of the synthesized data.
Specifically, we propose two retrieval pipelines: observation-based (line 5-14) and model-based retrieval (line 15-17).
In observation-based retrieval, we compare the current observation $o$ to the trajectory of each example $e$ in the synthesized data, where $e=[I',[o_0, a_1, o_1, ..., a_n, o_n]]$.
If $o$ matches one of the observations in $e$, i.e., $o=o_i$, then we consider $e$ as a helpful example to the current task.
For the model-based retrieval, we leverage LLMs to first write queries based on the instruction, the interaction history and the current observation (line 16), and then employ retrieval models to retrieve non-duplicate examples (line 17).
LLMs are then augmented with the retrieved examples to predict the next action (line 18).
Refer to Table~\ref{tab:inference_prompt_1} to \ref{tab:write_query} for prompts to write queries and predict actions.

Apart from using the synthesized data as demonstration examples in ICL, we further utilize them to fine-tune models.
For a given generated example, we convert it to the format of action prediction (Table~\ref{tab:inference_prompt_1}), and prepare input-output pairs for supervised fine-tuning.
More details on the experimental settings can be found in \S\ref{sec:settings}.

%% file: texts/algorithms/data_synthesis.tex
\begin{wrapfigure}{r}{0.5\textwidth}
\begin{minipage}{0.99\textwidth}
\vspace{-20pt}
\begin{algorithm}[H]
\fontsize{9pt}{10pt}\selectfont
\caption{Agent data synthesis}
\label{alg:data_synthesis}
\begin{algorithmic}[1]
\State \textbf{Input:} $LLM$: Large Language Model; $E$: environment; $Doc$: standard resources like documentation; $N$: the number of instructions to generate per document; $F$: data filter.
\State \textbf{Initialization:} $D=[]$: synthesized data.
\For{$d$ in $Doc$} 
    \State \textcolor{black}{// self-instruct to generate $N$ task instructions}
    \State $Instructions = LLM(d,N)$ 
    \For{$I$ in $Instructions$} 
        \State $E$.reset()
        \State $T$ = [] \quad \textcolor{rebuttal}{// initialize interaction trajectory}
        \While{not $E$.finished()}
            \State $o = E$.get\_observation()
            \State $a = \textcolor{rebuttal}{LLM}(I,T,o)$
            \State $T$ += $[o,a]$
        \EndWhile
        \State $T.append(E$.get\_observation())
        \State \textcolor{black}{// backward construction}
        \For{$i$ in range($0,len(T)-1,2$)} 
            \For{$j$ in range($i + 2,len(T),2$)} 
                \State $T' = T[i:j]$
                \State $I' = LLM(T')$ 
                \State $D.append([I',T'])$
            \EndFor
        \EndFor
    \EndFor
\EndFor
\State $D$ = $F(D)$  \textcolor{black}{// Filter low-quality data}
\State \textbf{Return:} $D$
\end{algorithmic}

\end{algorithm}
\vspace{-20pt}
\end{minipage}
\end{wrapfigure}

%% file: texts/algorithms/adaptation.tex
\begin{wrapfigure}{r}{0.5\textwidth}
\begin{minipage}{0.99\textwidth}
\vspace{-20pt}
  \begin{algorithm}[H]
  \fontsize{9pt}{10pt}\selectfont
    \caption{ICL with agentic retrieval}
    \label{alg:adaptation}
    \begin{algorithmic} [1]
      \State \textbf{Input:} $LLM$: Large \textcolor{rebuttal}{Language Model}; $E$: environment; $D$: synthesized data; $BM25$: BM25 retrieval model; $RM$: dense retriever; $I$: task instruction; $m1$: maximum number of examples from observation-based retrieval; $m2$: maximum number of examples from model-based retrieval.
      \State \textbf{Initialization}: $H=[]$: interaction history; $R$: retrieved examples.
      \While{not $E$.finished()}
        \State $o = E$.get\_observation()
        \State \textcolor{black}{// observation-based retrieval}
        \State $R = BM25(o,D,m1)$
        \State \textcolor{black}{// model-based retrieval}
        \State $q=LLM(I,H,o)$ 
        \State $R$ += $RM(q,D,m2,R)$
        \State $a=LLM(I,H,o,R)$
        \State $H += [o,a]$
      \EndWhile
    \end{algorithmic}
  \end{algorithm}
  \vspace{-25pt}
\end{minipage}
\end{wrapfigure}

%% file: texts/experiments.tex
\section{Experiments}

\subsection{Baselines}
\label{sec:baselines}
We compare ICL with agentic retrieval to the following prompt-based approaches.
\begin{itemize}[leftmargin=*]
    \item Baseline: The vanilla prediction pipeline in each benchmark that includes the task instruction, interaction history and the state observation in the prompt. See more implementation details in Appendix~\ref{app:baseline_implementation}.
    \item RAG: The conventional RAG pipeline that first retrieves from the resources like documentation based on the instruction, and augments LLMs with the retrieved content. 
    \item Data distill: We follow the same pipeline to synthesize data in Algorithm~\ref{alg:data_synthesis} except backward construction (replace lines 15-22 with $D.append(I,T)$), and follow Algorithm~\ref{alg:adaptation} during the evaluation.
    \item Reflexion~\citep{shinn2024reflexion}: A general framework to reinforce language agents through linguistic feedback from both executors and LLMs.
    \item Language Agent Tree Search (LATS)~\citep{zhou2023language}: It integrates the combinatorial tree search into expanding ReAct~\citep{yao2022react} and combine agent online reasoning, acting and planning throughout the trajectory.
\end{itemize}
For the training-based evaluation, we primarily compare to the data distillation, which also constructs data from scratch and requires no human effort to annotate seed or preference data.
Additionally, we include the model performance before training as another baseline.

\subsection{Datasets}
\label{sec:datasets}
We consider the four agentic datasets that involve multi-round interactions with realistic environments.
They span diverse domains of code, web, computer desktop and professional software.
Appendix~\ref{app:data_examples} illustrates details of each dataset with examples.
\begin{itemize}[leftmargin=*]
    \item SWE-bench~\citep{jimenez2023swe} is an evaluation benchmark on realistic software engineering problems from realistic Github issues. We use the verified version by default throughout the experiments.
    \item Webarena~\citep{zhou2023webarena} evaluates agent capabilities to perform tasks in the web environments such as e-commerce, social forum discussion, and beyond.
    \item OSWorld~\citep{xie2024osworld} is an integrated environment for assessing open-ended computer tasks, which involve diverse applications like Terminal, Chrome, etc.
    \item Spider2-V~\citep{cao2024spider2} is a multimodal agent benchmark focusing on professional data science and engineering workflows, which includes BigQuery, Airbyte and more.
\end{itemize}

\input{tables/data_statistics}

\subsection{Settings}
\label{sec:settings}
We synthesize one separate set of environment-specific data for each evaluated benchmark.
Throughout the data synthesis process, we employ the Claude-3.5-sonnet~\citep{claude35anthropic} as the generator model and both Gemini-1.5-pro~\citep{reid2024gemini} and Claude-3.5-sonnet as the LLM committee for filtering low-quality data.
For each document, we sample three task instructions from LLMs.
The statistics for generated raw trajectories, examples before and after filtering are shown in Table~\ref{tab:data_statistics}.
In Appendix~\ref{app:document_sources}, we list document sources used for each benchmark.
During ICL, we retrieve examples until the maximum length of LLMs and set an upper bound of 5 for both model-based and observation-based retrieval ($m1=5$, $m2=5$ in Algorithm~\ref{alg:adaptation}).
We leverage Gemini-1.5-pro~\citep{reid2024gemini} and Claude-3.5-sonnet~\citep{claude35anthropic}\footnote{In the subsequent descriptions, Gemini refers to Gemini-1.5-pro, and Claude refers to Claude-3.5-sonnet.}, Codegemma-7B~\citep{team2024codegemma} and Codestral-22B~\citep{codestralblog} in the ICL evaluation, and tune Codegemma-7B and  Codestral-22B with LoRA~\citep{hu2021lora} to evaluate the data quality as training sources.
By default, we do not include retrieval content in evaluating the trained model to avoid the confusion in understanding the effectiveness of our synthesized data in training.
We include more detailed hyper-parameter settings (both existing approaches and \ours) and machine information in Appendix~\ref{app:experiment_settings}.

\subsection{Evaluation}
We follow the default evaluation metrics designed by the original benchmarks.
On SWE-bench~\citep{jimenez2023swe}, we apply the generated patch program to the repository codebase, and measure the agent performance by execution accuracy (pass@1).
On WebArena~\citep{zhou2023webarena}, we employ both LLM-based fuzzy match and string match that checks keywords in predictions. Slightly different from the original work that uses gpt-4-0613 as the LLM judge, we use Claude-3.5-sonnet as a similar replacement.
On OSWorld~\citep{xie2024osworld}, we leverage the sample-specific evaluation scripts to assess the functional correctness of the task completion, which processes environment states and checks if agents finish the task as expected.
On Spider2-V~\citep{cao2024spider2}, we utilize file-based comparison, information-based validation, execution-based verification to determine whether a task is successfully completed.
\textcolor{rebuttal}{All performance numbers throughout the paper are shown in the percentage of resolved instances with \% omitted for brevity.}

\subsection{Results}
\label{sec:results}
\input{tables/inference_results}

\subsubsection{Training-free Evaluation}
We first consider \ours in the training-free setting, where the proposed methods can be applied to the commercial LLMs even with prediction-only API access.

Results on Table~\ref{tab:inference_results} show marginal improvement of RAG compared to the baseline, which suggests limited effectiveness by simply concatenating standard resources to LLM prompts.
By retrieving examples from distilled data, we observe better performance compared to RAG, but still no more than 2\% improvement over the baseline, which indicates that the distilled data tend to be noisy in the setting with multi-round agent-environment interactions.
This highlights the critical role of backward construction, which corrects the misalignment between instructions and trajectories by curating new task objectives.


Both Reflexion and LATS consistently improve over the baseline across 4 benchmarks, which demonstrate their general applicability to agent tasks.
Using the data synthesized from the \ours, we can see a significant performance gain compared to all other frameworks in both Gemini and Claude.
For example, on OSWorld, augmenting Claude with synthesized environment-specific data almost doubles the result compared to the baseline.
This signifies the high quality of the generated data and the effectiveness of the \ours framework.

\subsubsection{Training-based Evaluation}
\input{tables/train_results}
We consider the data synthesized by \ours in the scenario of LLM tuning, which is applicable to the LLMs with access to weight updates.

The results presented in Table~\ref{tab:train_results} reveal that \ours substantially surpasses both the baseline and data distillation, suggesting its capacity to generate high-quality training data that enables language models to learn and adapt efficiently. 
We discover that utilizing our synthesized data for model training yields better results compared to using it as in-context learning (ICL) examples. 
A notable instance is in WebArena, where Codestral-22B's performance jumps from 4.7\% to 24.2\% when trained on our synthesized data, while only showing a 5.5\% improvement in the ICL scenario. 
Remarkably, the Codestral-22B model trained with our synthesized data even outperforms Gemini when the latter uses our data as demonstration examples.


%% file: tables/data_statistics.tex
\begin{table}
\caption{Statistics for the number of crawled documents, generated raw trajectories, examples (instruction-trajectory pairs) and examples after filtering.
}
\centering
\begin{tabular}{ccccc}
\toprule
& SWE-bench & WebArena & OSWorld & Spider2-V \\
\midrule
Documents & 6,464 & 3,578 & 7,362 & 11,231 \\
Raw trajectories & 4,568 & 3,967 & 1,125 & 1,226 \\
Examples & 41,237 & 32,319 & 19,688 & 21,525 \\
Filtered examples & 10,232 & 10,456 & 11,782 & 10,169\\
\bottomrule
\end{tabular}

\label{tab:data_statistics}
\end{table}

%% file: tables/inference_results.tex
\begin{table}[t!]
\caption{Comparison of \ours to other existing training-free approaches. SWE refers to SWE-bench, Web refers to WebArena and OS refers to OSWorld. The best results are highlighted in bold. 
}
\centering
\resizebox{\textwidth}{!}{
\begin{tabular}{l|cccc|cccc}
\toprule
Benchmark $\rightarrow$ & SWE & Web & OS & Spider2-V & SWE & Web & OS & Spider2-V\\
\midrule
Approach $\downarrow$ & \multicolumn{4}{c|}{Gemini-1.5-pro} & \multicolumn{4}{c}{Claude-3.5-sonnet} \\
\midrule
& \multicolumn{8}{c}{\textit{Existing approaches}} \\
\midrule
Baseline & 13.3 & 17.9 & 4.9 & 8.3 & 51.2 & 35.8 & 12.4 & 8.4 \\
RAG & 13.7 & 19.5 & 5.1 & 9.1 & 51.8 & 36.9 & 12.8 & 9.2 \\
Data distill & 14.0 & 19.8 & 5.7 & 9.1 & 54.0 & 39.2 & 12.9 & 9.7 \\
Reflexion & 14.3 & 20.2 & 5.7 & 9.3 & 54.4 & 40.4 & 15.6 & 10.5 \\
LATS & 15.3 & 21.0 & 6.5 & 11.3 & 55.2 & 41.3 & 16.8 & 11.2 \\
\midrule
& \multicolumn{8}{c}{\textit{Ours}} \\
\midrule
Learn-by-interact & \textbf{18.7} & \textbf{25.6} & \textbf{10.3} & \textbf{16.4} & \textbf{60.0} & \textbf{48.0} & \textbf{22.5} & \textbf{16.6}\\
$\Delta$ over baseline & \textcolor{blue}{+5.4} &  \textcolor{blue}{+7.7} & \textcolor{blue}{+5.4} & \textcolor{blue}{+8.1} & \textcolor{blue}{+8.8} & \textcolor{blue}{+12.2} & \textcolor{blue}{+10.1} & \textcolor{blue}{+8.2}\\
\bottomrule
\end{tabular}

}
\label{tab:inference_results}
\end{table}

%% file: tables/train_results.tex
\begin{table}[t!]
\caption{Downstream task performance of models trained from data generated by Learning-by-interact and data distillation. We include the models results before training, where the synthesized data is used as demonstration examples, and after training, where the synthesized data is used to train models.
}
\centering
\resizebox{\textwidth}{!}{
\begin{tabular}{l|cc|cc|cc|cc}
\toprule
Benchmark $\rightarrow$ & Web & OS & Web & OS & Web & OS & Web & OS \\
\midrule
Model $\rightarrow$ & \multicolumn{2}{c|}{Codegemma-7B} & \multicolumn{2}{c|}{Codestral-22B} & \multicolumn{2}{c|}{Codegemma-7B} & \multicolumn{2}{c}{Codestral-22B} \\
\midrule
Approach $\downarrow$ & \multicolumn{4}{c|}{\textit{\textbf{Before tuning}}} & \multicolumn{4}{c}{\textit{\textbf{After tuning}}} \\
\midrule
& \multicolumn{8}{c}{\textit{Existing approaches}} \\
\midrule
Baseline & 3.3 & 0.0 & 4.7 & 2.2 & - & - & - & - \\
Data distill & 4.2 & 0.0 & 5.8 & 2.7 & 6.2 & 1.4 & 10.2 & 5.4 \\
\midrule
& \multicolumn{8}{c}{\textit{Ours}} \\
\midrule
Learn-by-interact & 7.6 & 3.5 & 9.9 & 5.4 & 14.6 & 6.5 & 24.2 & 11.7\\
$\Delta$ over baseline & \textcolor{blue}{+4.3} &  \textcolor{blue}{+3.5} & \textcolor{blue}{+5.2} & \textcolor{blue}{+3.2} & \textcolor{blue}{+11.3} & \textcolor{blue}{+6.5} & \textcolor{blue}{+19.5} & \textcolor{blue}{+9.5}\\
\bottomrule
\end{tabular}
}
\label{tab:train_results}
\end{table}

%% file: texts/analysis.tex
\section{Analysis}


\subsection{Inference Efficiency}
\label{sec:efficiency}
\begin{figure}[t]
\centering
    \includegraphics[width=\textwidth]{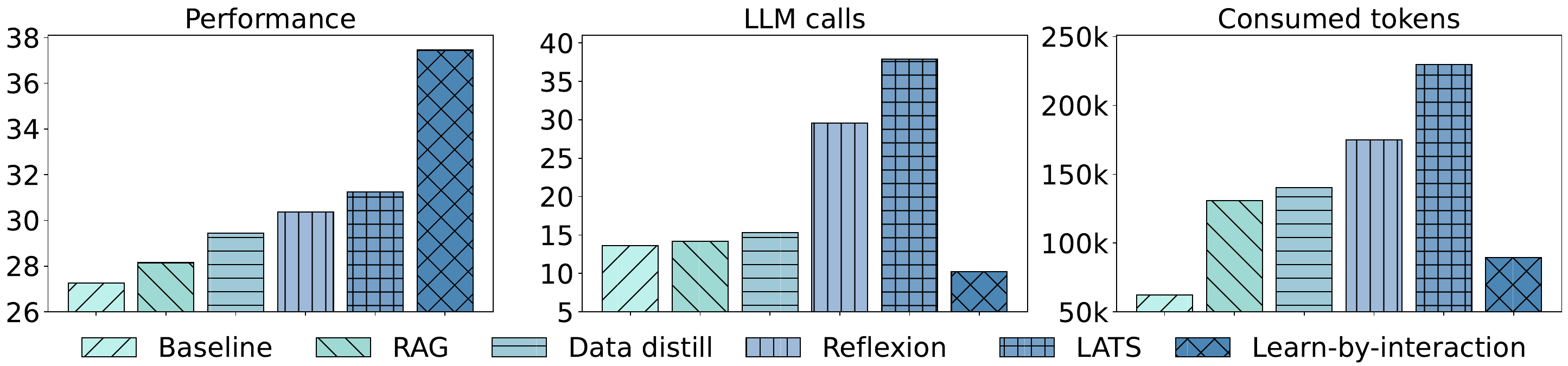}
\caption{Evaluation performance, the number of LLM calls and consumed tokens (per example) of various training-free pipelines \textcolor{rebuttal}{during inference}, which are averaged across four benchmarks: SWE-bench, Webarena, OSWorld and Spider2-V. 
}
\label{fig:efficiency}
\end{figure}

We compare the efficiency of different pipelines at inference.
We analyze the trade-off between downstream task performance and the required computational costs.
We focus on measuring the number of LLM calls and consumed tokens per example, which are averaged across four evaluated datasets (\S\ref{sec:datasets}) using Claude-3.5-sonnet.
As illustrated in Fig. \ref{fig:efficiency}, while Reflexion and LATS demonstrate enhanced performance, this comes at the cost of significantly increased computational resources during inference. 
Specifically, LATS achieves an average improvement of 2.5 \%, albeit at the cost of requiring nearly four times more tokens per instance compared to the baseline.
In contrast, \ours exhibits superior performance while utilizing fewer LLM calls and slightly more tokens compared to the baseline.
Thanks to the rich environment information stored in the examples of synthesized data, LLMs can potentially make better decisions and thus finish the task in fewer steps.
This removes the performance-efficiency trade-off during inference at the cost of data synthesis in advance and suggests that \ours is particularly well-suited for real-world deployment that demands both low latency and high performance.

\subsection{The Impact of Retrieval}
\label{subsec:retrieval}
As mentioned in \S\ref{sec:adaptation}, we employ both model-based and observation-based retrieval in our evaluation with ICL.
We analyze their effectiveness by incorporating only one of them (skip lines 5-14 in Algorithm~\ref{alg:adaptation} for model-based retrieval only and skip lines 15-17 for observation-based retrieval only).
In addition, we compare to two baselines: (1) no retrieval: LLMs predict each action in the zero-shot setting; and (2) instruction-based: only use instructions to retrieve synthesized data and apply the same demonstration examples in every action prediction throughout the trajectory.
\input{tables/retrieval}

The results presented in Table~\ref{tab:retrieval} illustrate how various retrieval methods impact LLMs when using the synthetic data as the retrieval source. 
Despite having access to the same example pool (except the baseline without using retrieval), there are notable differences in performance across different retrieval strategies, highlighting the crucial role of agentic retrieval in effectively utilizing synthesized data. 
Conventional Retrieval-Augmented Generation (RAG) methods, which only employs instructions for retrieval, show the least improvement across four benchmarks and two LLMs. 
In contrast, the observation-based approach proves particularly effective for agent-based tasks, significantly outperforming the instruction-based retrieval, for instance, achieving a 4.4\% absolute improvement on Spider-2V when using Gemini. 
By leveraging task instructions, interaction history and the current observation, model-based retrieval demonstrates even better results compared to using the observation-based version.
Ultimately, the most impressive scores are achieved by combining both model-based and observation-based retrieval, which results in our agentic retrieval pipeline.
These findings underscore the importance of carefully designing retrieval pipelines to maximize the potential of synthetic data and LLMs in agent scenarios.

\subsection{Data granularity}
\label{sec:data_granularity}
\input{tables/data_granularity}

As mentioned in \S\ref{sec:data_synthesis}, we synthesize data by taking contiguous sub-trajectories from the full generation paths of LLMs, i.e. $T'=T[i:j]$, which results in trajectories of diverse lengths in the synthesized data.
We divide the synthetic data into three groups: (1). trajectory steps $<5$ (short); (2). $5\leq$ trajectory steps $<10$ (medium); (3). trajectory steps $\geq10$ (long), and leverage each group and their combinations in both the training-free and the training-based process.
To ensure a fair comparison, we constraint the data size in each group  and combined group to 200M tokens\footnote{We use the number of tokens to measure the data size due to the fact that long-trajectory example may contain more information compared to the short version.}, utilizing \citet{su2022selective} for sub-sampling. 
Table~\ref{tab:data_granularity} presents the results.
In both training-free and training-based evaluation, LLMs derive greater advantages from short-trajectory data, as demonstrated by its consistently superior performance compared to medium and long-trajectory data with Claude-3.5-sonnet and Codestral-22B. 
This can be attributed to the versatility of short-trajectory data, which usually serves as a sub-step or a partial workflow in downstream tasks. 
The combination of any two data groups proves more effective than relying on a single group, showcasing the complementary nature of diverse data sets. 
For instance, in Webarena with Codestral-22B, incorporating examples with both short and medium-length trajectories shows additional improvement over using either one exclusively. 
This underscores the value of considering the trajectory length as a unique dimension of agentic data synthesis.

\begin{wrapfigure}{r}{0.5\textwidth}
  \begin{center}
    \includegraphics[width=0.99\textwidth]{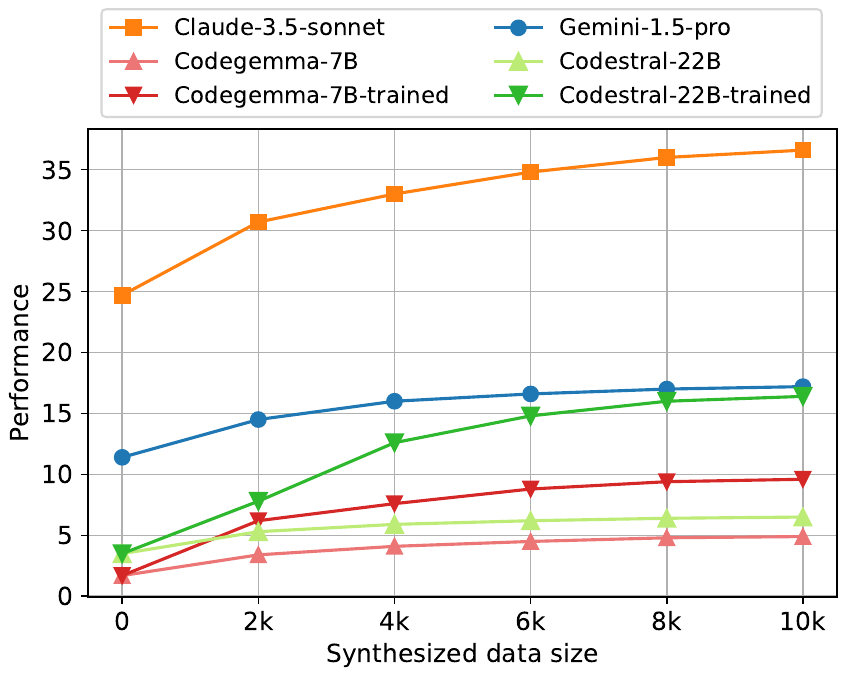}
  \end{center}
  \caption{Scaling laws for the synthesized data. Compared to in-context learning, tuning achieves more significant improvements as the data scales up. The performance is averaged across WebArena and OSWorld.}
  \label{fig:data_scale}
\end{wrapfigure}

\subsection{Scaling Laws}
\label{sec:scaling}
We examine how the model performance improves as the synthetic data size scales up. 
Figure~\ref{fig:data_scale} presents two sets of results, with training-free (where Claude, Gemini, Codegemma and Codestral use retrieval augmentation without training) and with training-based (where fine-tuned Codegemma and Codestral models are evaluated without retrieval). 
All results are averaged across Webarena and OSworld due to computational resource constraints.
The findings indicate that both learning paradigms benefit from larger data, suggesting the synthetic data is diverse and high-quality.
In the training-free evaluation, more substantial improvements are observed for larger models (Claude and Gemini) compared to smaller ones (Codegemma and Codestral), possibly due to the their enhanced in-context learning abilities. 
Our analysis also reveals that for a given amount of synthetic data, fine-tuning smaller models is more effective than using the data as demonstration examples during evaluation.



%% file: tables/retrieval.tex
\begin{table}[t!]
\caption{Model performance based on different retrieval paradigms. Observation-based and Model-based retrieval prove to be particularly effective in agent tasks, whose combination (ours) gives the best results.}
\centering
\begin{tabular}{l|cccc|cccc}
\toprule
Benchmark $\rightarrow$ & SWE & Web & OS & Spider2-V & SWE & Web & OS & Spider2-V\\
\midrule
Retrieval $\downarrow$ & \multicolumn{4}{c|}{Gemini-1.5-pro} & \multicolumn{4}{c}{Claude-3.5-sonnet} \\
\midrule
No retrieval & 13.3 & 17.9 & 4.9 & 8.3 & 51.2 & 35.8 & 12.4 & 8.4 \\
Instruction-based & 14.7 & 21.6 & 7.0 & 10.2 & 52.4 & 36.6 & 15.0 & 9.6 \\
Observation-based & 16.3 & 23.5 & 8.7 & 14.6 & 53.6 & 42.5 & 17.2 & 10.5 \\
Model-based & 17.0 & 24.3 & 9.5 & 15.4 & 57.8 & 44.8 & 20.3 & 13.7 \\
Ours & 18.7 & 25.6 & 10.3 & 16.4 & 60.0 & 48.0 & 22.5 & 16.6 \\
\bottomrule
\end{tabular}

\label{tab:retrieval}
\end{table}

%% file: tables/data_granularity.tex
\begin{table}[t!]
\caption{Effectiveness of synthetic data with various granularity. In general, short-trajectory data is more advantageous to both training and ICL, while mixing all of short, medium and long-trajectory data provides the best performance. }
\centering
\begin{tabular}{l|cccc|cc}
\toprule
Benchmark $\rightarrow$ & SWE & Web & OS & Spider2-V & Web & OS \\
\midrule
Granularity $\downarrow$ & \multicolumn{4}{c|}{Claude-3.5-sonnet} & \multicolumn{2}{c}{Codestral-22B} \\
\midrule
Baseline & 51.2 & 35.8 & 12.4 & 8.4 & 4.6 & 2.2 \\
Short & 54.2 & 39.4 & 17.9 & 10.8 & 13.5 & 4.9 \\
Medium & 53.6 & 38.8 & 16.6 & 9.7 & 12.6 & 4.0 \\
Long & 52.2 & 37.6 & 15.2 & 9.2 & 10.6 & 3.4 \\
Short+Medium & 54.6 & 41.2 & 18.8 & 11.3 & 14.6 & 5.7 \\
Short+Long & 54.0 & 40.5 & 17.8 & 10.7 & 14.4 & 5.3\\
Medium+Long & 53.8 & 38.6 & 17.2 & 10.4 & 13.2 & 4.5 \\
Short+Medium+Long & 55.0 & 42.0 & 19.8 & 12.3  & 15.4 & 6.3 \\
\bottomrule
\end{tabular}

\label{tab:data_granularity}
\end{table}

%% file: texts/related_work.tex
\section{Related work}

Various agents based on LLMs have been developed ~\citep{wang2024executable,zhang2024ui,shinn2024reflexion,huang2022language,wang2023voyager,wang2023describe}.
React~\citep{yao2022react} proposes to synergize reasoning and acting in LLMs.
By integrating Monte Carlo Tree Search~\citep{kocsis2006bandit,coulom2006efficient}, \citet{zhou2023language} leverages LLM-powered value functions and self-reflection~\citep{madaan2024self} to encourage proficient exploration and decision-making.
However, it comes with increased computational costs and relies on the premise that the environment allows for state reversals. 
In contrast, \ours removes such assumptions and improves both agent efficiency and performance by synthesizing high-quality data in advance.

Another line of research to improve agent models relies on training on human-labeled examples~\citep{zeng2023agenttuning,yin2023lumos,deng2024mind2web,chen2024agent,wang2022scienceworld} or data distilled from LLMs like GPT-4~\citep{chen2023fireact,zhao2024we}. 
\textcolor{rebuttal}{AgentGen~\citep{hu2024agentgen} explores automatic synthesis of both environments and tasks
and then leverages FastDownward\footnote{https://www.fast-downward.org/} to generate trajectory data.
AgentTuning~\citep{zeng2023agenttuning} utilizes both existing datasets and self-instruct~\citep{wang2022self} to derive instructions and then samples trajectories from GPT-4~\citep{achiam2023gpt}.
In contrast, \ours focuses on realistic environments and generate tasks and trajectories using backward construction.
}
Some other researchers are also exploring ways to use data more efficiently with reinforcement learning~\citep{ball2023efficient,schwarzer2020data,nachum2018data,thomas2016data,schwarzer2021pretraining}. 
\citet{gulcehre2023reinforced} suggests using data created by an LLM's policy can enhance itself via offline reinforcement learning algorithms. 
\citet{aksitov2023rest} takes this further by combining with ReAct~\citep{yao2022react} to train agent models iteratively on experience trajectories. 
These typically require a reward model as the scoring function or LLM/execution-generated feedback to enhance data quality. 
Our work, however, takes a different approach by employing the backward construction to improve the data quality by aligning instructions and trajectories.






%% file: texts/conclusion.tex
\section{Conclusion}
We introduce \ours, a data-centric framework to adapt LLM agents to any given environments without human annotations.
Based on commonly-accessible resources like documentaion, LLMs propose downstream tasks and complete them with multi-round interactions with environments.
We address the misalignment between instructions and trajectories by updating objectives with new instructions derived from trajectories.
Additionally, we design innovative retrieval approaches that leverage agent instructions, interaction histories, and current observations to retrieve synthesized examples.
Through extensive experiments, we demonstrate that the synthetic data from \ours significantly enhances model performance with both ICL and training.
Compared with other leading approaches in agent tasks, \ours shows much better performance with lower latency and computational costs, which make it particularly suitable for large-scale deployment.
Further analysis has also shown the superiority of \ours over the classical RAG.
In future work, we plan to explore multi-modal settings and train general agent models widely applicable in realistic environments.
We anticipate that \ours will inspire future research to push the state-of-the-art in this direction.

%% file: texts/limitations.tex
\section{\textcolor{rebuttal}{Limitations}}

\textcolor{rebuttal}{Although \ours effectively synthesizes high-quality agentic data with trajectories, it requires a lot of LLM calls in generation and filtering. 
We hope that future works will explore more efficient approaches to complete annotations without sacrificing quality.
Additionally, \ours leverages the environment-related resources to generate instructions.
In some scenarios, however, these resources may be incomplete or not available.}

%% file: texts/appendix.tex

\section{Baseline implementations}
\label{app:baseline_implementation}
We follow the existing frameworks to set up baselines in each benchmark.
In SWE-bench~\citep{jimenez2023swe}, we follow CodeAct~\citep{openhands}, where LLMs interact with environments to solve problems.
In WebArena~\citep{zhou2023webarena}, we follow the implementation in \citet{workarena2024}, which concatenates task objectives, action space descriptions, general instructions (e.g., output formats) and webpage observations in the prompt, and ask LMs to predict the next action. By default, we use the accessibility tree\footnote{\url{https://developer.mozilla.org/en-US/docs/Glossary/Accessibility_tree}} as the observation space.
In OSWorld~\citep{xie2024osworld} and Spider2-V~\citep{cao2024spider2}, we follow the original prompt style designed by the benchmark, which also concatenates task objectives, action space descriptions, general instructions and computer observations in the prompt.
By default, we use the accessibility tree as the observation space for OSWorld, and use the set-of-mark for Spider2-V due to the significant information loss of the accessibility tree in the original benchmark.
See an example in Table~\ref{tab:observation_space_spider2v_1} and \ref{tab:observation_space_spider2v_2} for more details.

\section{Dataset examples}
\label{app:data_examples}
From Table~\ref{tab:webarena_1_visual} to \ref{tab:spider2v_6}, we provide one example for each dataset with full instructions, interaction history with the environment.




\section{Experimental settings}
\label{app:experiment_settings}
We retrieve documents until the maximum length of LLMs for RAG and set an upper bound number of 50 documents, where the retrieved documents remain unchanged throughout agent interaction trajectory because only instructions are used as the query for retrieval.
For Reflexion~\citep{shinn2024reflexion}, we use the maximum trials 3.
In LATS~\citep{zhou2023language}, we use the number of generated action 5, depth limit 15, value function weight 0.8, following the original setting in paper with WebShop~\citep{yao2022webshop}, which is also an agent task based on website.
By default, we use https://cloud.google.com/vertex-ai/generative-ai/docs/embeddings/get-text-embeddings as the dense retriever for model-based retrieval.
We use the temperature 0 throughout the experiments to ensure better reproductivity of the experiments.
During training, we the batch size 128, learning rate 0.00002, warmup ratio 0.03 and maximum length 8192, and tune the model for 3 epochs.
All experiments are conducted in H100 machines with 80GB memeory.



\section{Document sources}
\label{app:document_sources}
We use all the non-repeated python files in SWE-bench-Verified~\citep{jimenez2023swe} as the document sources.
Although we may not always find abundant documentations and tutorials for each environment, we believe that documentations in the same domain still have a good coverage of frequent operations.
For example, one subset of WebArena~\citep{zhou2023webarena} focuses on the navigation of the shopping website OneStopMarket, we use the Amazon documentation as a good replacement.
Regardless of the shopping websites, the frequent tasks usually include order change, product search, delivery checking, etc.
Therefore, we use other documentations in the same domain to sample task instructions when the exact version for the target environment is not available.
Concretely, we use the following sources for WebArena:
\begin{itemize}
    \item https://docs.gitlab.com/ee/tutorials/
    \item https://support.google.com/maps
    \item https://www.amazon.com/hz/contact-us/foresight/hubgateway
    \item https://support.reddithelp.com/hc/en-us/articles
\end{itemize}
The following sources are used for OSWorld:
\begin{itemize}
    \item https://support.google.com/chrome/?hl=en
    \item https://www.gimp.org/tutorials/
    \item https://books.libreoffice.org/en/CG72/CG72.html
    \item https://books.libreoffice.org/en/WG73/WG73.html
    \item https://ubuntu.com/tutorials/command-line-for-beginners
    \item https://support.mozilla.org/en-US/products/thunderbird
    \item https://wiki.videolan.org/Documentation:Documentation
    \item https://code.visualstudio.com/docs
\end{itemize} ,
The following sources are used for Spider2-V:
\begin{itemize}
    \item https://docs.getdbt.com/
    \item https://release-1-7-2.dagster.dagster-docs.io/
    \item https://docs.astronomer.io/
    \item https://docs.airbyte.com/
    \item https://airbyte.com/tutorials/
    \item https://airbyte-public-api-docs.s3.us-east-2.amazonaws.com/rapidoc-api-docs.html
    \item https://superset.apache.org/docs/
    \item https://www.metabase.com/docs/v0.49/
    \item https://www.metabase.com/learn/
    \item https://docs.snowflake.com/en/
    \item https://cloud.google.com/bigquery/docs/
    \item https://jupyterlab.readthedocs.io/en/4.1.x/
\end{itemize}

\section{Synthesized data examples}
\label{app:data_synthesis_example}
From Table~\ref{tab:data_synthesis_bigquery_1} to \ref{tab:data_synthesis_bigquery_7}, we provide a complete example of data synthesis.
To begin with, an LLM generates instructions based on standard resources like tutorials, documentations and FAQs: Upload CSV data in Google Drive to BigQuery. (See prompt in Table~\ref{tab:self_instruct_prompt})
It then attempts solve the task by predicting actions and collecting feedback from environments (interactions).
This produces a long trajectory showing how LLMs try to achieve the goal.

However, it is not guaranteed that the trajectory successfully achieves the target.
In our example, the LLM makes a wrong prediction in the action 4.
It selects the table source Google Cloud Storage, while the correct action should select ``Drive" to align with the instruction that reuiqres to upload CSV data in Google Drive.
This results in wrong actions in the subsequent predictions, and the generated trajectory is not aligned with the initial instruction, which leads to noisy data in this case.

Instead of using the original instruction-trajectory pairs for downstream training and in-context learning, we fix the mentioned misalignment by crafting new instructions for each sub-trajectory (backward construction).
Concretely, we feed the generated trajectory into LLM prompts, and ask it to summarize the trajectory or propose a new task based on it.
For example, the LLM updates the task objective to ``Link CSV file in Google Cloud Storage to BigQuery" after observing the trajectory, which makes the task instrucion and the trajectory aligned.
Additionally, we also generate new instructions for each sub-trajectory, which would increase the utility of a generated full trajectory.
For instance, based on the sub-trajectory (observation 0, Action 1, observation 1), the LLM generates a new instruction: When is dataset ``demo" created?
In Table~\ref{tab:generated_instructions_1} and \ref{tab:generated_instructions_2}, we list more generated instructions based on sub-trajectories.

\section{\textcolor{rebuttal}{Case study on filtered examples}}
\textcolor{rebuttal}{In Table~\ref{tab:osworld_filtered_1}-\ref{tab:webarena_filtered_4}, we demonstrate the representative synthesized examples that fail to meet our designed criteria.
The example in Table~\ref{tab:osworld_filtered_1}-\ref{tab:osworld_filtered_6} is filtered because the trajectory shows detour in accomplishing the goal, i.e. Action 1-6 are not necessary.
The example in Table~\ref{tab:webarena_filtered_1}-\ref{tab:webarena_filtered_4} is filtered because it goes back and forth in states, i.e. repeat the actions of clicking "My Orders" and clicking "View Order".
We filter these low-quality examples to avoid their negative influences in the downstream applications.}

\section{Synthesized data from environments}
We compare \ours with the version without relying on existing resources in WebArena. 
Except for sampling task instructions from LLMs based on given environments, we follow the same procedures in Learn-by-interact to synthesize 10k examples.
The results of in-context learning with Claude-3.5-sonnet are shown  in Table \ref{tab:envdata}
\input{tables/envdata}

However, we note the following potential concerns regarding the version without replying on existing resources: the distribution and the diversity of the generated data are hard to control. 
Without conditioning on prior documents, one will need intensive prompt engineering to guide LLMs in generating diverse task instructions. 
On the other hand, the related resources are usually crafted by experts or written by real users, which cover most important scenarios that people who interact with the environment are interested in.

\section{Cross-website generalization}
To evaluate the generalization capabilities of \ours, in WebArena, we consider the content management systems (CMS) as a held-out test set and leverage the synthetic data from the remaining websites as the training data. To ensure a fair comparison and avoid the influences of the discrepancies in the training set size, we downsample the original data that covers all the WebArena domains so that both set contains the same number of instances. Following the same evaluation pipelines in the paper, we assess the performance of in-context learning with Claude-3.5-sonnet and training with Codestral-22B.
\input{tables/crosswebsite}

From the results in Table~\ref{tab:crosswebsite}, we observe that, even without specifically using the data sampled from the CMS environment, Learn-by-interact demonstrates significant improvements over the baseline in both training and in-context learning. This indicates that the proposed approach holds the potential for cross-website generalization and is likely to achieve better performance when utilizing data from more websites.

\input{tables/data_samples/webarena}

\input{tables/data_samples/osworld}

\input{tables/data_samples/spider2v}

\input{tables/observation_space_spider2v}

\input{tables/data_samples/data_synthesis_bigquery}

\input{tables/prompts/self_instruct}
\input{tables/prompts/sub_trajectory}

\input{tables/prompts/filter}
\input{tables/prompts/inference}

\input{tables/prompts/write_query}

\input{tables/data_samples/osworld_filtered}

\input{tables/data_samples/webarena_filtered}

%% file: tables/envdata.tex
\begin{table}[t!]
\caption{The comparison of \ours to the version without replying on existing resources.}
\centering
\begin{tabular}{l|ccc}
\toprule
Number of synthesized examples & 0 & 5k & 10k \\
\midrule
Task instructions generated based on environments & 35.8 & 37.3 & 39.6 \\
Task instructions generated based on related resources & 35.8 & 43.2 & 48.0 \\
\bottomrule
\end{tabular}

\label{tab:envdata}
\end{table}

%% file: tables/crosswebsite.tex
\begin{table}[t!]
\caption{Results for cross-website generalization.}
\centering
\begin{tabular}{l|cc}
\toprule
Model & Claude-3.5-sonnet & Codestral-22B \\
\midrule
Baseline & 26.0 & 3.3 \\
Learn-by-interact with synthetic data that excludes CMS	& 28.3	& 12.6 \\
Learn-by-interact with all WebArena data that contains CMS & 29.2	& 17.6 \\
\bottomrule
\end{tabular}

\label{tab:crosswebsite}
\end{table}

%% file: tables/data_samples/webarena.tex
\begin{table}[ht]
\caption{\textbf{Webarena example}}
\centering
\begin{tabular}{p{13cm}}
\toprule
\textbf{Instruction} \\
\midrule
Tell me the total cost of my latest cancelled order? \\
\midrule
\textbf{Observation 0} \\
\midrule
\includegraphics[width=0.8\textwidth]{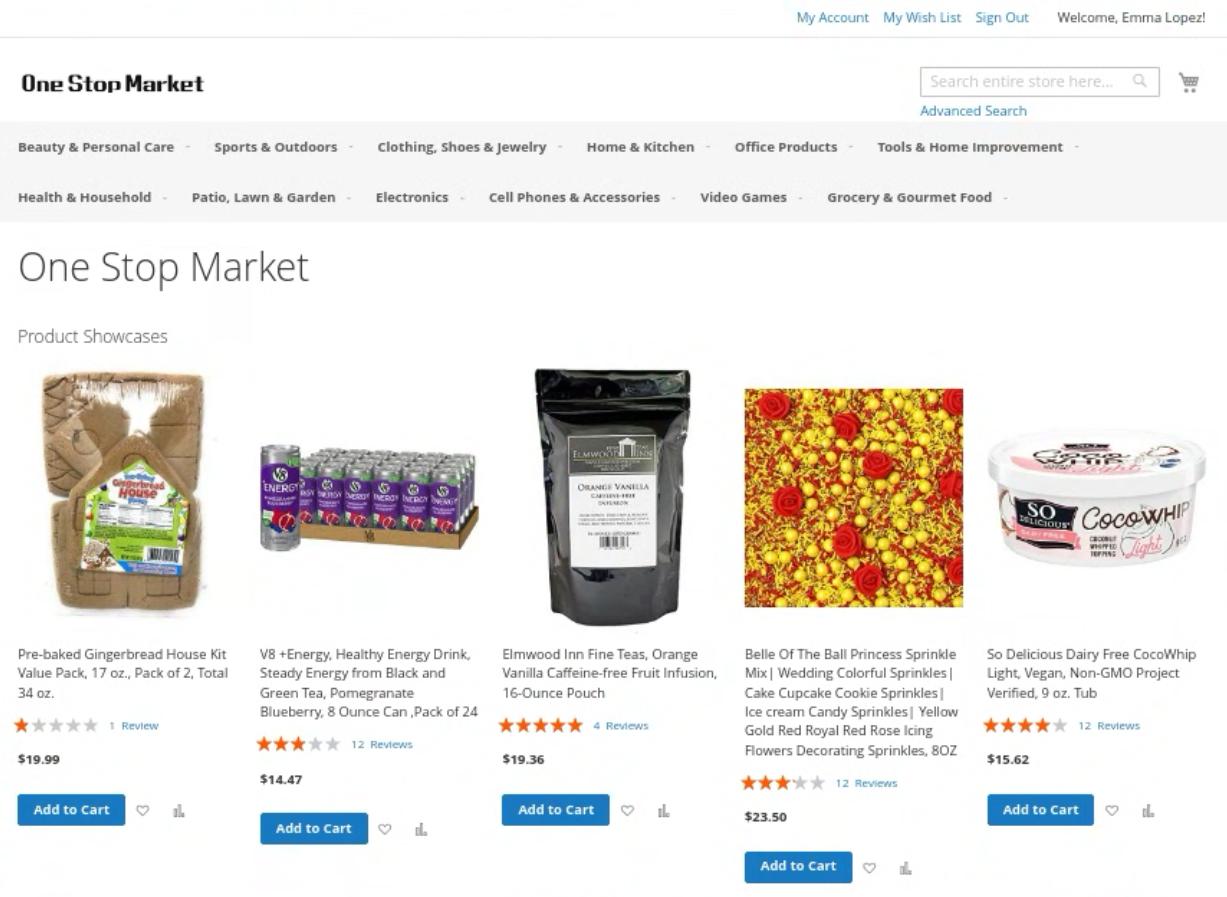} \\
\midrule
\textbf{Action 1} \\
\midrule
click [4918] // click the button `My Account' \\
\bottomrule
\end{tabular}

\label{tab:webarena_1_visual}
\end{table}

\begin{table}[ht]
\caption{\textbf{Webarena example cont.}}
\centering
\begin{tabular}{p{13cm}}
\toprule
\textbf{Observation 1} \\
\midrule
\includegraphics[width=0.8\textwidth]{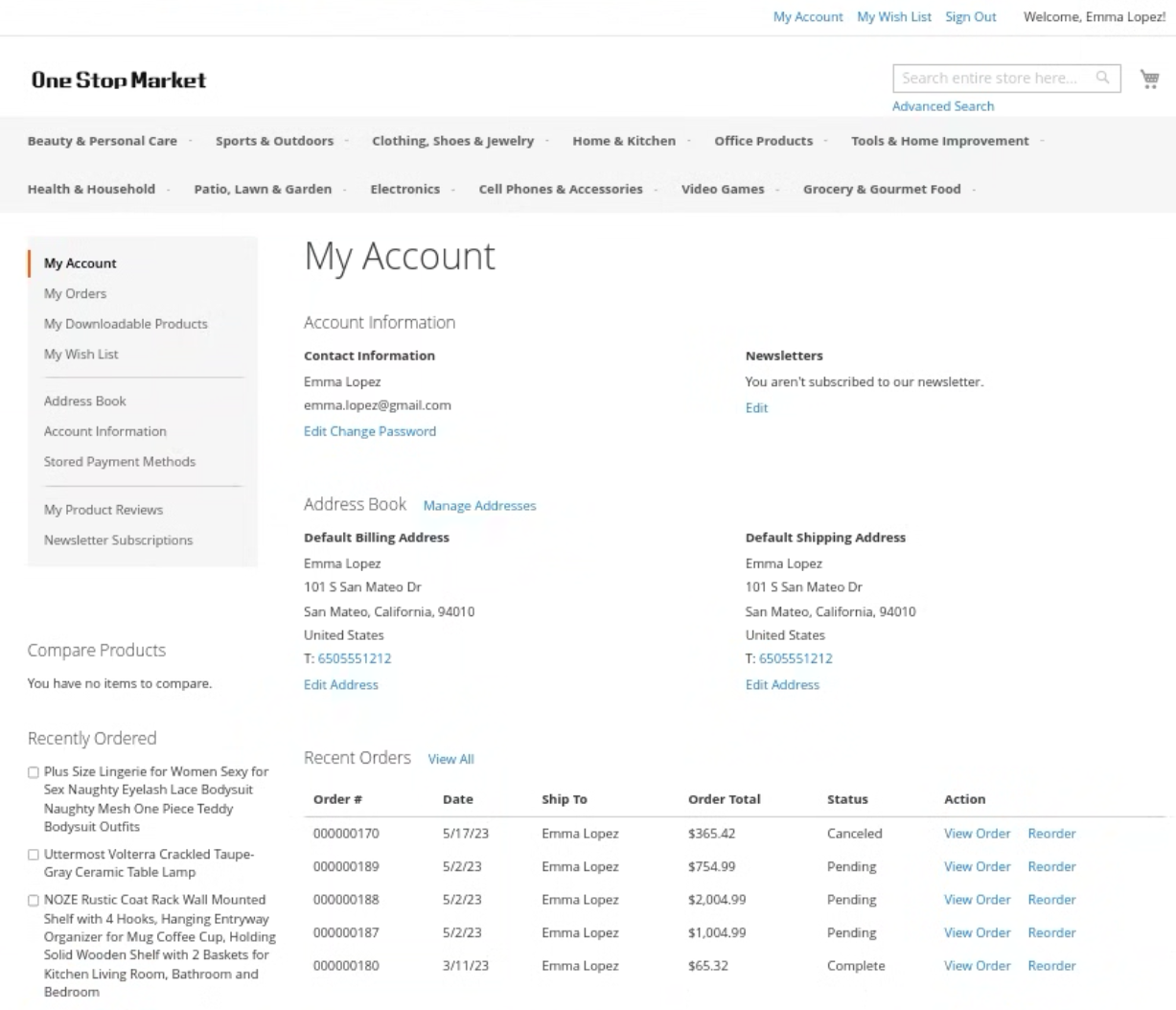} \\
\midrule
\textbf{Action 2} \\
\midrule
Stop: The total cost of the latest cancelled order is \$365.42\\
\bottomrule
\end{tabular}

\label{tab:webarena_2_visual}
\end{table}

%% file: tables/data_samples/osworld.tex
\begin{table}[ht]
\caption{\textbf{OSWorld example}}
\centering
\begin{tabular}{p{13cm}}
\toprule
\textbf{Instruction} \\
\midrule
Could you assist me in adding a new layer and naming it `Square'? \\
\midrule
\textbf{Observation 0 (Interface of the software GIMP)} \\
\midrule
\includegraphics[width=0.8\textwidth]{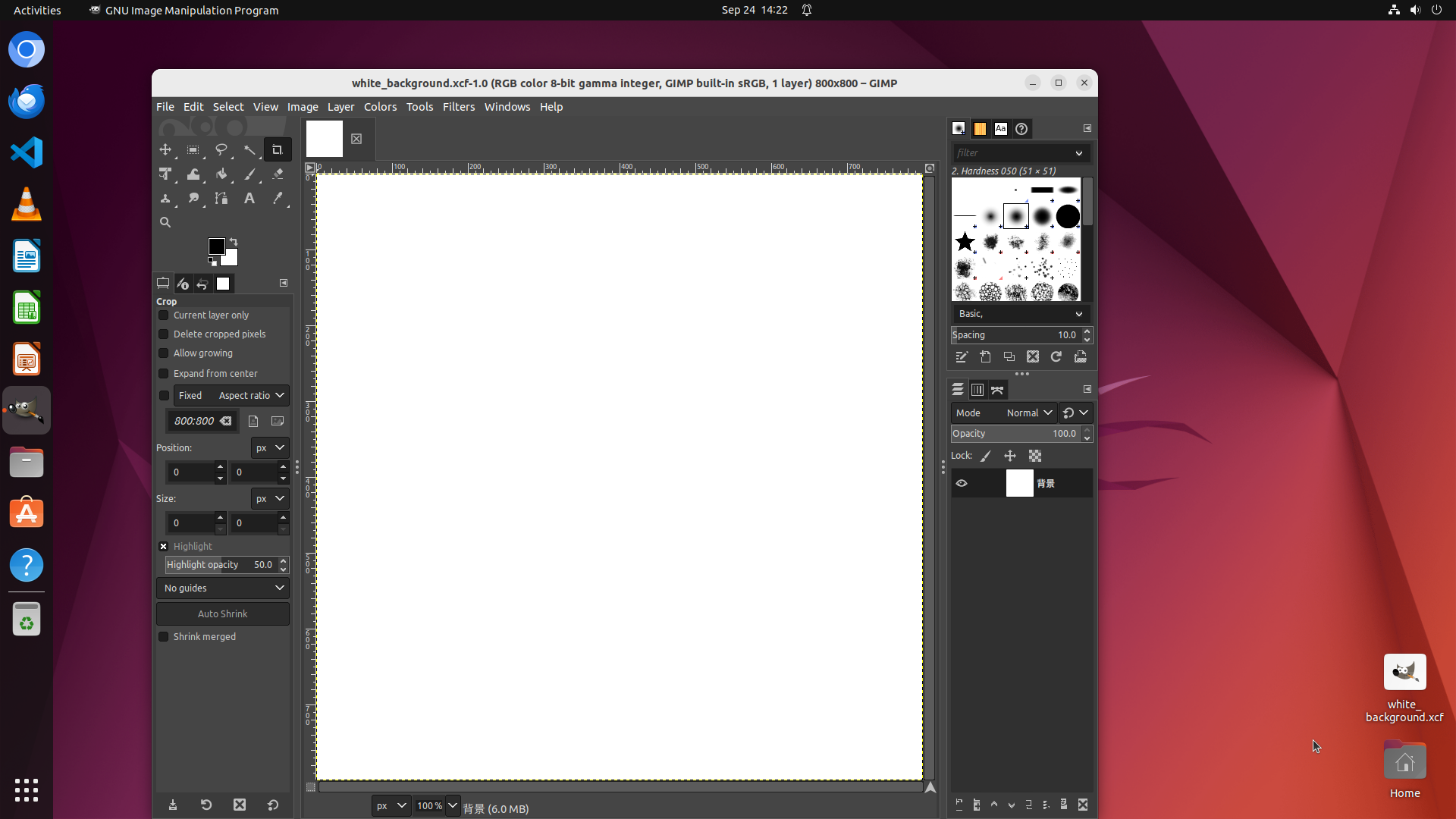} \\
\midrule
\textbf{Action 1} \\
\midrule
import pyautogui \\
pyautogui.hotkey(`shift', `ctrl', `n') // shortcut to initialize a new layer. \\
\midrule
\textbf{Observation 1 (Interface of the software GIMP)} \\
\midrule
\includegraphics[width=0.8\textwidth]{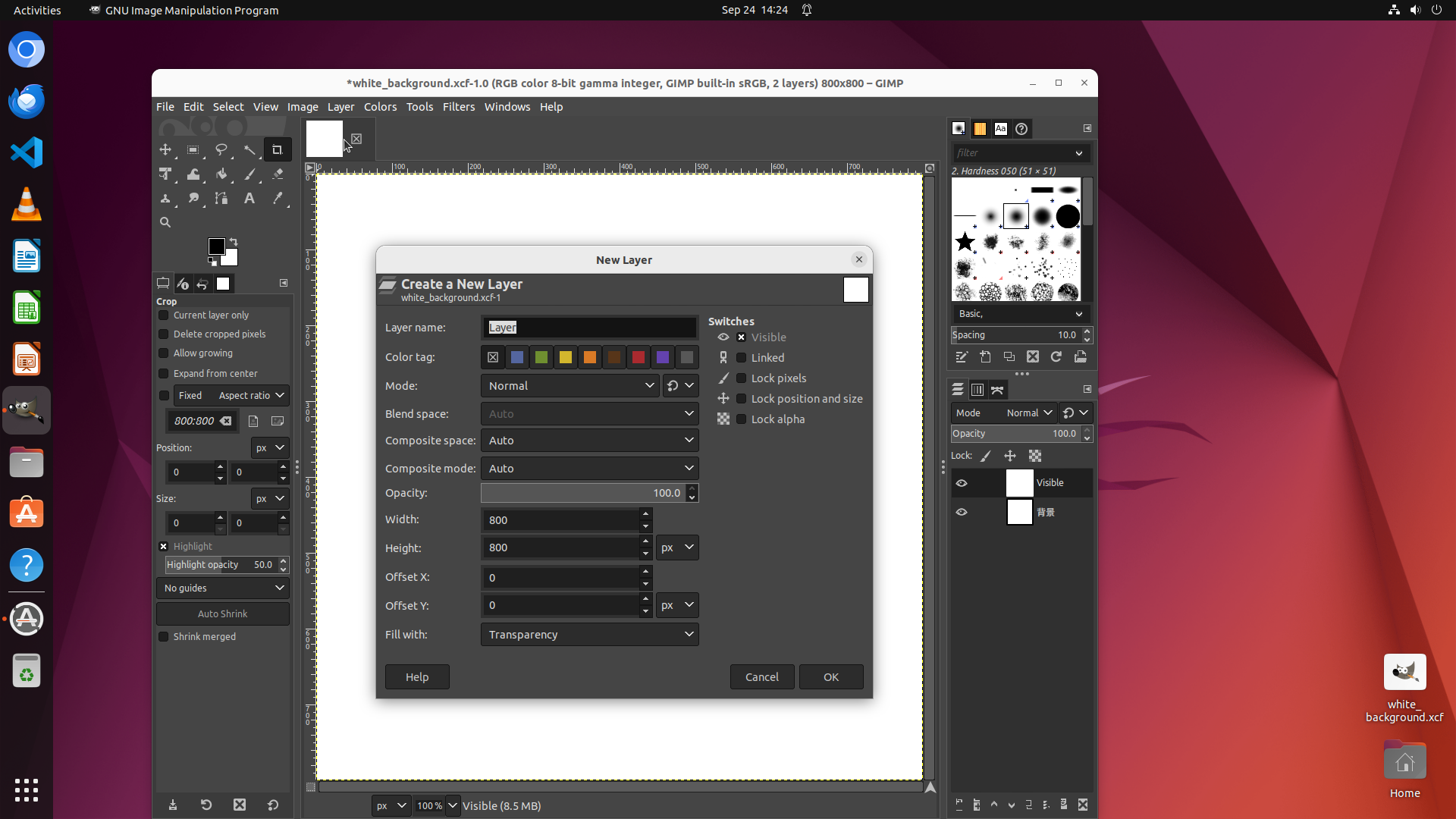} \\
\midrule
\bottomrule
\end{tabular}

\label{tab:osworld_1_visual}
\end{table}

\begin{table}[ht]
\caption{\textbf{OSWorld example cont.}}
\centering
\begin{tabular}{p{13cm}}
\toprule
\textbf{Action 2} \\
\midrule
import pyautogui \\
pyautogui.typewrite(`Square') // change the layer name to Square. \\
\midrule
\textbf{Observation 2 (Interface of the software GIMP)} \\
\midrule
\includegraphics[width=0.8\textwidth]{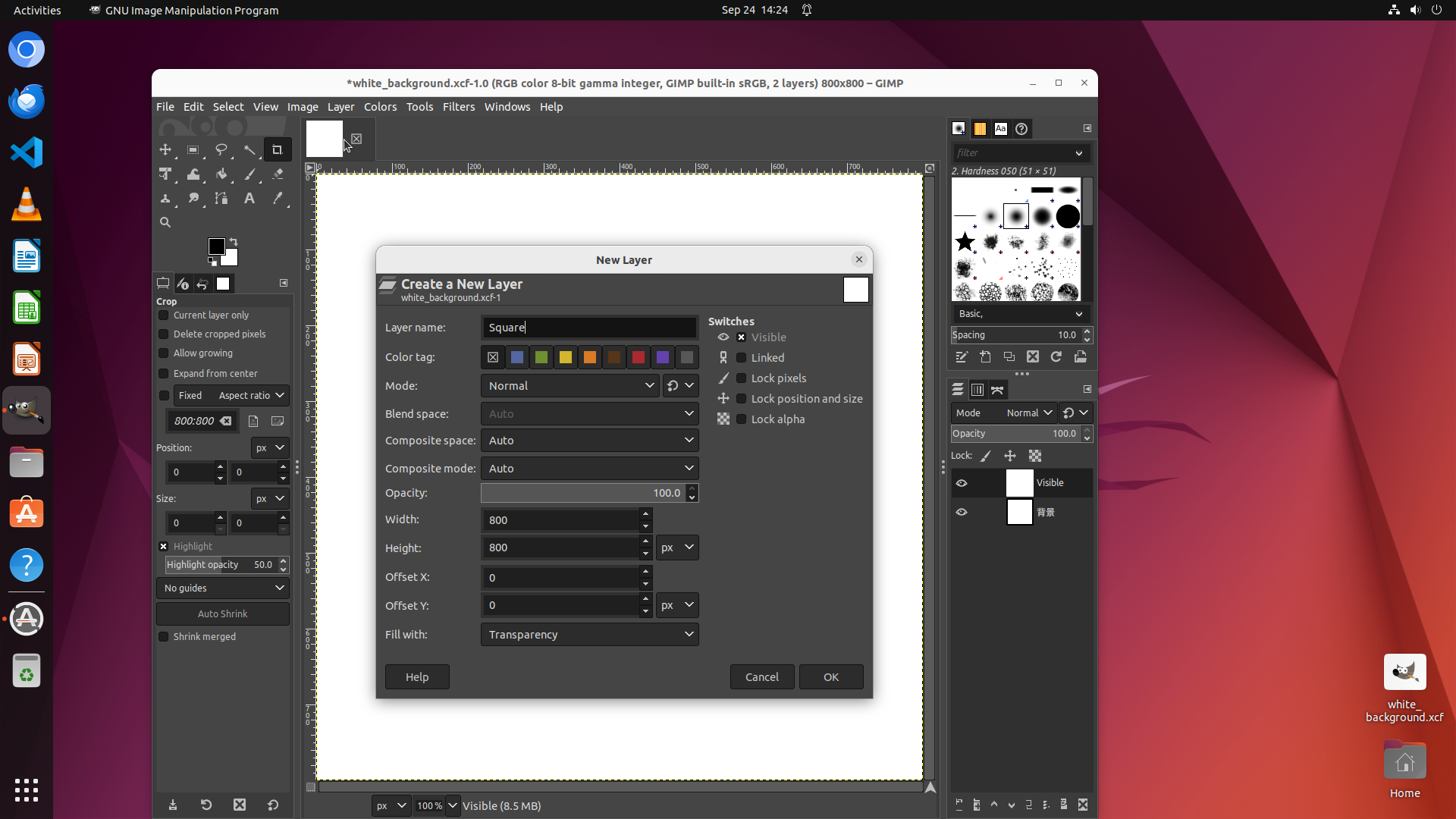} \\
\midrule
\textbf{Action 3} \\
\midrule
import pyautogui \\
pyautogui.click(1096, 892) // click OK to save the change. \\
\midrule
\textbf{Observation 3 (Interface of the software GIMP)} \\
\midrule
\includegraphics[width=0.8\textwidth]{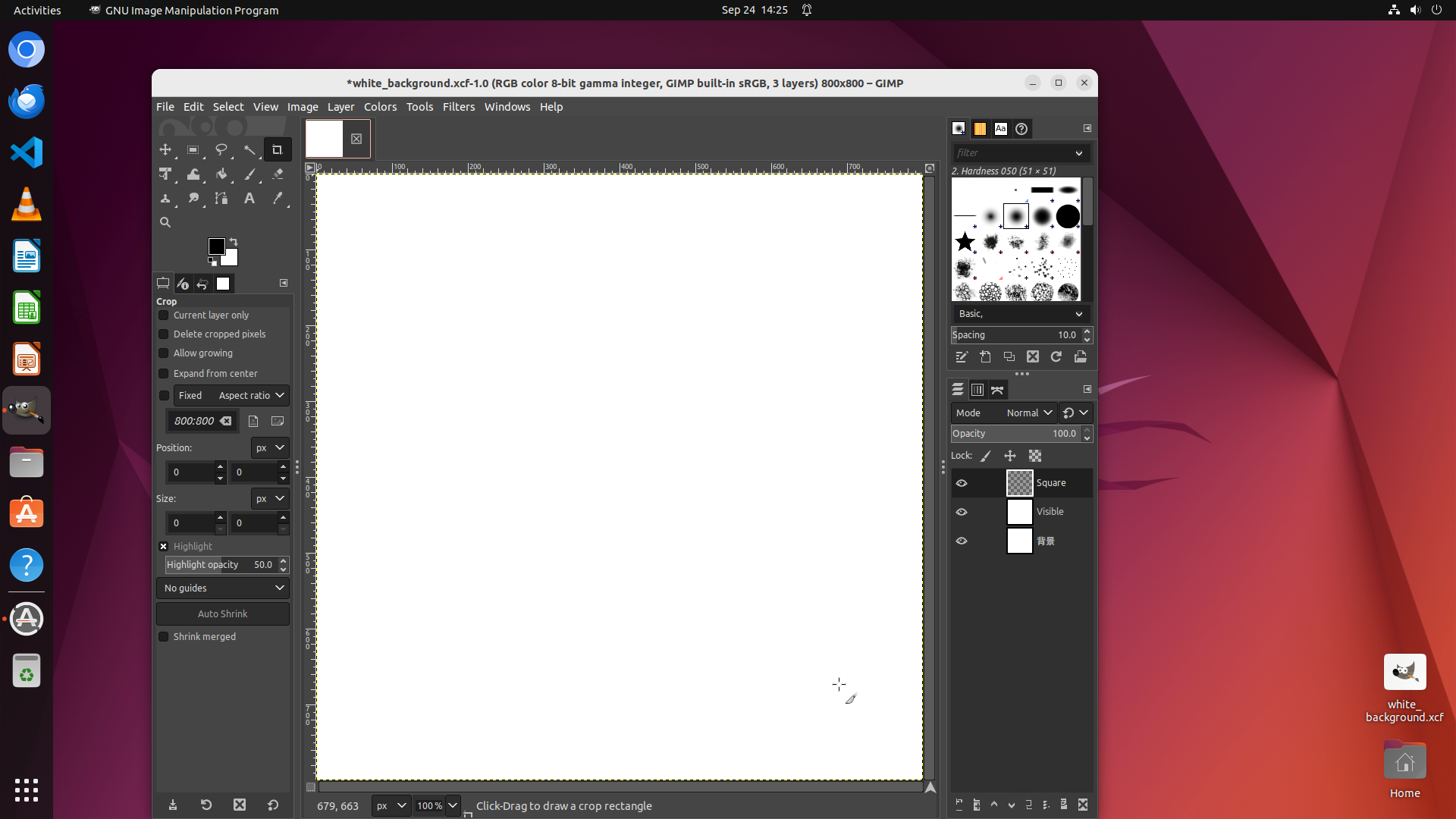} \\
\midrule
\bottomrule
\end{tabular}

\label{tab:osworld_2_visual}
\end{table}

%% file: tables/data_samples/spider2v.tex
\begin{table}[ht]
\caption{\textbf{Spider2-V example, cont.}}
\centering
\begin{tabular}{p{13cm}}
\toprule
\textbf{Instruction} \\
\midrule
I have established a connection from Faker to local .csv file. Could you help me change the running schedule? I hope it can be replicated at 18:00 pm every day. \\
\midrule
\textbf{Observation 0 (Interface of the software Airbyte)} \\
\midrule
\includegraphics[width=0.8\textwidth]{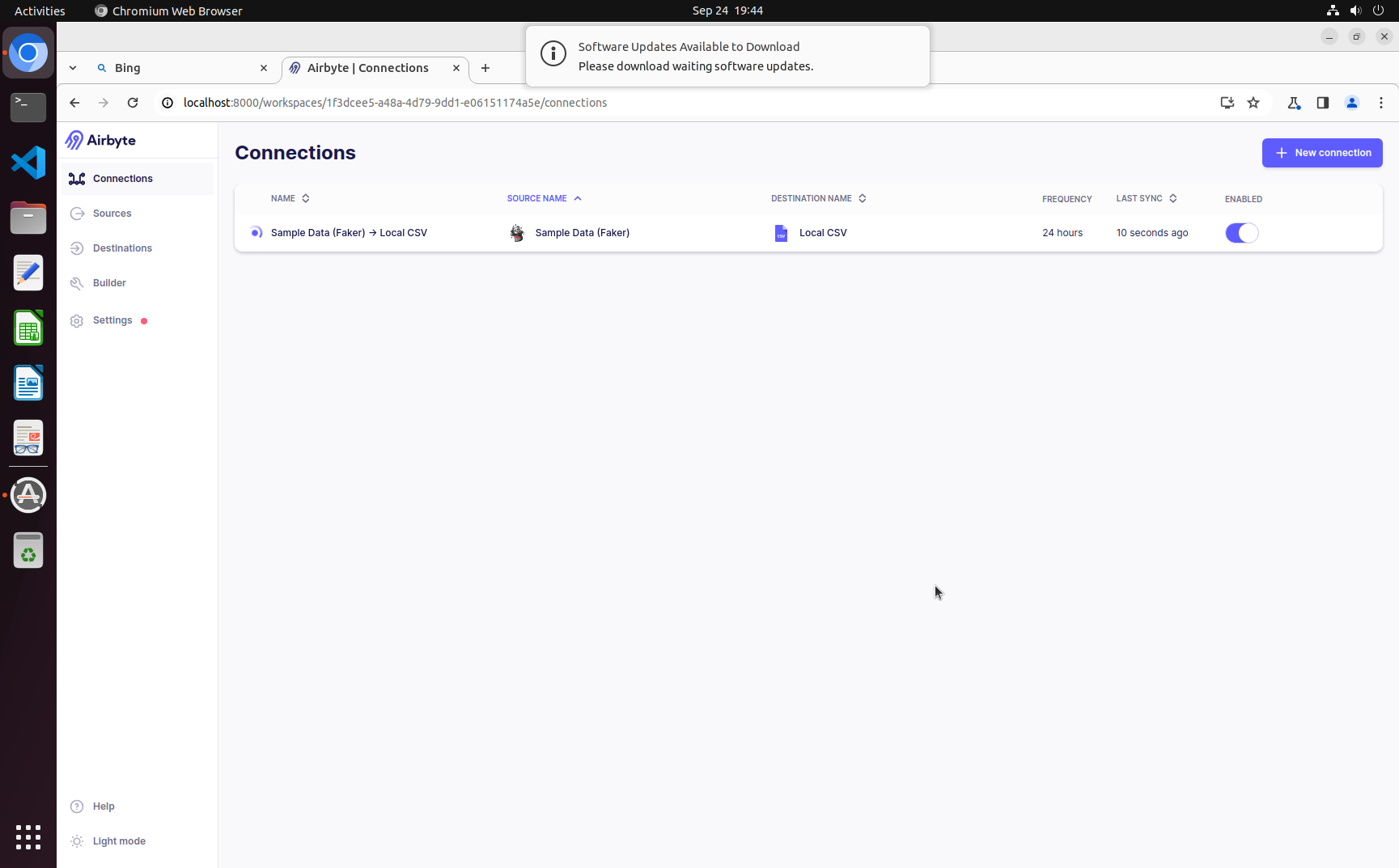} \\
\midrule
\textbf{Action 1} \\
\midrule
import pyautogui \\
pyautogui.click(550,280) // click the connection row with the name ``Sample Data (Faker) $\rightarrow$ Local CSV" \\
\midrule
\textbf{Observation 1 (Interface of the software Airbyte)} \\
\midrule
\includegraphics[width=0.8\textwidth]{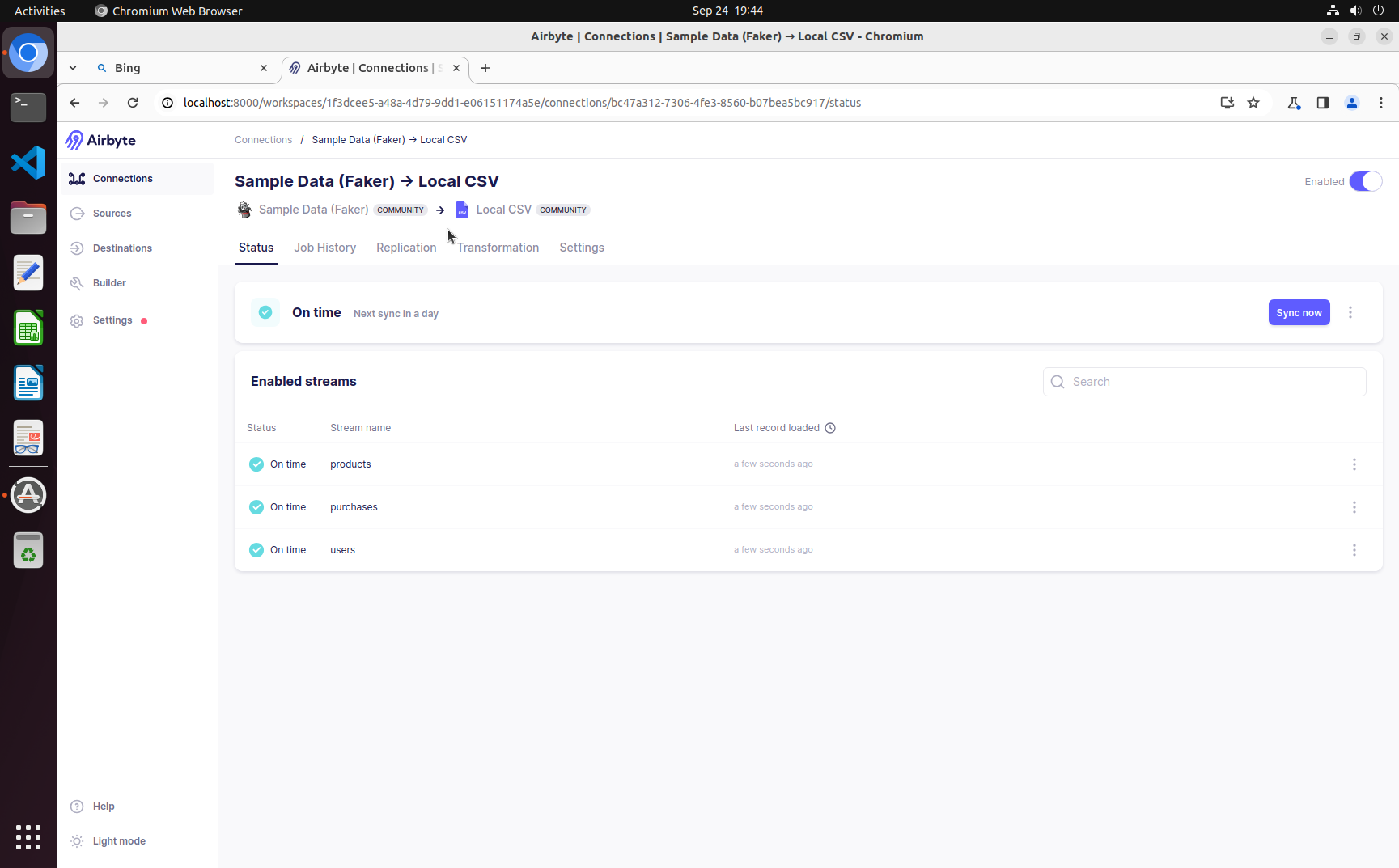} \\
\midrule
\bottomrule
\end{tabular}

\label{tab:spider2v_1}
\end{table}

\begin{table}[ht]
\caption{\textbf{Spider2-V example, cont.}}
\centering
\begin{tabular}{p{13cm}}
\toprule
\textbf{Action 2} \\
\midrule
import pyautogui \\
pyautogui.click(500,300) // click the button Replication \\
\midrule
\textbf{Observation 2 (Interface of the software Airbyte)} \\
\midrule
\includegraphics[width=0.8\textwidth]{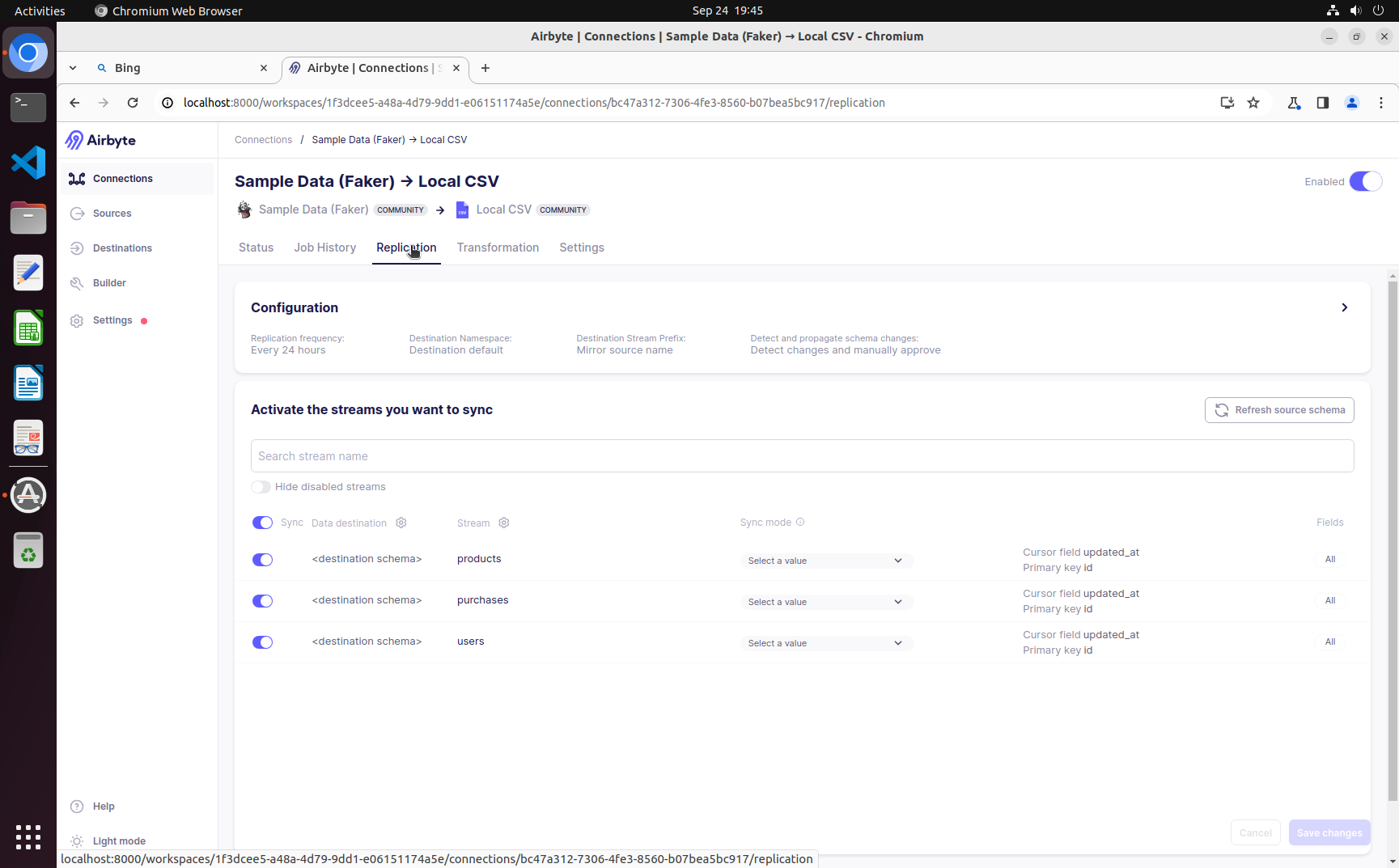} \\
\midrule
\textbf{Action 3} \\
\midrule
import pyautogui \\
pyautogui.click(500,450) // click the panel Configuration \\
\midrule
\textbf{Observation 3 (Interface of the software Airbyte)} \\
\midrule
\includegraphics[width=0.8\textwidth]{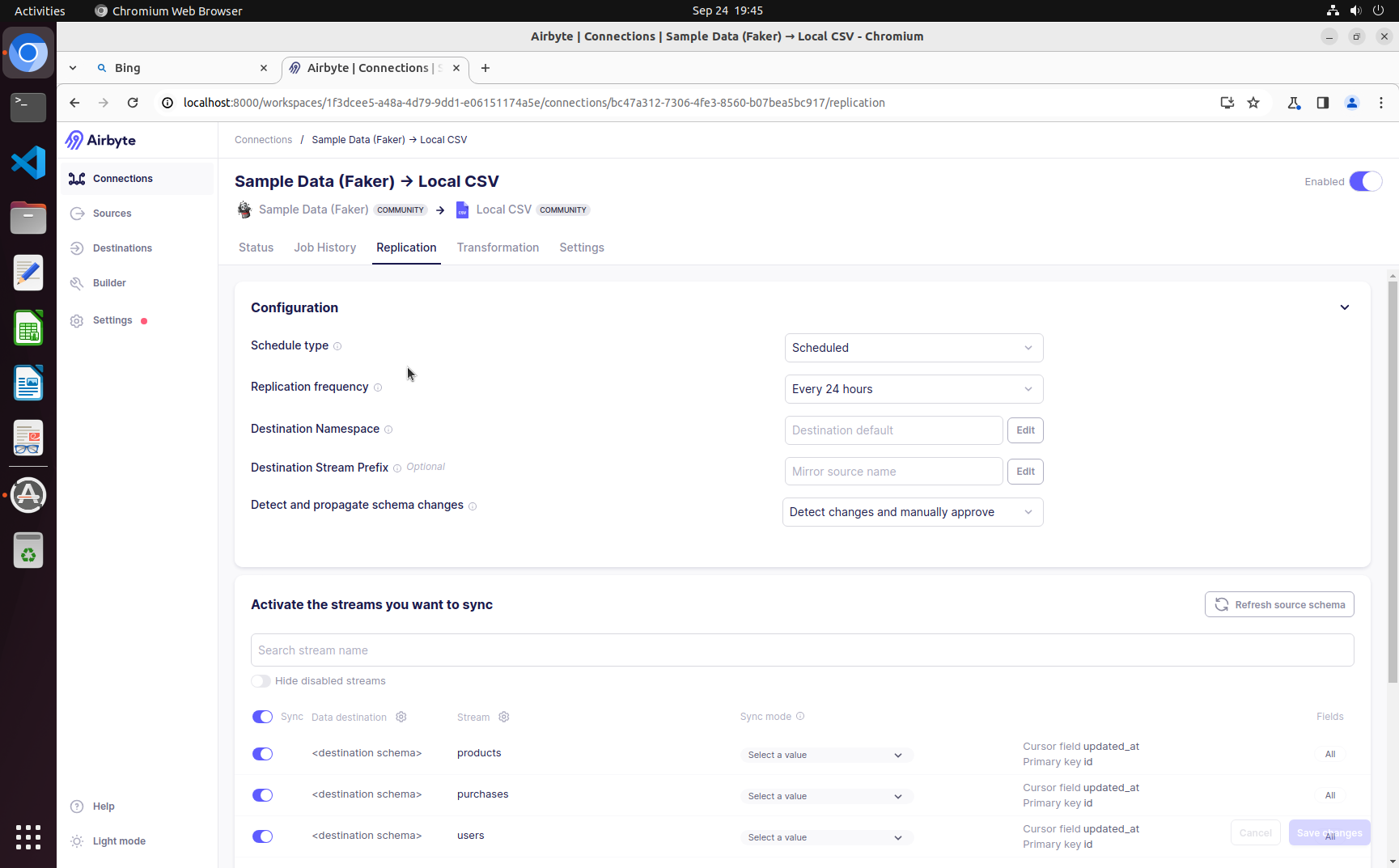} \\
\midrule
\bottomrule
\end{tabular}

\label{tab:spider2v_2}
\end{table}

\begin{table}[ht]
\caption{\textbf{Spider2-V example, cont.}}
\centering
\begin{tabular}{p{13cm}}
\toprule
\textbf{Action 4} \\
\midrule
import pyautogui \\
pyautogui.click(1270,430) // reveal the dropdown menu of the schedule type \\
\midrule
\textbf{Observation 4 (Interface of the software Airbyte)} \\
\midrule
\includegraphics[width=0.8\textwidth]{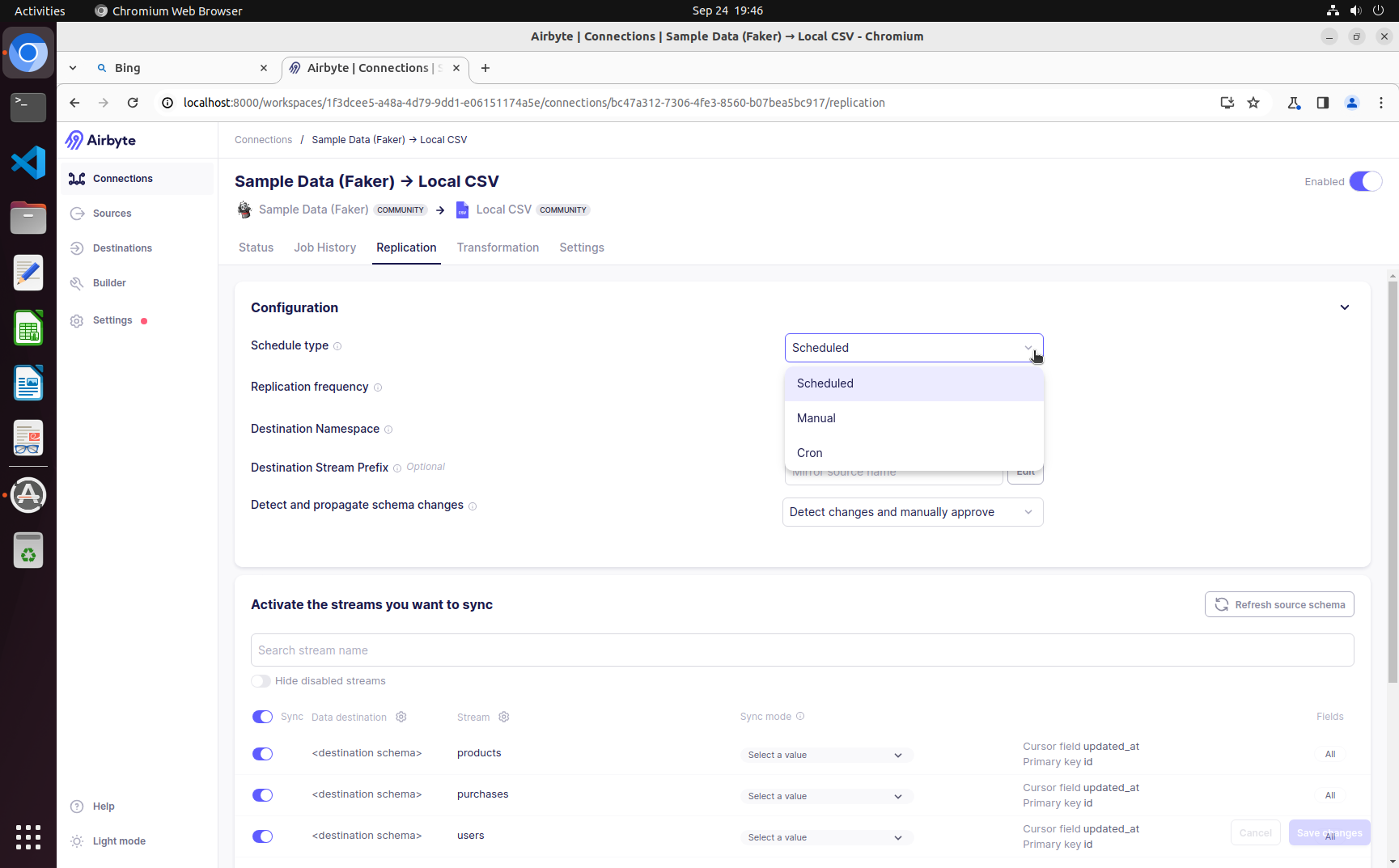} \\
\midrule
\textbf{Action 5} \\
\midrule
import pyautogui \\
pyautogui.click(1200,565) // select the schedule type Cron \\
\midrule
\textbf{Observation 5 (Interface of the software Airbyte)} \\
\midrule
\includegraphics[width=0.8\textwidth]{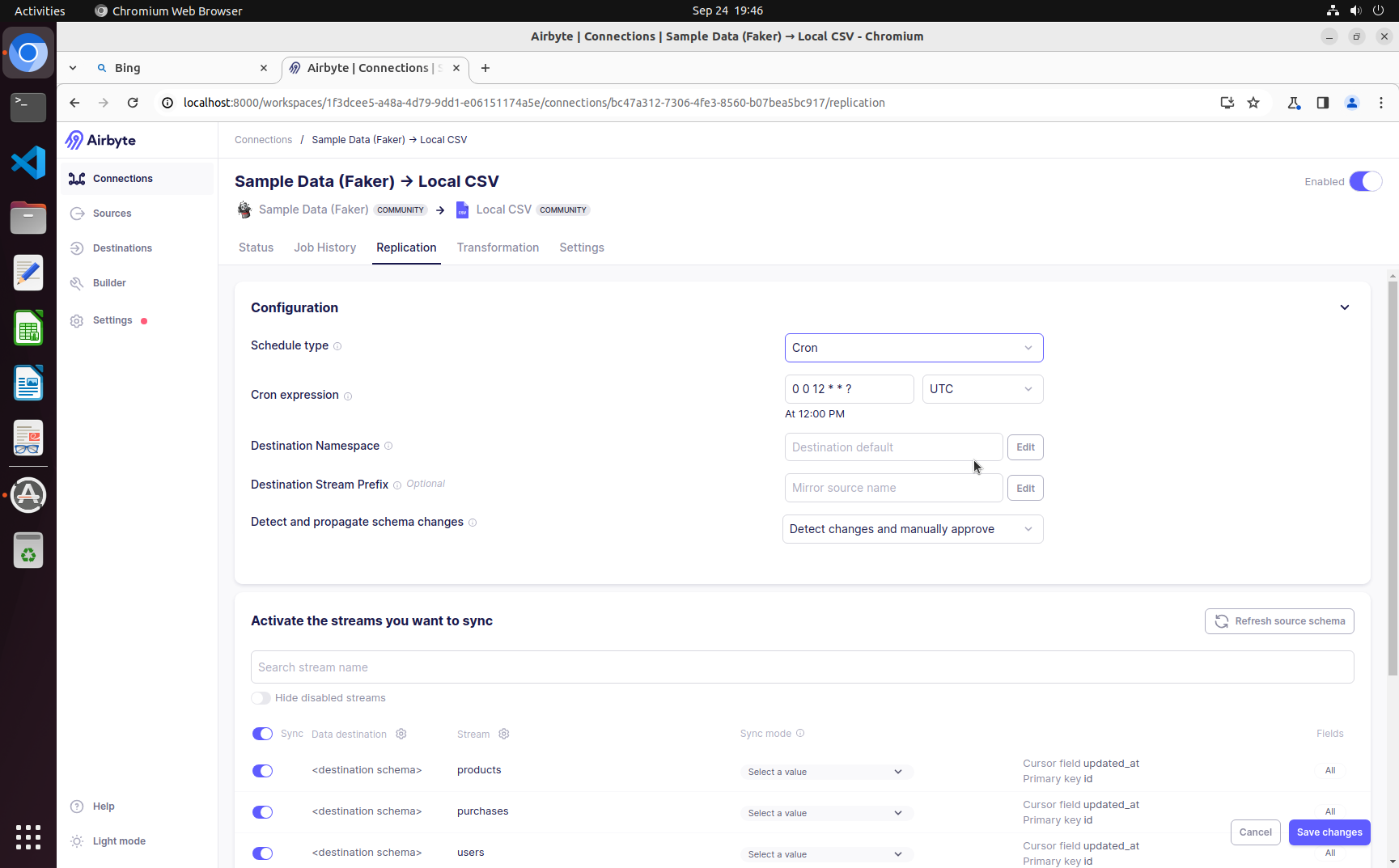} \\
\midrule
\bottomrule
\end{tabular}

\label{tab:spider2v_3}
\end{table}

\begin{table}[ht]
\caption{\textbf{Spider2-V example, cont.}}
\centering
\begin{tabular}{p{13cm}}
\toprule
\textbf{Action 6} \\
\midrule
import pyautogui \\
pyautogui.click(1020,480) // focus on the content of the Cron expression  \\
\midrule
\textbf{Observation 6 (Interface of the software Airbyte)} \\
\midrule
\includegraphics[width=0.8\textwidth]{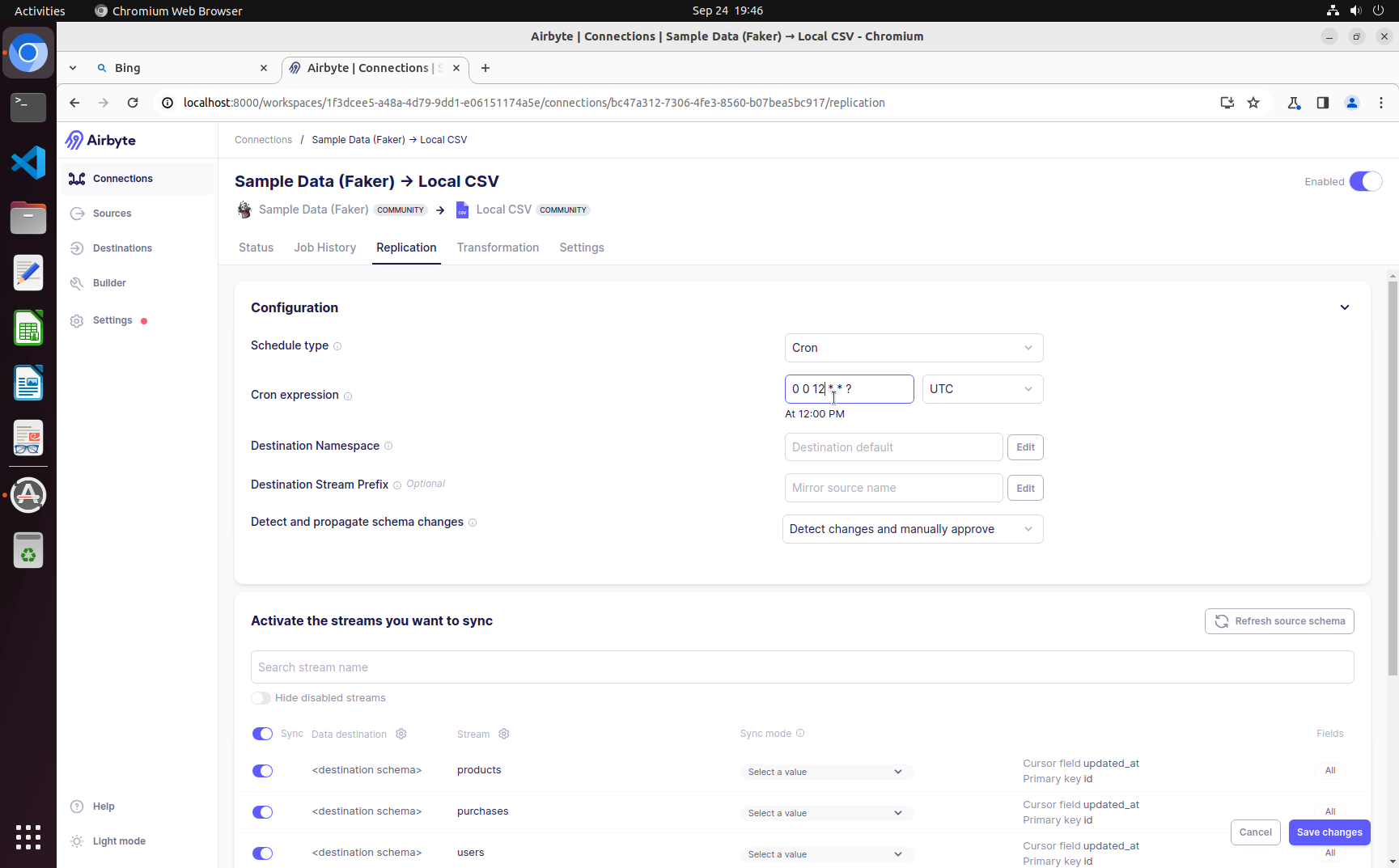} \\
\midrule
\textbf{Action 7} \\
\midrule
import pyautogui \\
pyautogui.hotkey(``ctrl",``a") // select all content in the textbox \\
\midrule
\textbf{Observation 7 (Interface of the software Airbyte)} \\
\midrule
\includegraphics[width=0.8\textwidth]{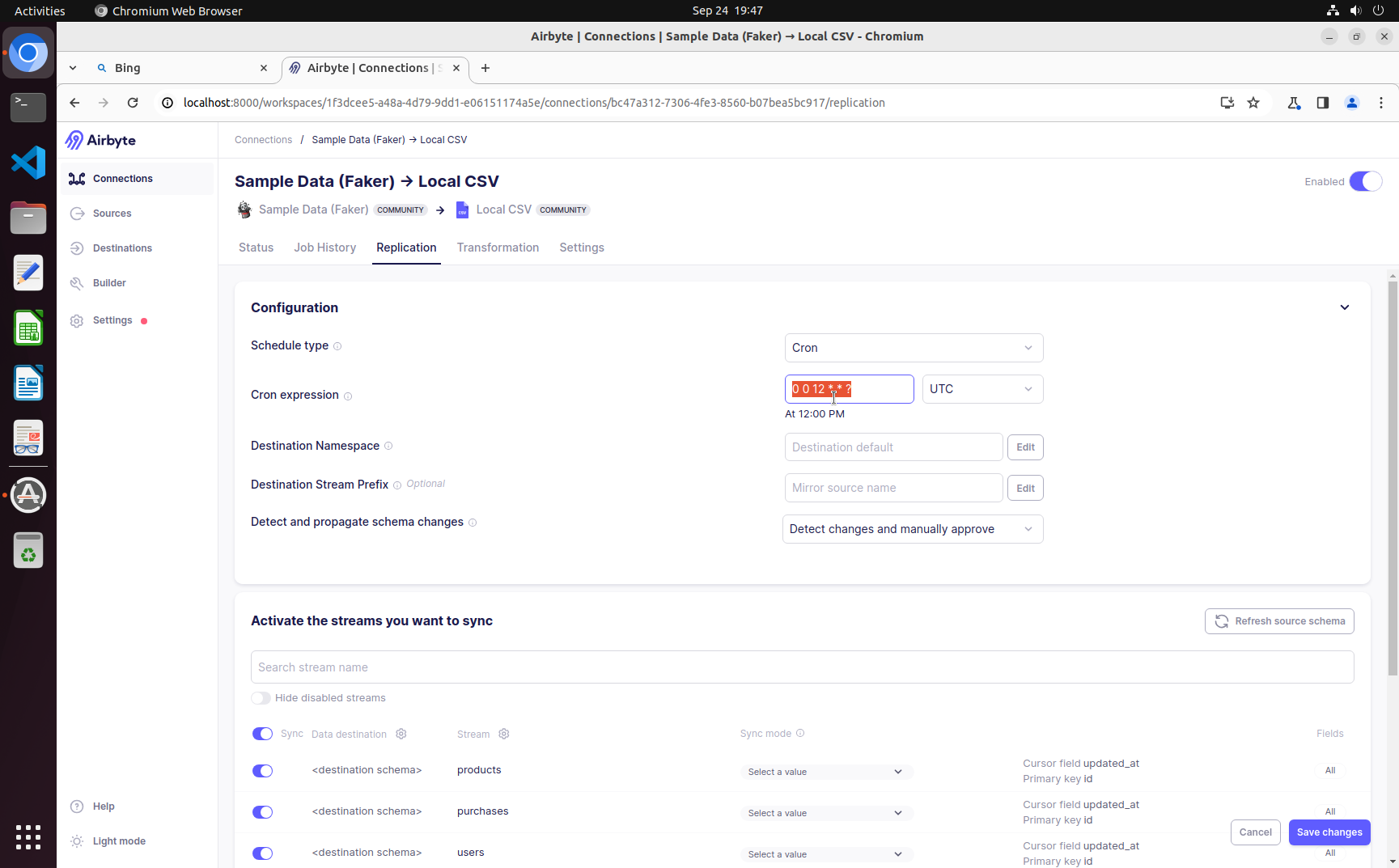} \\
\midrule
\bottomrule
\end{tabular}

\label{tab:spider2v_4}
\end{table}

\begin{table}[ht]
\caption{\textbf{Spider2-V example, cont.}}
\centering
\begin{tabular}{p{13cm}}
\toprule
\textbf{Action 8} \\
\midrule
import pyautogui \\
pyautogui.press(``delete") // delete the default content  \\
\midrule
\textbf{Observation 8 (Interface of the software Airbyte)} \\
\midrule
\includegraphics[width=0.8\textwidth]{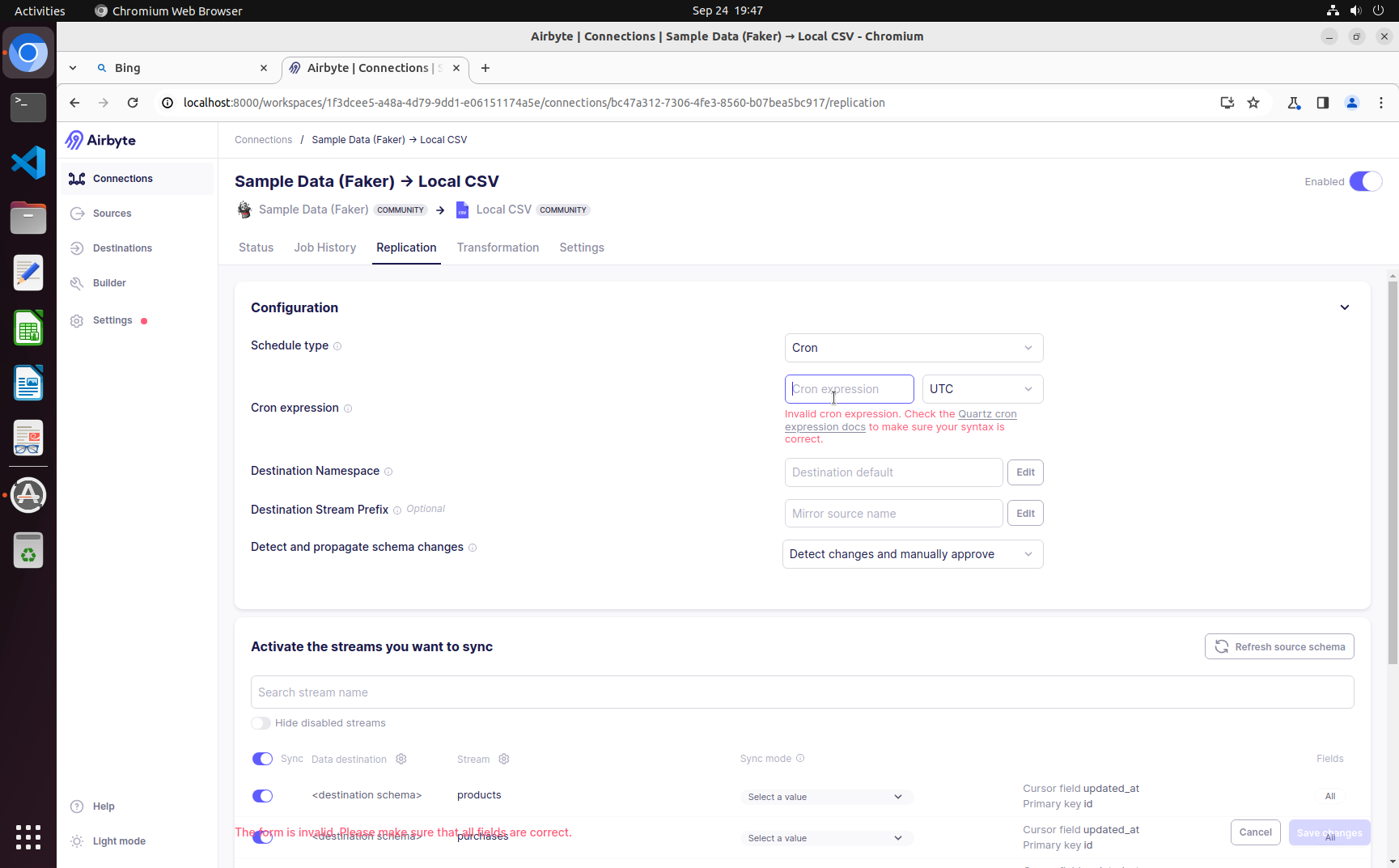} \\
\midrule
\textbf{Action 9} \\
\midrule
import pyautogui \\
pyautogui.write(``0 0 18 * * ?") // update the content to align 18:00 pm in the instruction \\
\midrule
\textbf{Observation 9 (Interface of the software Airbyte)} \\
\midrule
\includegraphics[width=0.8\textwidth]{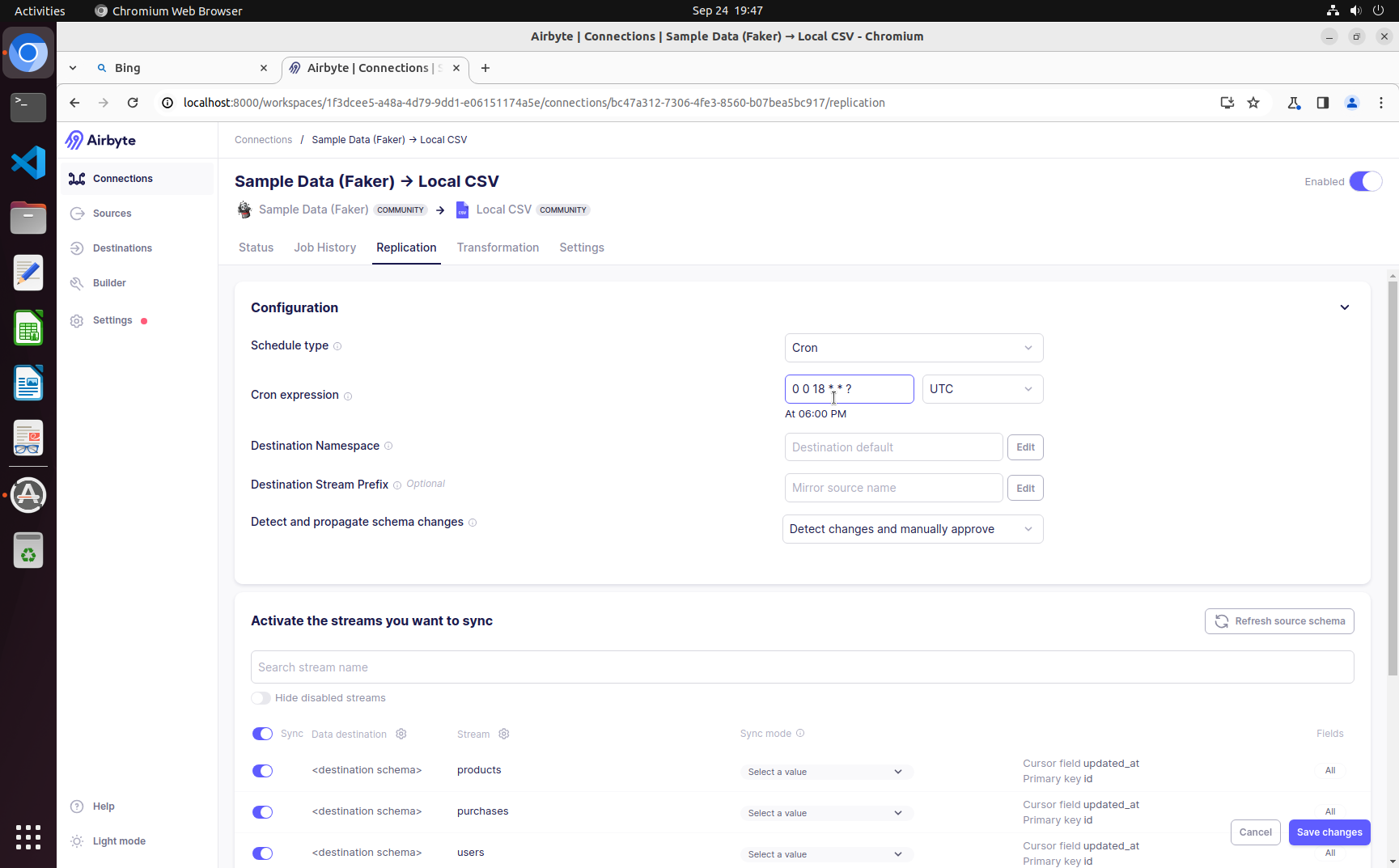} \\
\midrule
\bottomrule
\end{tabular}

\label{tab:spider2v_5}
\end{table}

\begin{table}[ht]
\caption{\textbf{Spider2-V example, cont.}}
\centering
\begin{tabular}{p{13cm}}
\toprule
\textbf{Action 10} \\
\midrule
import pyautogui \\
pyautogui.click(1450,900) // click the button save changes  \\
\midrule
\textbf{Observation 10 (Interface of the software Airbyte)} \\
\midrule
\includegraphics[width=0.8\textwidth]{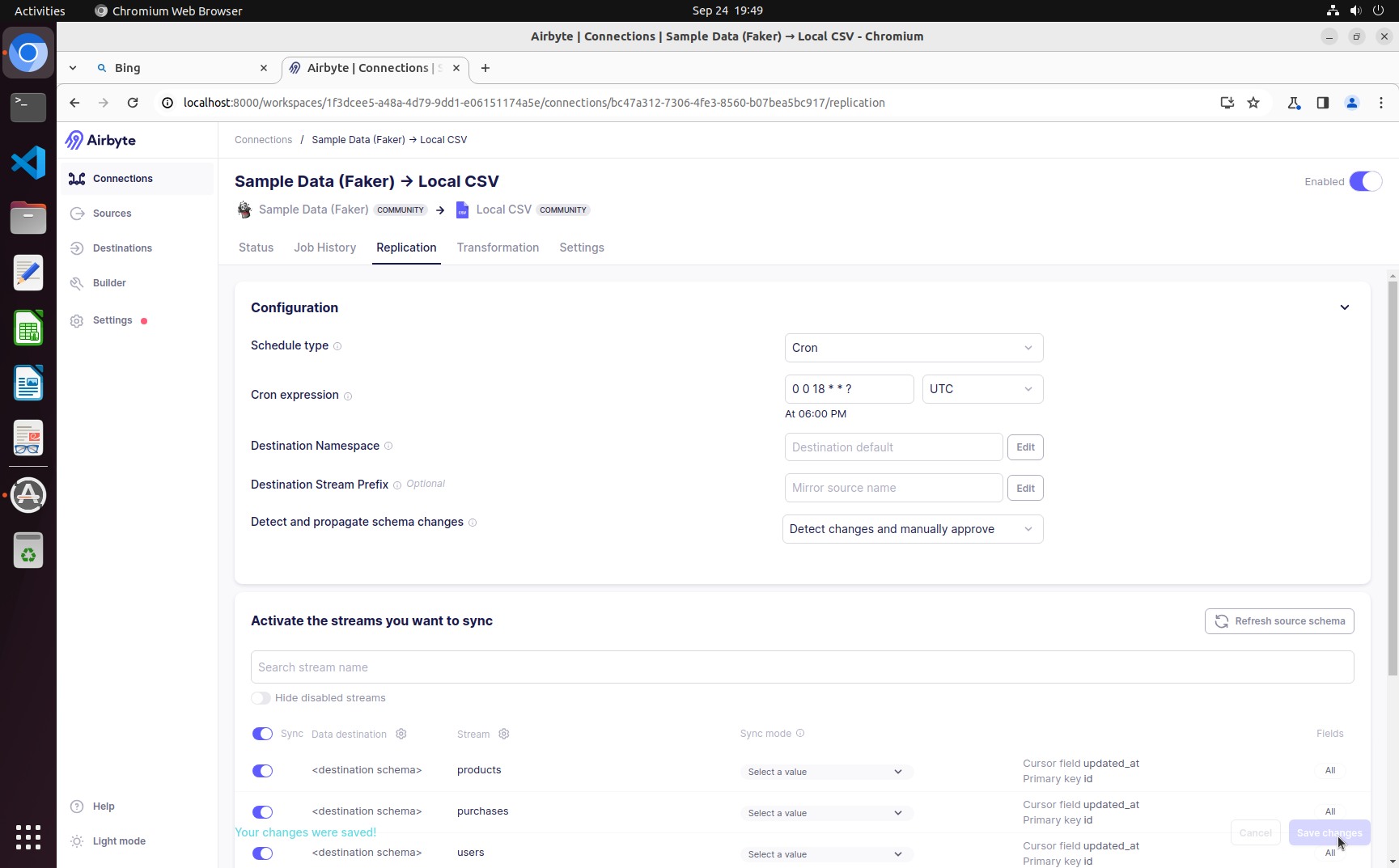} \\
\bottomrule
\end{tabular}

\label{tab:spider2v_6}
\end{table}

%% file: tables/observation_space_spider2v.tex
\begin{table}[ht]
\caption{Observation space of Spider2-V.}
\centering
\begin{tabular}{p{13cm}}
\toprule
\textbf{Screenshot} \\
\midrule
\includegraphics[width=0.8\textwidth]{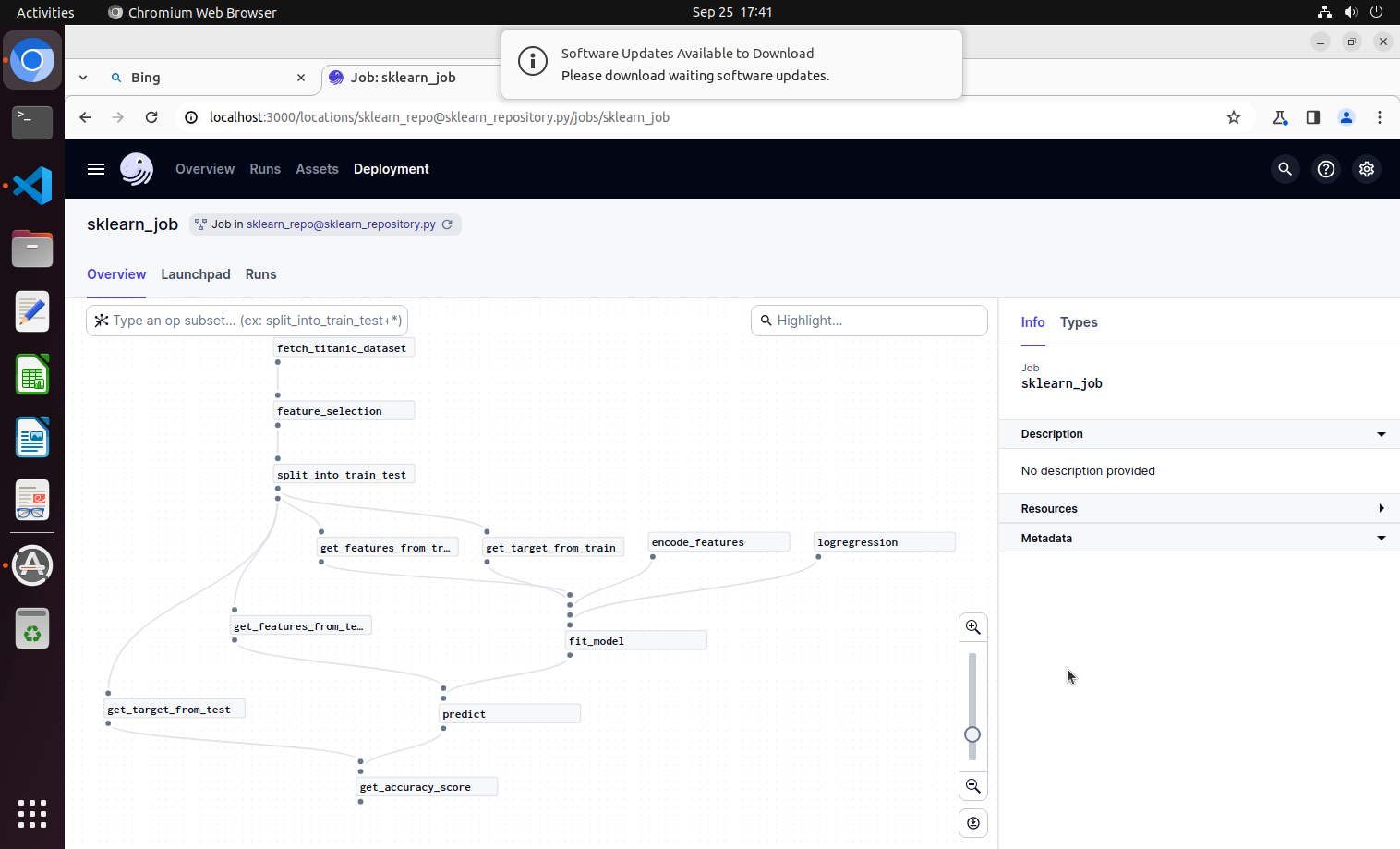} \\
\midrule
\textbf{Set-of-mark} \\
\midrule
\includegraphics[width=0.8\textwidth]{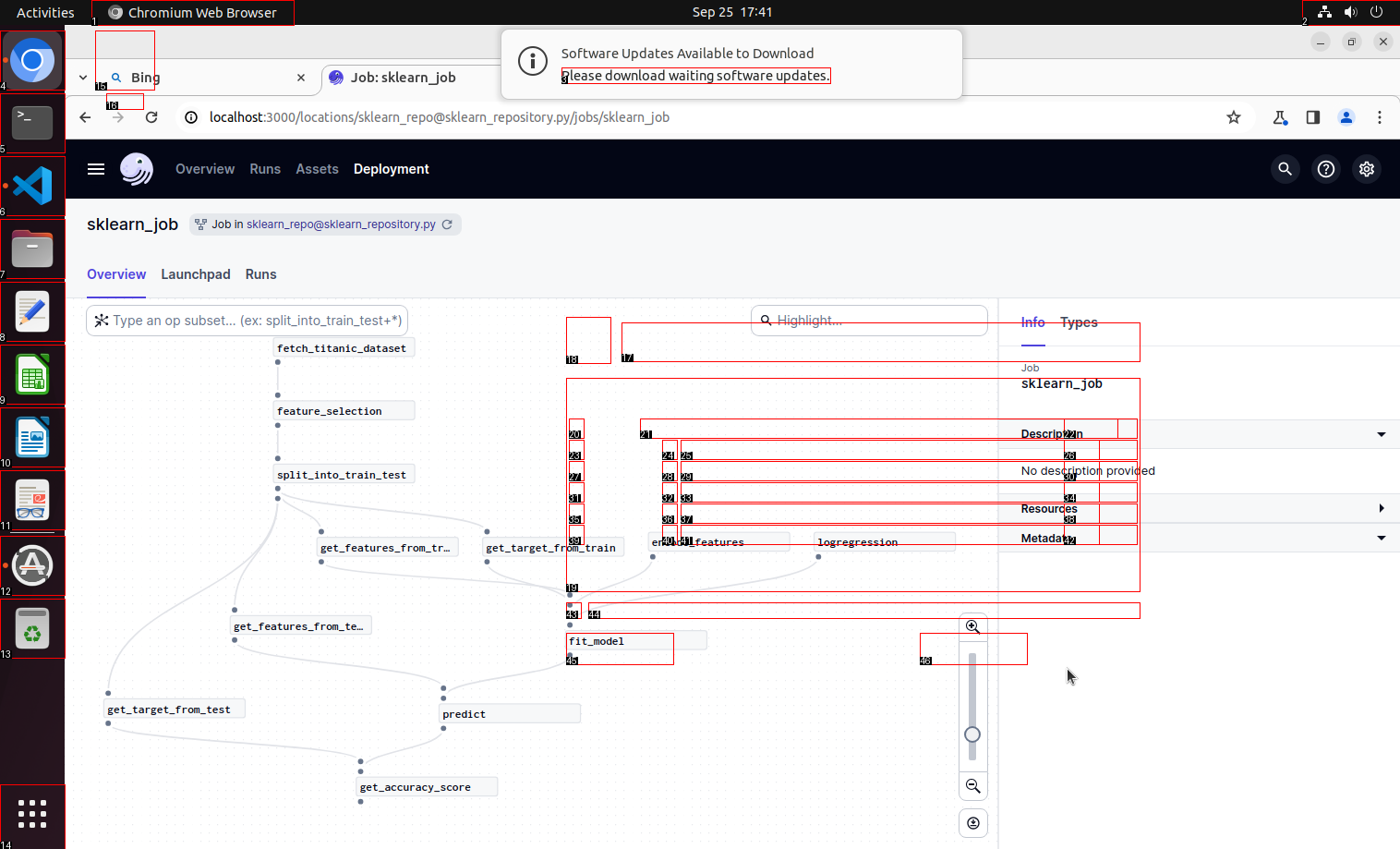} \\
\bottomrule
\end{tabular}
\label{tab:observation_space_spider2v_1}
\end{table}

\begin{table}[ht]
\caption{Observation space of Spider2-V. The accessibility tree suffers from significant information loss. Compared to the screenshot and set-of-mark shown in Table~\ref{tab:observation_space_spider2v_1}, the presented accessibility tree fails to retrieve webpage information, and only shows the details of the desktop icons in the left panel.}
\centering
\begin{tabular}{p{13cm}}
\toprule
$[$208, 13$]$ menu Chromium Web Browser ``" \\
$[$1463, 13$]$ menu System ``"\\
$[$35, 65$]$ push-button Chromium Web Browser ``" \\
$[$753, 81$]$ label Please download waiting software updates. ``" \\
$[$135, 109$]$ label Home \\
$[$35, 133$]$ push-button Terminal ``" \\
$[$35, 201$]$ push-button Visual Studio Code ``" \\
$[$35, 269$]$ push-button Files ``" \\
$[$35, 337$]$ push-button Text Editor ``" \\
$[$953, 370$]$ label Updated software is available for this computer. Do you want to install it now? \\
$[$35, 405$]$ push-button LibreOffice Calc ``" \\
$[$951, 463$]$ table-cell Security updates \\
$[$1191, 463$]$ table-cell 638.8 MB \\
$[$35, 473$]$ push-button LibreOffice Writer ``" \\
$[$963, 486$]$ table-cell LibreOffice \\
$[$1191, 486$]$ table-cell 23.4 MB \\
$[$963, 509$]$ table-cell LibreOffice Calc \\
$[$1191, 509$]$ table-cell 8.7 MB \\
$[$923, 524$]$ toggle-button Details of updates ``" \\
$[$963, 532$]$ table-cell LibreOffice Draw \\
$[$1191, 532$]$ table-cell 3.0 MB \\
$[$35, 541$]$ push-button Document Viewer ``" \\
$[$963, 555$]$ table-cell LibreOffice Impress \\
$[$1191, 555$]$ table-cell 1.3 MB \\
$[$963, 578$]$ table-cell LibreOffice Math \\
$[$1191, 578$]$ table-cell 673 kB \\
$[$35, 612$]$ push-button Software Updater ``" \\
$[$935, 660$]$ label 1157.8 MB will be downloaded. \\
$[$35, 680$]$ push-button Trash ``" \\
$[$671, 702$]$ push-button Settings… ``" \\
$[$1054, 702$]$ push-button Cancel ``" \\
$[$1176, 702$]$ push-button Install Now ``" \\
$[$35, 884$]$ toggle-button Show Applications ``" \\
\bottomrule
\end{tabular}
\label{tab:observation_space_spider2v_2}
\end{table}

%% file: tables/data_samples/data_synthesis_bigquery.tex
\begin{table}[ht]
\caption{\textbf{Example of data synthesis - Bigquery}}
\centering
\begin{tabular}{p{13cm}}
\toprule
\textbf{Instruction} \\
\midrule
Upload CSV data in Google Drive to BigQuery. \\
\midrule
\textbf{Observation 0 (Bigquery Interface)} \\
\midrule
\includegraphics[width=0.8\textwidth]{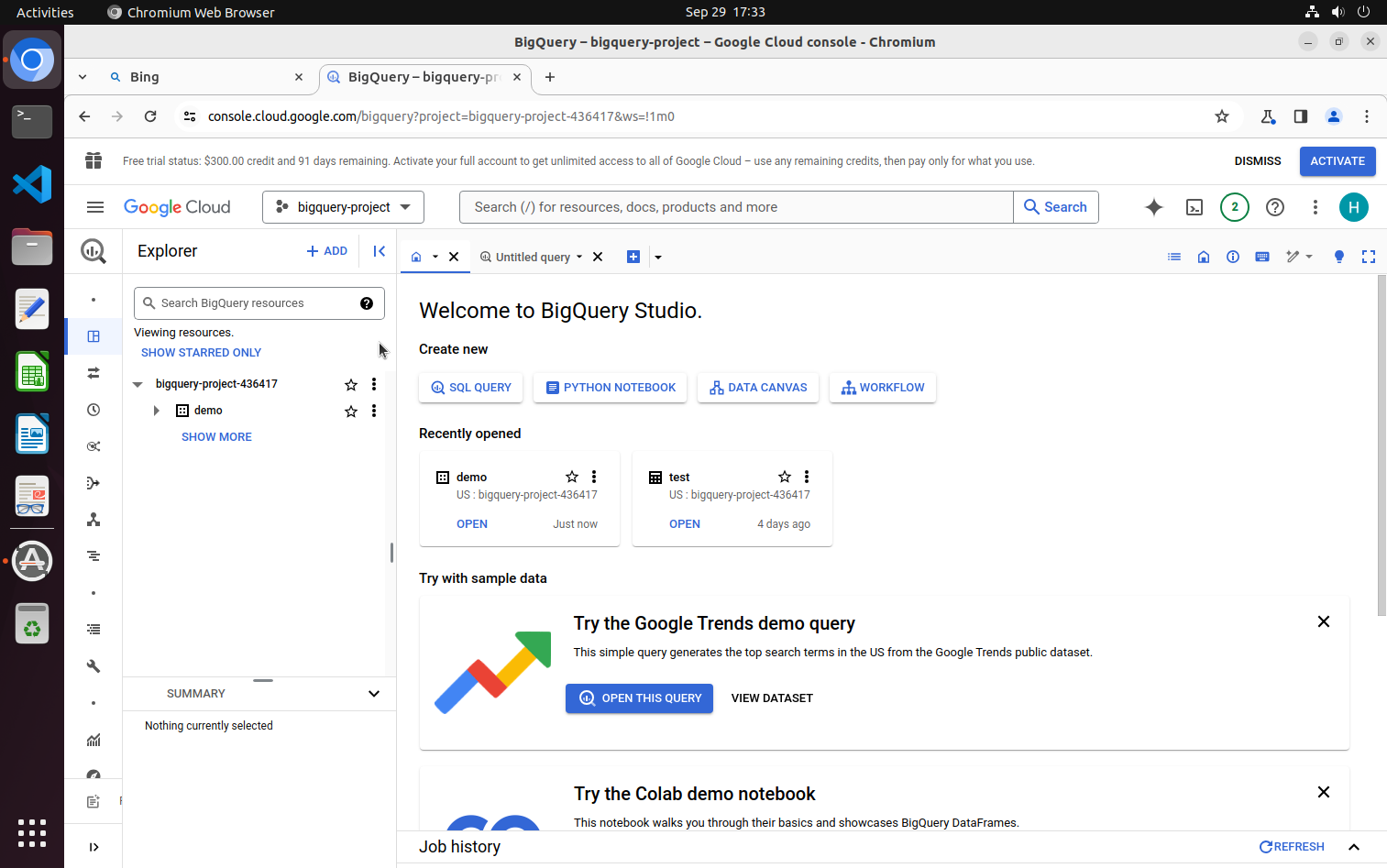} \\
\midrule
\textbf{Action 1} \\
\midrule
import pyautogui \\
pyautogui.doubleClick(332,447) // double click the dataset demo. \\
\midrule
\textbf{Observation 1 (Bigquery Interface)} \\
\midrule
\includegraphics[width=0.8\textwidth]{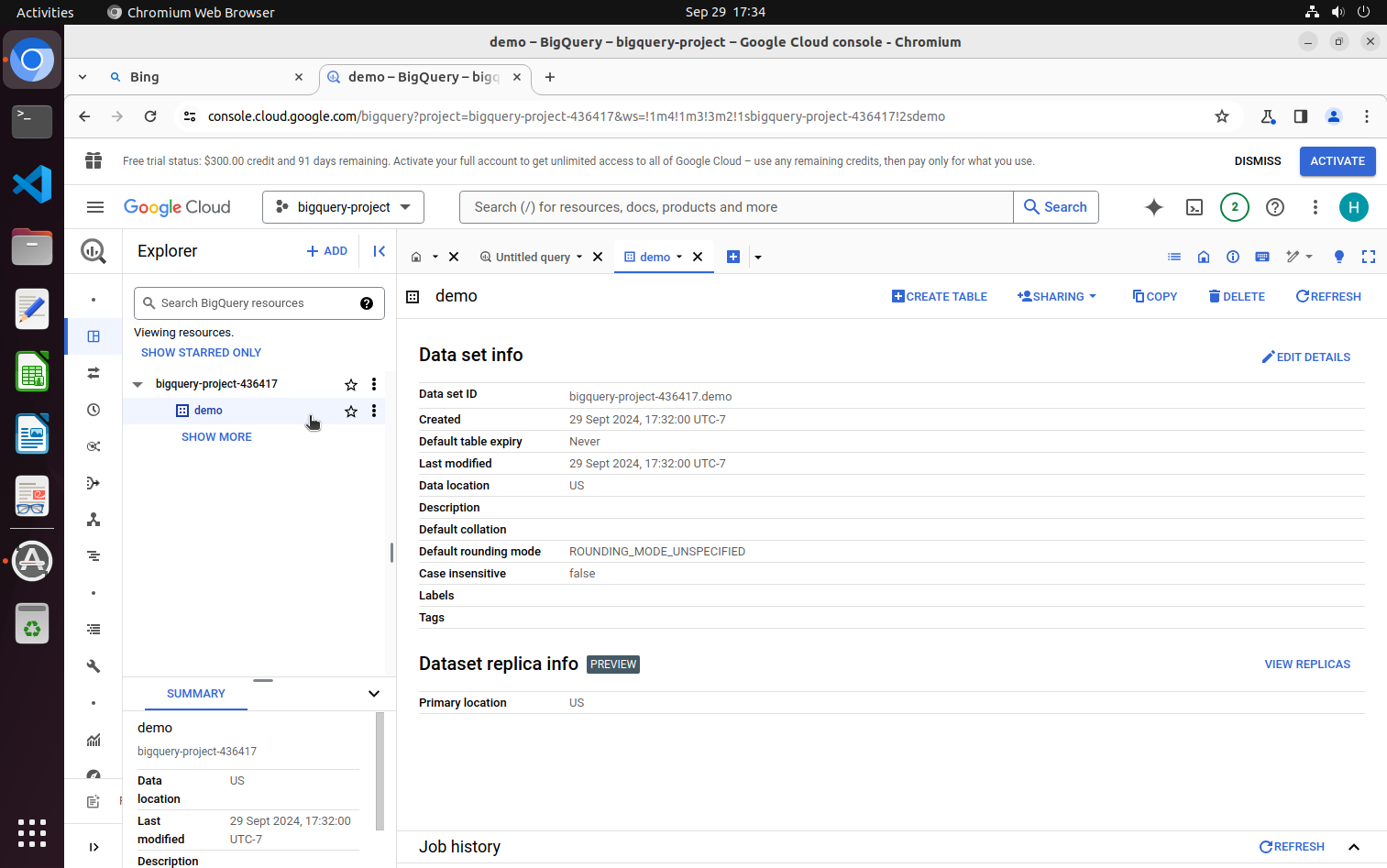} \\
\midrule
\bottomrule
\end{tabular}
\label{tab:data_synthesis_bigquery_1}
\end{table}

\begin{table}[ht]
\caption{\textbf{Example of data synthesis - Bigquery}}
\centering
\begin{tabular}{p{13cm}}
\toprule
\textbf{Action 2} \\
\midrule
import pyautogui \\
pyautogui.doubleClick(1002,321) // double click the button CREATE TABLE. \\
\midrule
\textbf{Observation 2 (Bigquery Interface)} \\
\midrule
\includegraphics[width=0.8\textwidth]{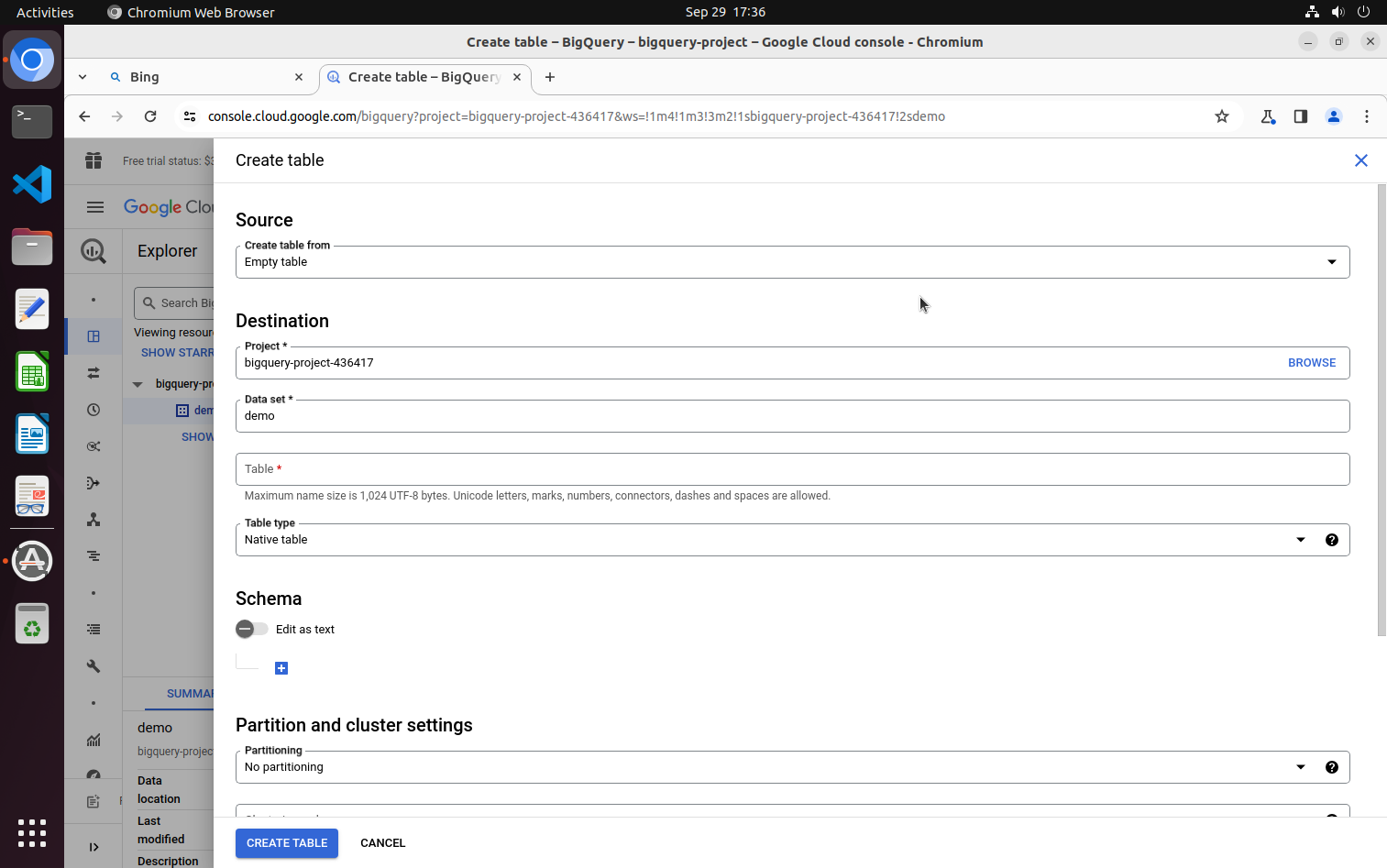} \\
\midrule
\textbf{Action 3} \\
\midrule
import pyautogui \\
pyautogui.click(1458,279) // click the dropdown menu to select table source. \\
\midrule
\textbf{Observation 3 (Bigquery Interface)} \\
\midrule
\includegraphics[width=0.8\textwidth]{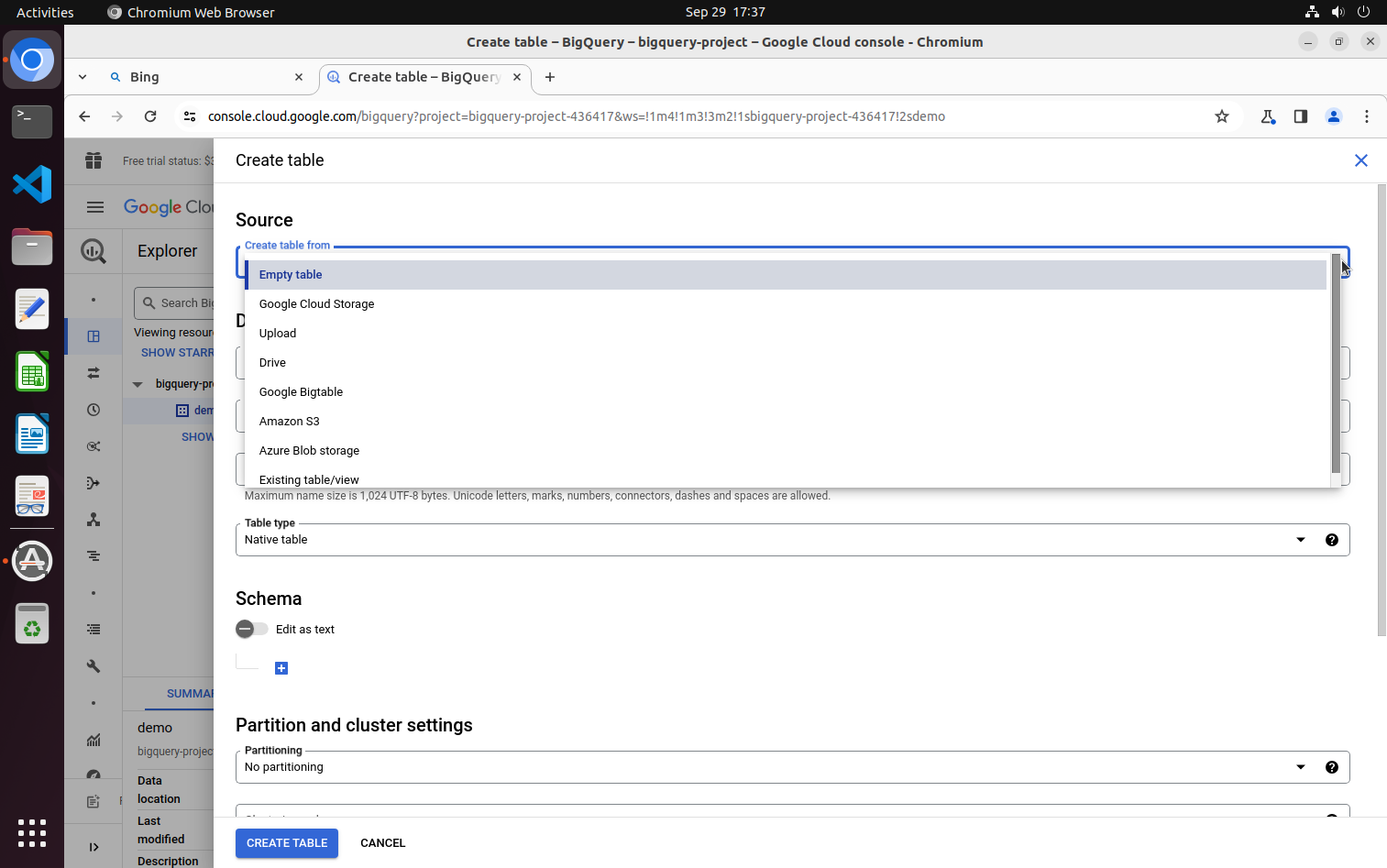} \\
\midrule
\bottomrule
\end{tabular}
\label{tab:data_synthesis_bigquery_2}
\end{table}

\begin{table}[ht]
\caption{\textbf{Example of data synthesis - Bigquery}}
\centering
\begin{tabular}{p{13cm}}
\toprule
\textbf{Action 4} \\
\midrule
import pyautogui \\
// wrong prediction: click the table source Google Cloud Storage.  \\
// The correct prediction should select Drive to align with the instruction. \\
pyautogui.click(1302,331)  \\
\midrule
\textbf{Observation 4 (Bigquery Interface)} \\
\midrule
\includegraphics[width=0.8\textwidth]{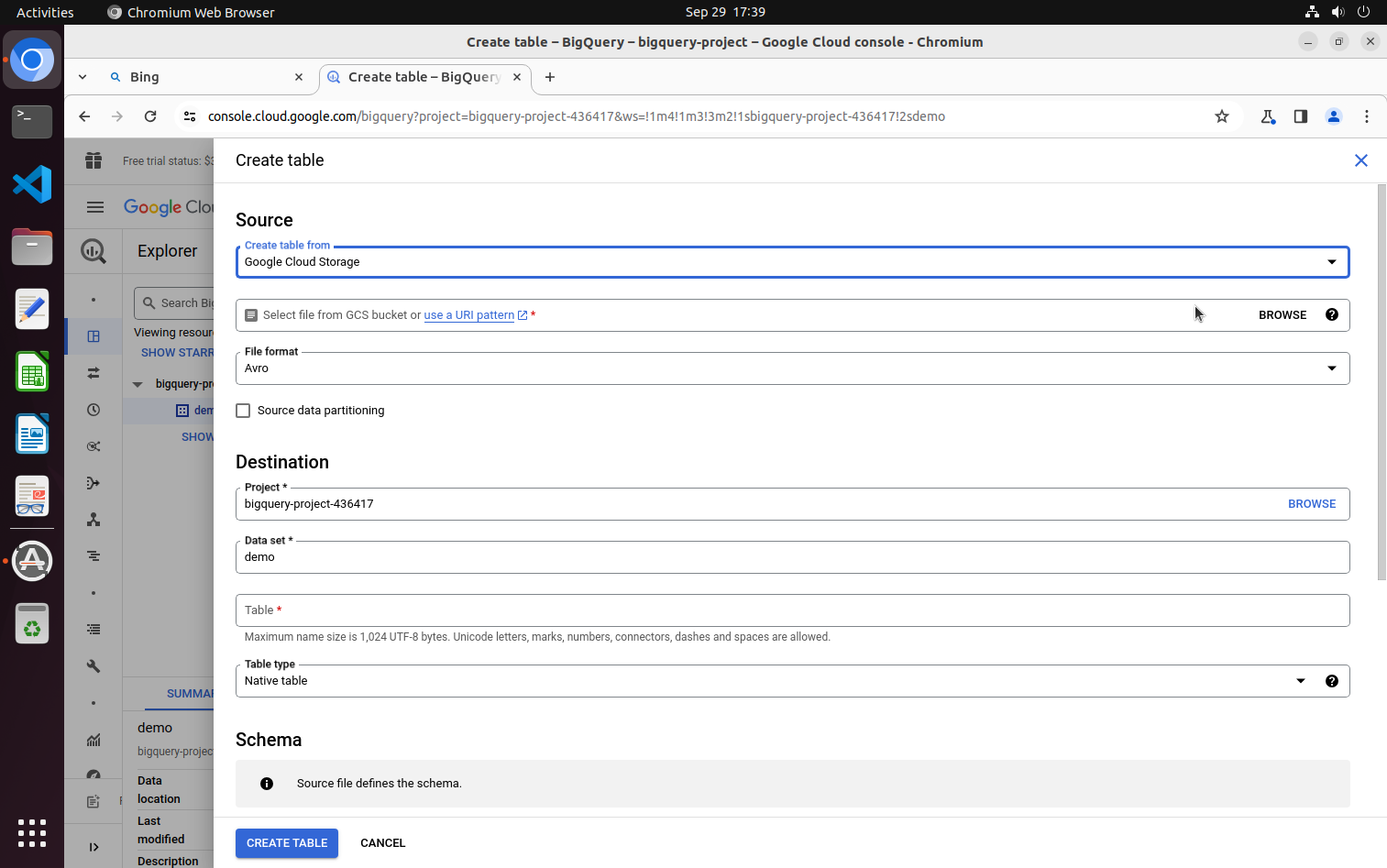} \\
\midrule
\textbf{Action 5} \\
\midrule
import pyautogui \\
pyautogui.click(1389,340) // click BROWSE to find file. \\
\midrule
\textbf{Observation 5 (Bigquery Interface)} \\
\midrule
\includegraphics[width=0.8\textwidth]{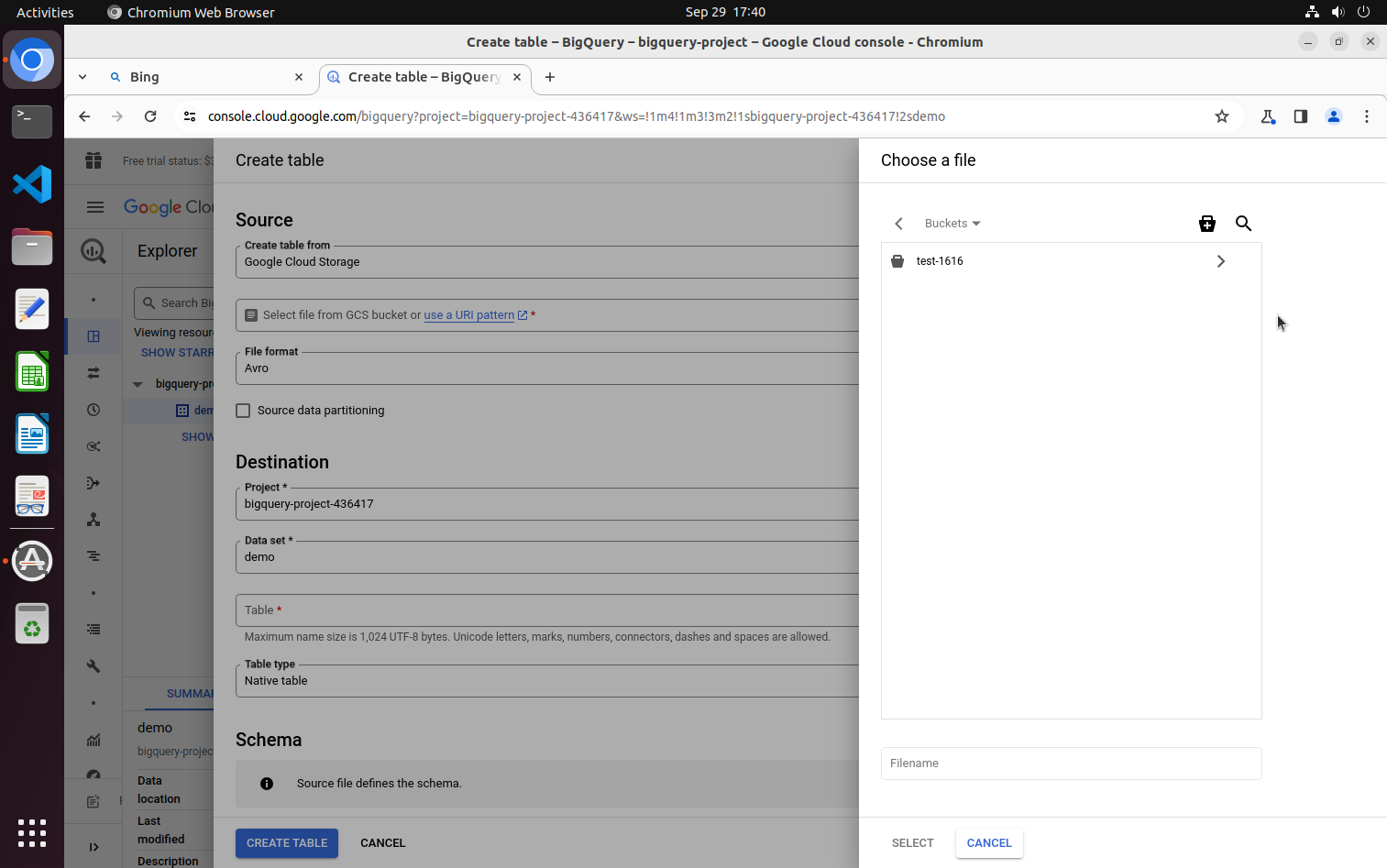} \\
\midrule
\bottomrule
\end{tabular}
\label{tab:data_synthesis_bigquery_3}
\end{table}

\begin{table}[ht]
\caption{\textbf{Example of data synthesis - Bigquery}}
\centering
\begin{tabular}{p{13cm}}
\toprule
\textbf{Action 6} \\
\midrule
import pyautogui \\
pyautogui.click(1341,282) // click to find files under directory. \\
\midrule
\textbf{Observation 6 (Bigquery Interface)} \\
\midrule
\includegraphics[width=0.8\textwidth]{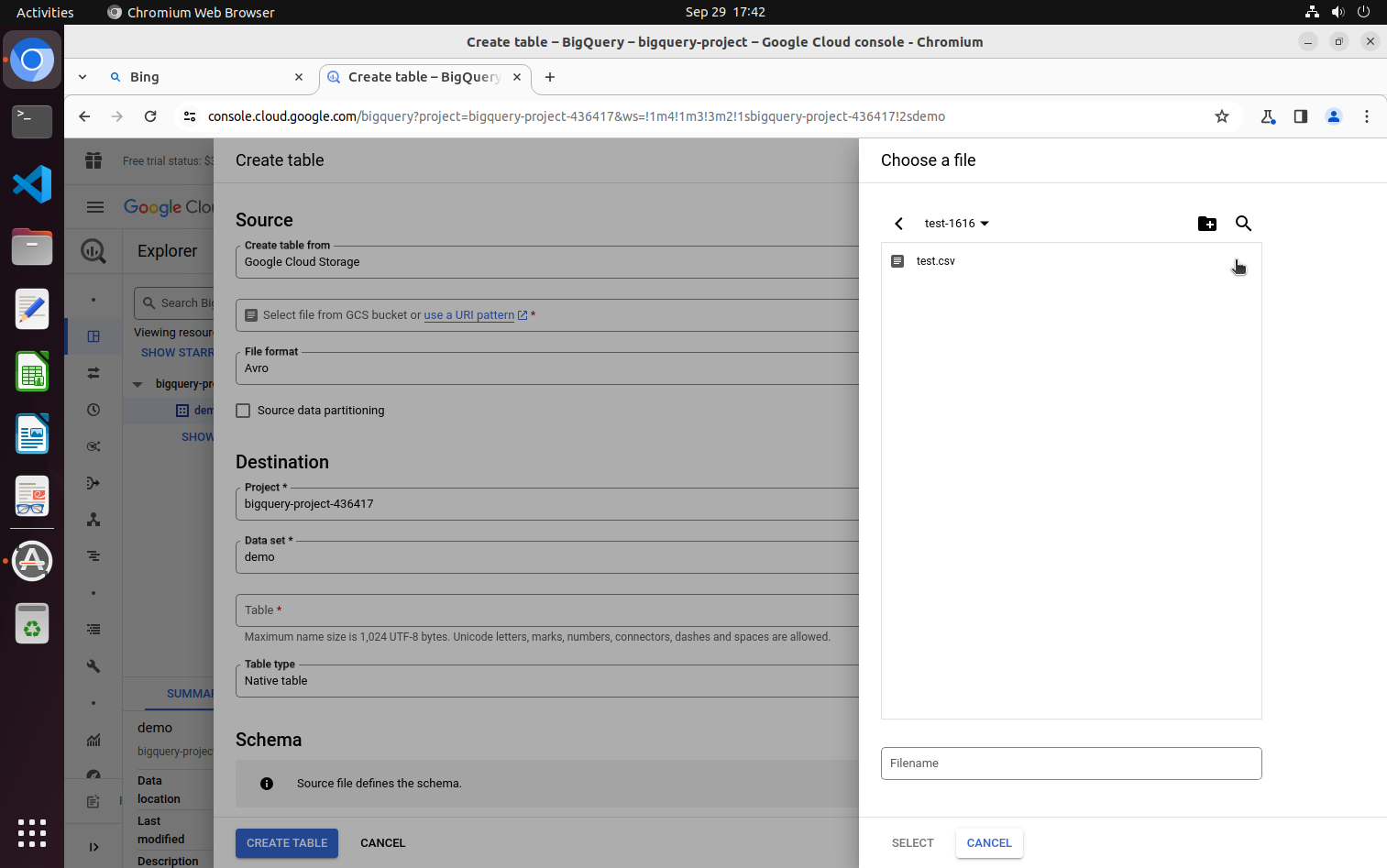} \\
\midrule
\textbf{Action 7} \\
\midrule
import pyautogui \\
pyautogui.doubleClick(1203,284) // select the CSV file. \\
\midrule
\textbf{Observation 7 (Bigquery Interface)} \\
\midrule
\includegraphics[width=0.8\textwidth]{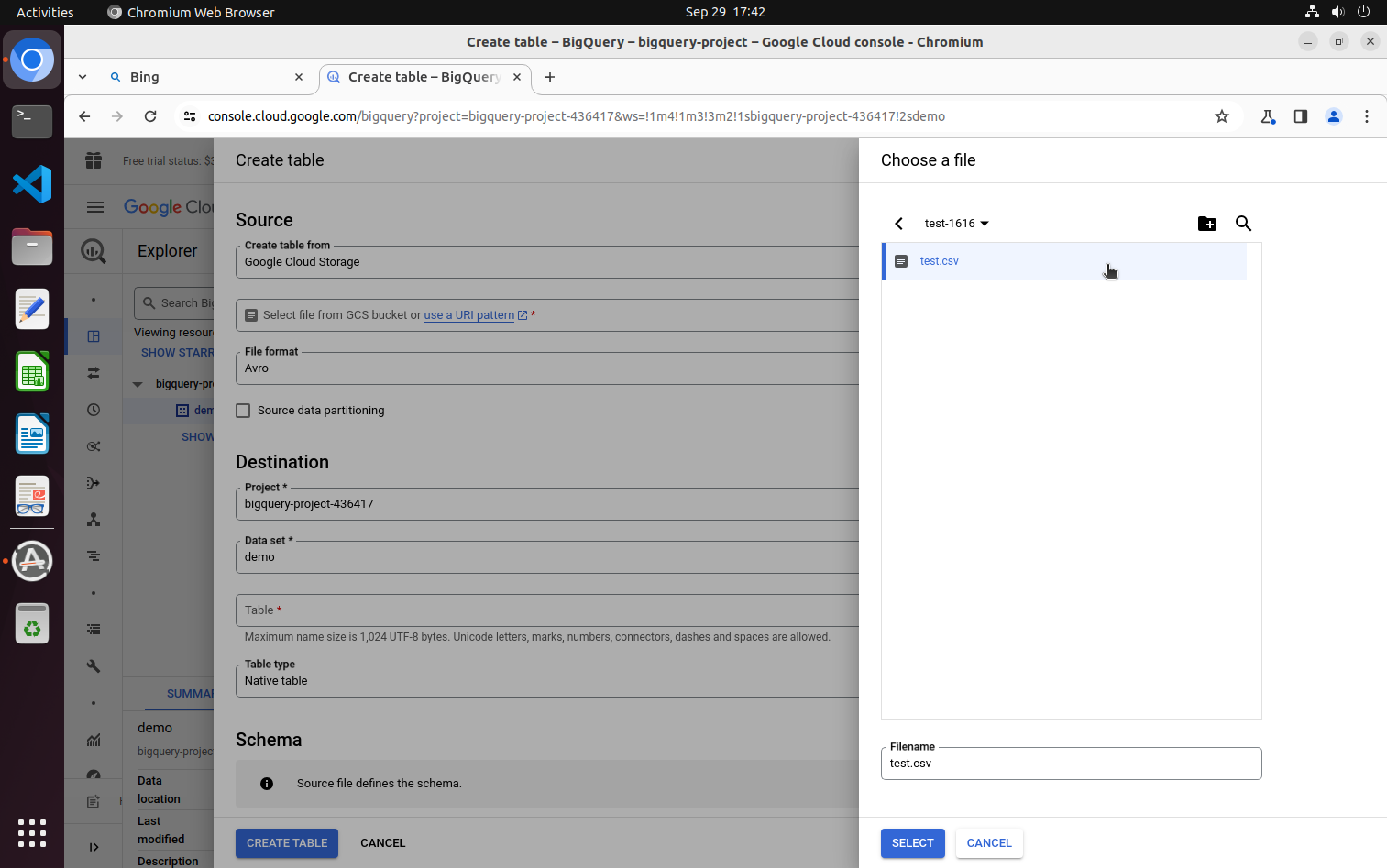} \\
\midrule
\bottomrule
\end{tabular}
\label{tab:data_synthesis_bigquery_4}
\end{table}

\begin{table}[ht]
\caption{\textbf{Example of data synthesis - Bigquery}}
\centering
\begin{tabular}{p{13cm}}
\toprule
\textbf{Action 8} \\
\midrule
import pyautogui \\
pyautogui.click(1000,915) // click to select file. \\
\midrule
\textbf{Observation 8 (Bigquery Interface)} \\
\midrule
\includegraphics[width=0.8\textwidth]{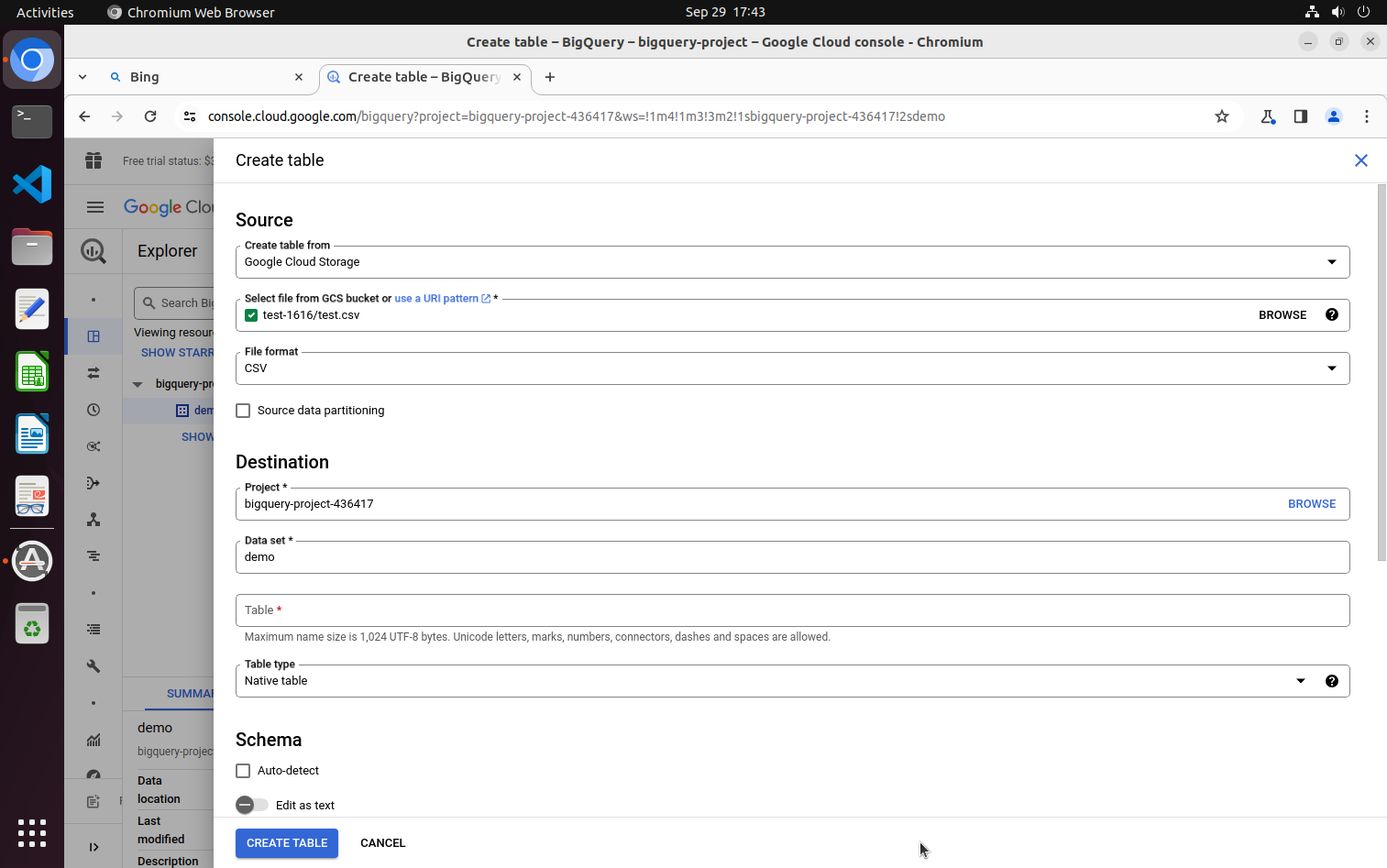} \\
\midrule
\textbf{Action 9} \\
\midrule
import pyautogui \\
pyautogui.click(1000,668) // click to focus on the text box Table. \\
\midrule
\textbf{Observation 9 (Bigquery Interface)} \\
\midrule
\includegraphics[width=0.8\textwidth]{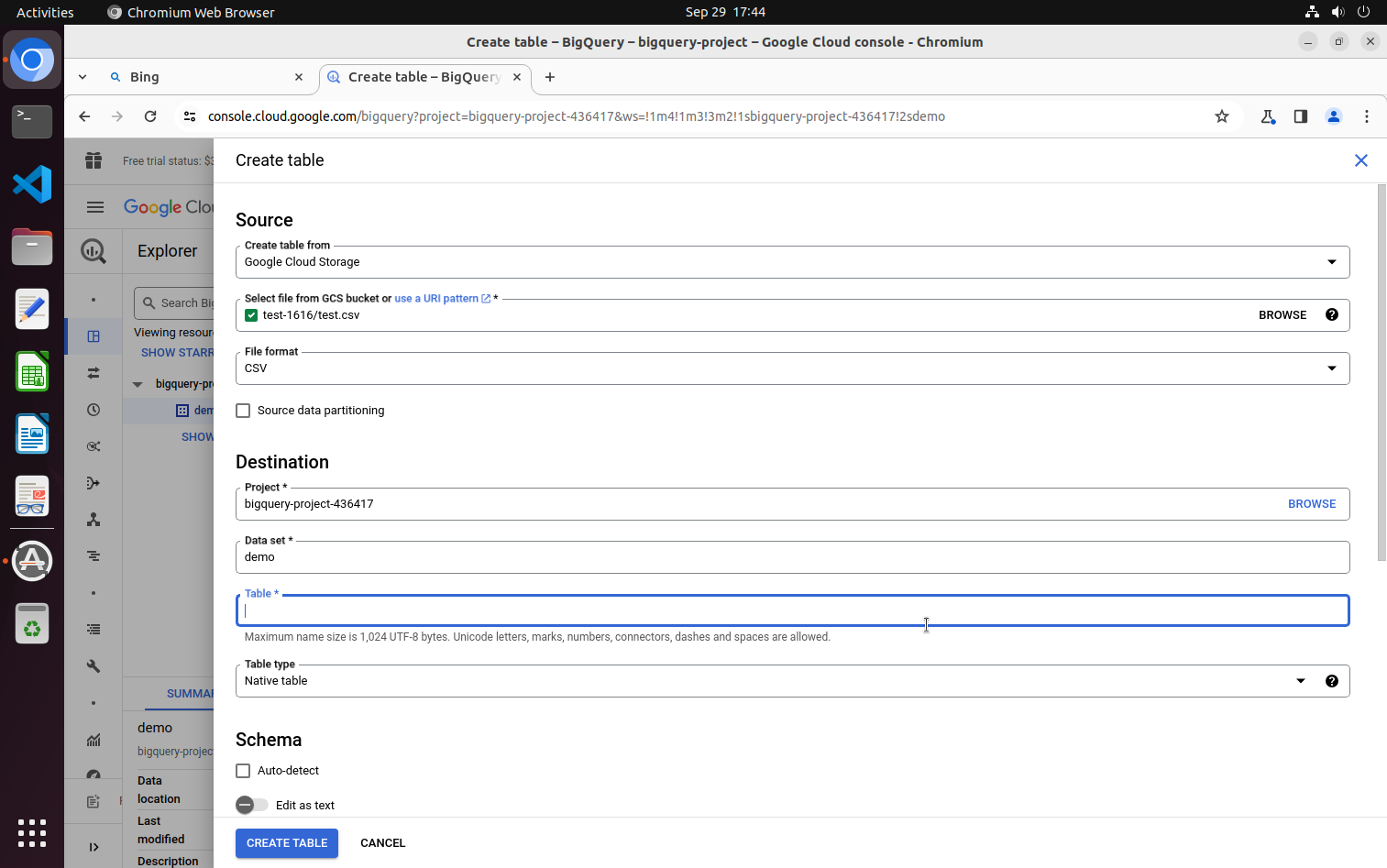} \\
\midrule
\bottomrule
\end{tabular}
\label{tab:data_synthesis_bigquery_5}
\end{table}

\begin{table}[ht]
\caption{\textbf{Example of data synthesis - Bigquery}}
\centering
\begin{tabular}{p{13cm}}
\toprule
\textbf{Action 10} \\
\midrule
import pyautogui \\
pyautogui.write("test") // name the file "test". \\
\midrule
\textbf{Observation 10 (Bigquery Interface)} \\
\midrule
\includegraphics[width=0.8\textwidth]{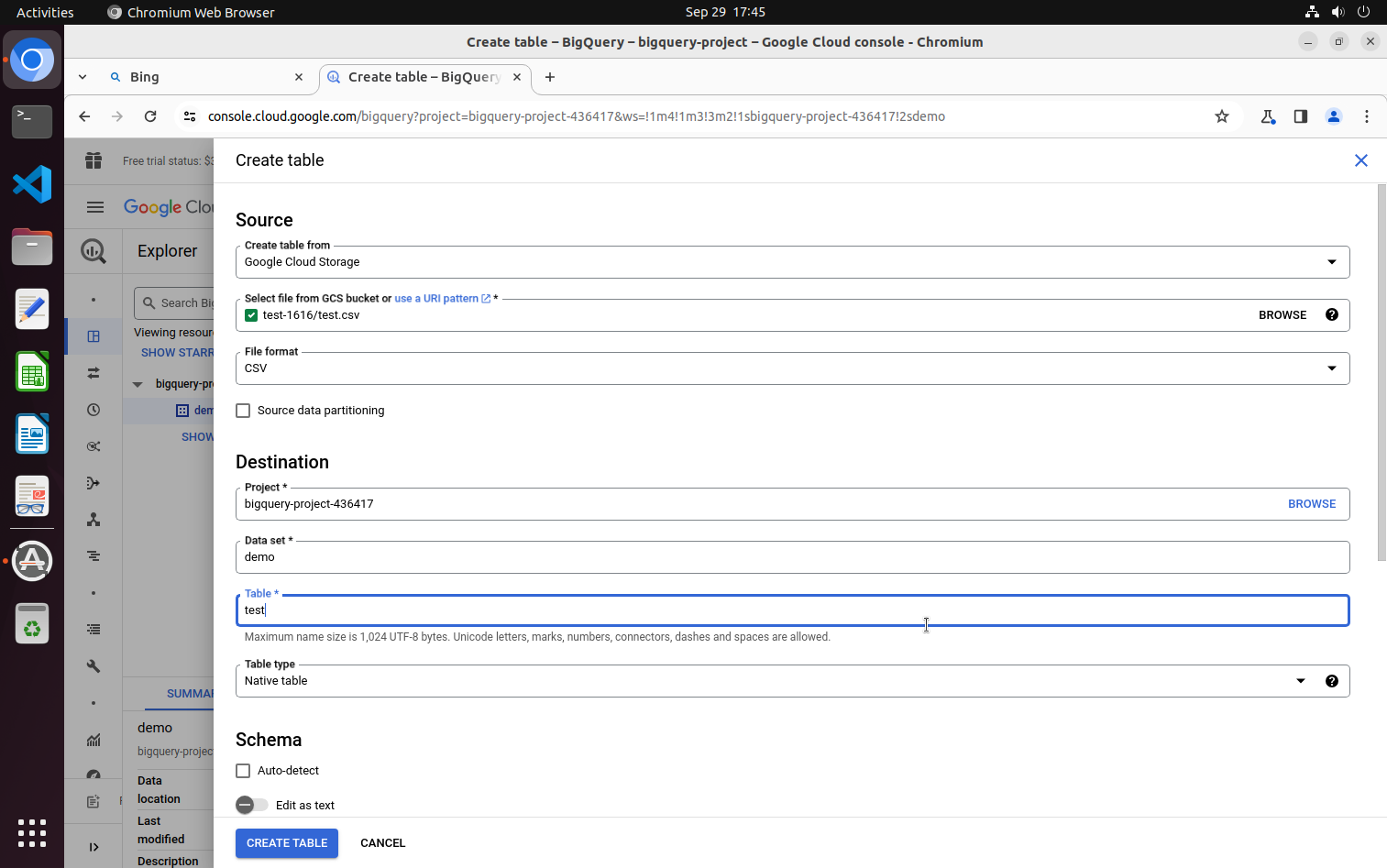} \\
\midrule
\textbf{Action 11} \\
\midrule
import pyautogui \\
pyautogui.click(1425,726) // click to reveal dropdown menu of the table type. \\
\midrule
\textbf{Observation 11 (Bigquery Interface)} \\
\midrule
\includegraphics[width=0.8\textwidth]{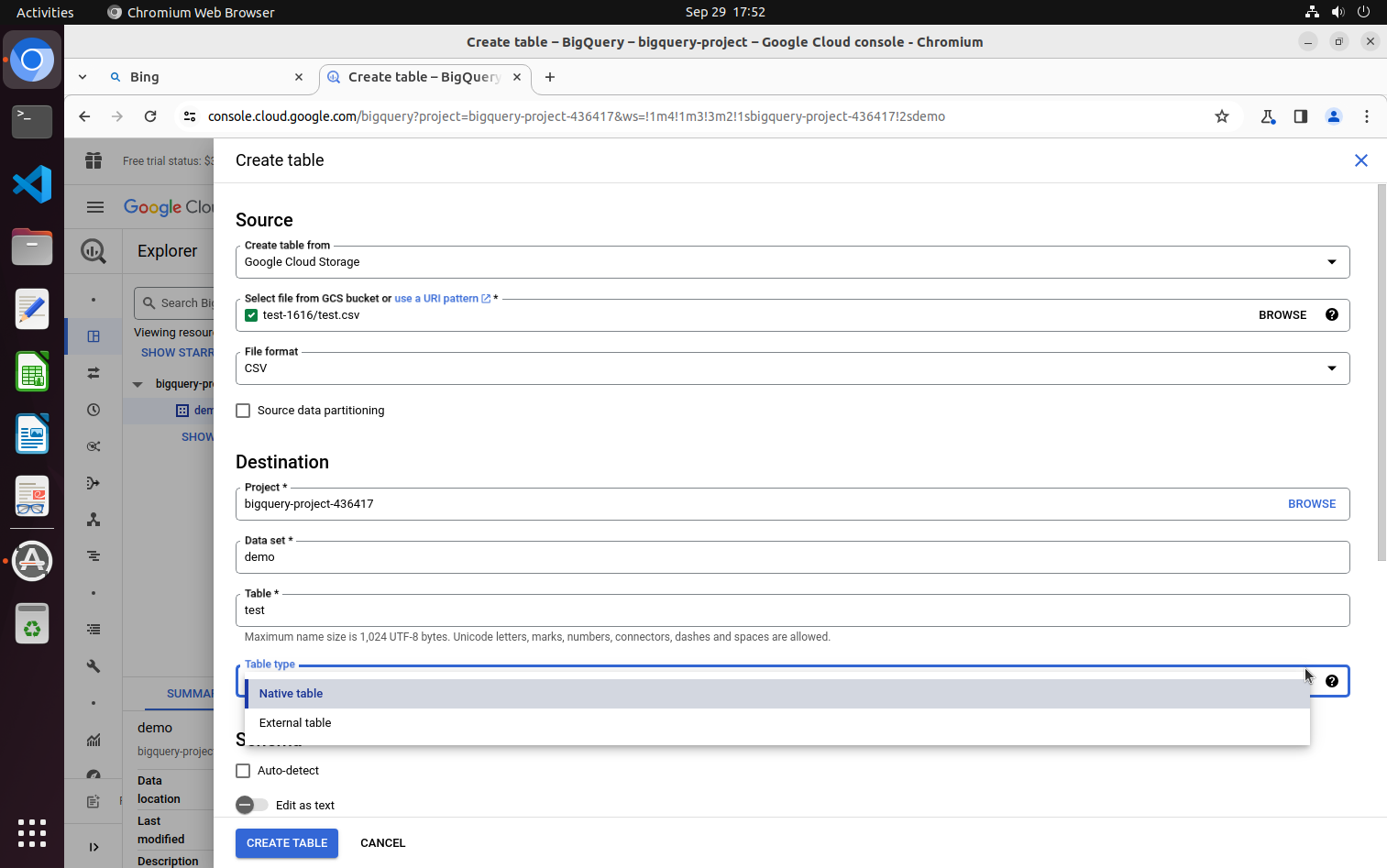} \\
\midrule
\bottomrule
\end{tabular}
\label{tab:data_synthesis_bigquery_6}
\end{table}

\begin{table}[ht]
\caption{\textbf{Example of data synthesis - Bigquery}}
\centering
\begin{tabular}{p{13cm}}
\toprule
\textbf{Action 12} \\
\midrule
import pyautogui \\
pyautogui.click(1297,801) // select the table source external table. \\
\midrule
\textbf{Observation 12 (Bigquery Interface)} \\
\midrule
\includegraphics[width=0.8\textwidth]{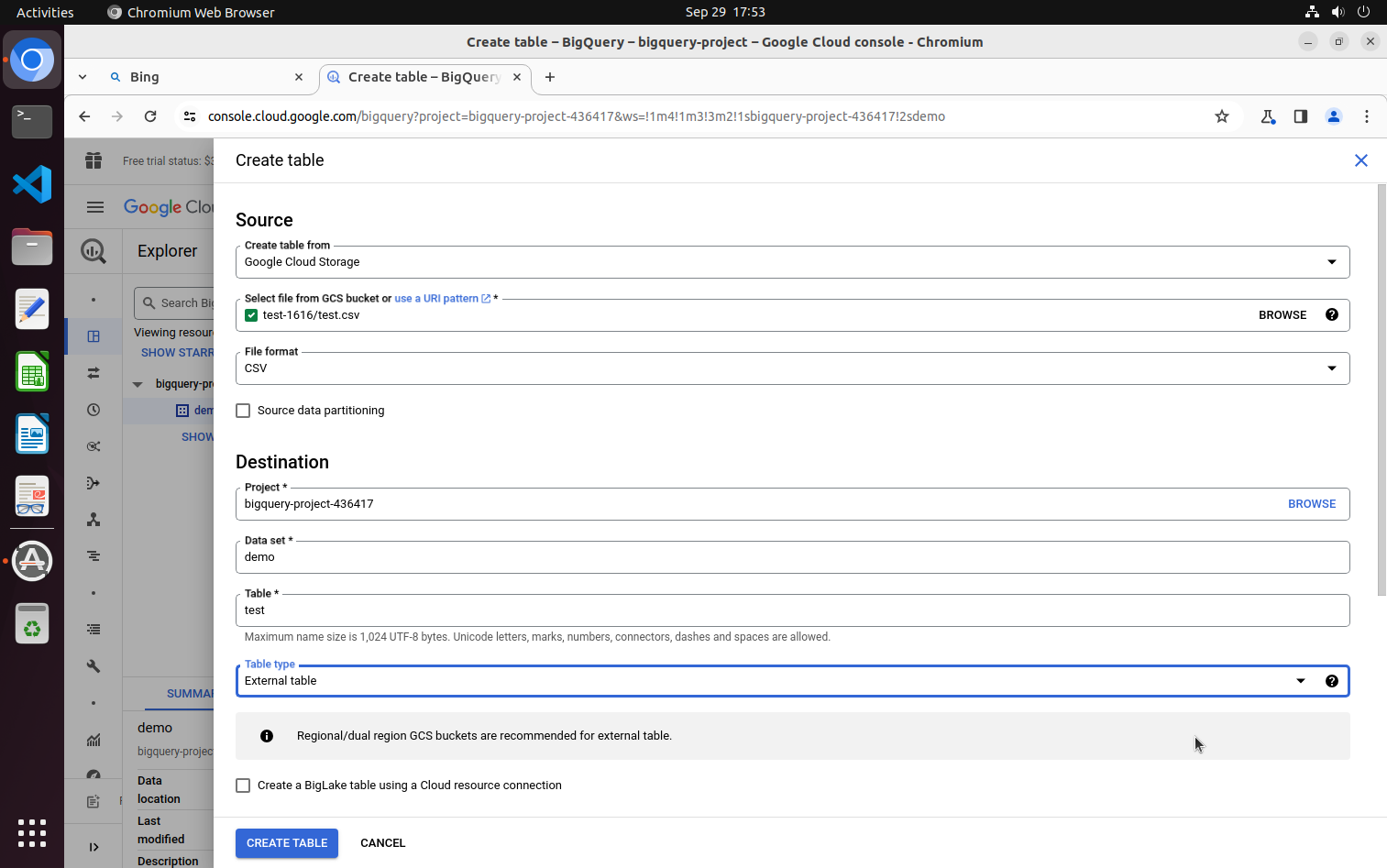} \\
\midrule
\textbf{Action 13} \\
\midrule
import pyautogui \\
pyautogui.click(303,914) // click CREATE TABLE. \\
\midrule
\textbf{Observation 13 (Bigquery Interface)} \\
\midrule
\includegraphics[width=0.8\textwidth]{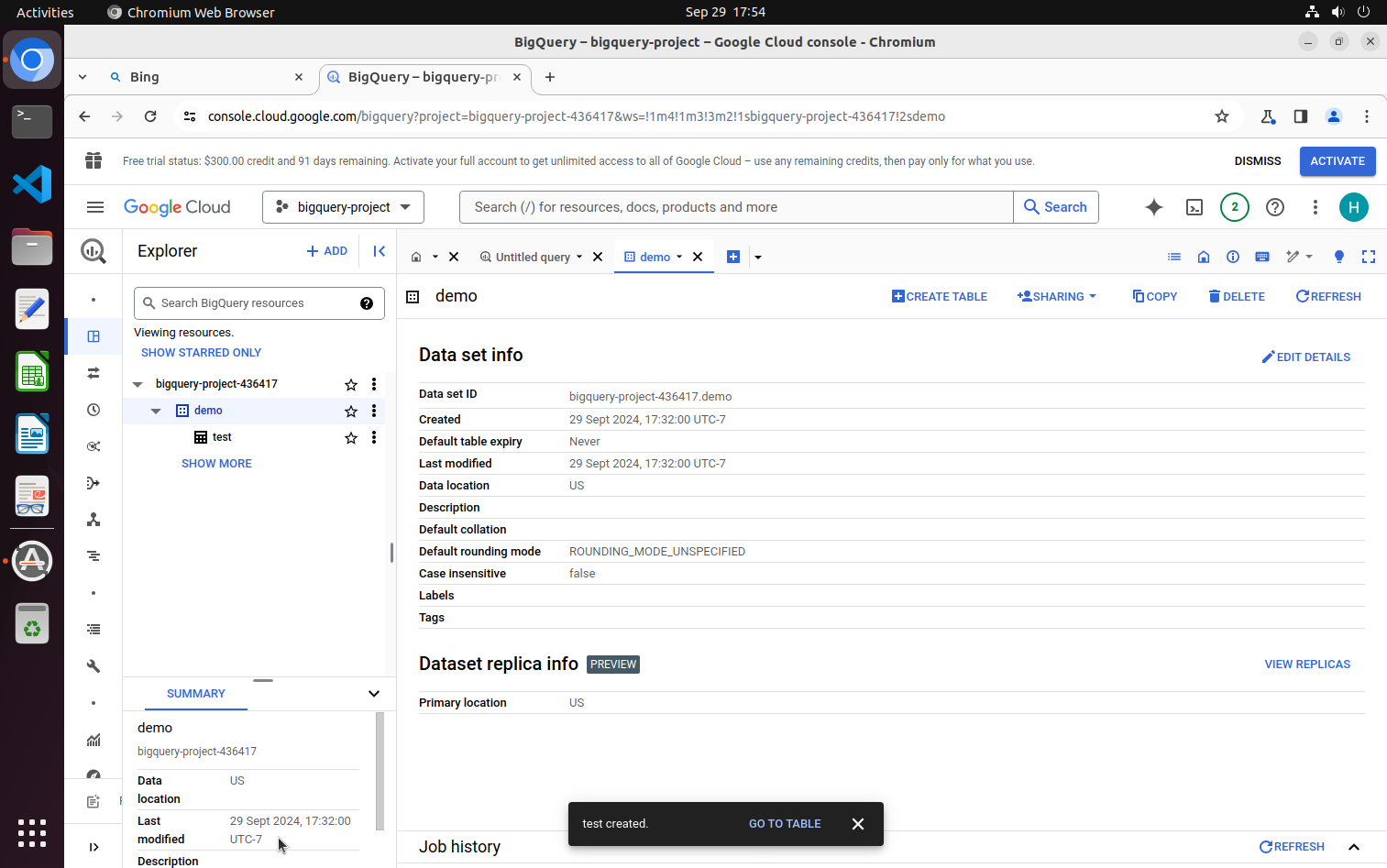} \\
\midrule
\bottomrule
\end{tabular}
\label{tab:data_synthesis_bigquery_7}
\end{table}

\begin{table}[t!]
\caption{Instructions generated from trajectory from Table~\ref{tab:data_synthesis_bigquery_1} to \ref{tab:data_synthesis_bigquery_7}}
\centering
\resizebox{\textwidth}{!}{
\begin{tabular}{c|c|l}
\toprule
sub-trajectory & type & instruction \\
\midrule
Observation 0 & & \\
$\downarrow$ & & \\
Action 1 & New task & When is dataset ``demo" created? \\
$\downarrow$ & & \\
Observation 1)  &  & \\
\midrule
Observation 1 &  & Replicate the following: We are currently at the Google Cloud  \\
$\downarrow$ & & Console interface, specifically focused on a BigQuery project. \\
Action 2 & Replicate trajectory & The browser window displays details of a dataset named "demo" \\
$\downarrow$ & & within a BigQuery project. The interface provides information \\
Observation 2 & & about the dataset, including its creation date, last modified time, \\
& & data location (US), and other properties like default table expiry \\
& & and rounding mode. On the left side of the screen, there's a \\
& & navigation panel showing the Explorer view with the "demo"  \\
& & dataset selected. The top of the screen shows the Google Cloud \\
& & header with project selection and search functionality. \\
& & The overall layout is characteristic of a cloud-based data \\
& & management platform, with options to create tables, share data, \\
& & and manage dataset properties. \\
& & After taking the action to click the CREATE TABLE button,  \\
& & we go to the user interface for creating a table. The screen \\
& & displays a form titled "Create table" with various fields and \\
& & options. The source section allows selecting a table to create \\
& & from, while the destination section includes fields for project, \\
& & dataset, and table name. There's also a schema section and  \\
& & partition and cluster settings. The interface is part of the Google \\
& & Cloud Console, as evident from the sidebar on the left showing \\
& & different Cloud services and project navigation.
\\
\midrule
Observation 4 & & \\
$\downarrow$ & & \\
Action 5 & & \\
$\downarrow$ & & \\
Observation 5 & & \\
$\downarrow$ & & \\
Action 6 & & \\
$\downarrow$ & & \\
Observation 6 & New task & Select test.csv in the bucket test-1616 in Google Cloud Storage \\
$\downarrow$ & & as the table source.\\
Action 7 & & \\
$\downarrow$ & & \\
Observation 7 & & \\
$\downarrow$ & & \\
Action 8 & & \\
$\downarrow$ & & \\
Observation 8 & & \\
\bottomrule
\end{tabular}

}
\label{tab:generated_instructions_1}
\end{table}

\begin{table}[t!]
\caption{Instructions generated from trajectory from Table~\ref{tab:data_synthesis_bigquery_1} to \ref{tab:data_synthesis_bigquery_7}}
\centering
\resizebox{\textwidth}{!}{
\begin{tabular}{c|c|l}
\toprule
sub-trajectory & type & instruction \\
\midrule
Observation 8 & & Replicate the following: We are in the the interface for creating \\
$\downarrow$ & &  a table in Google Cloud's BigQuery service. The page is divided \\
Action 9 & & into several sections. At the top, it indicates the user is creating \\
$\downarrow$ & & a table from a Google Cloud Storage source, with a CSV file \\
Observation 9 & Replicate trajectory & selected. The destination section shows the project ID and allows \\
$\downarrow$ & & input for the dataset and table name. The destination table is \\
Action 10 & & empty. The table type is set to ``Native table". At the bottom,  \\
$\downarrow$ & & there's an option for schema detection, with buttons to  create the  \\
Observation 10 & & table or cancel the operation. The left side of the screen displays a  \\
& & navigation menu for the Google Cloud Console, including options \\
& & like Explorer and various project-related items. The overall layout \\
& & suggests this is part of a larger cloud data management and \\
& & analysis platform. After we click on the text box Table, we select \\
& & and focus on the text box. We then type ``test" into the box, which  \\
& & gives the table a name. Except the textbox we are working on,  \\
& & the other parts of the webpage has not changed after clicking \\
& &  and typing. \\
\midrule
Observation 0 & & \\
$\downarrow$ & & \\
Action 1 & & \\
$\downarrow$ & & \\
Observation 1 & New task & Link CSV file in Google Cloud Storage to BigQuery \\
$\downarrow$ & & \\
Action 2 & & \\
$\downarrow$ & & \\
...... & & \\
$\downarrow$ & & \\
Observation 13 &  &  \\
\bottomrule
\end{tabular}

}
\label{tab:generated_instructions_2}
\end{table}

%% file: tables/prompts/self_instruct.tex
\begin{table}[ht]
\caption{self-instruct prompts to propose instructions based on tutorials, documentations and FAQs.}
\centering
\begin{tabular}{p{13cm}}
\toprule
\{Documentation\}
\bigbreak
Based on the tutorial, examplify 3 tasks that users frequently perform. \\
User the following format to output: \\
\* ... \\
\* ... \\
\bottomrule
\end{tabular}
\label{tab:self_instruct_prompt}
\end{table}

%% file: tables/prompts/sub_trajectory.tex
\begin{table}[ht]
\caption{Prompts to summarize (sub-)trajectories or propose new tasks based on the (sub-)trajectories.}
\centering
\begin{tabular}{p{13cm}}
\toprule
\textbf{Prompt 1} \\
\midrule
Below is a trajectory to complete a task. \\
Observation: \\
\{Observation$_i$\} \\
Action: \\
\{Action$_{i+1}$\} \\
Observation: \\
\{Observation$_{i+1}$\} \\
Action: \\
\{Action$_{i+2}$\} \\
... \\
Action: \\
\{Action$_{j-1}$\} \\
Observation: \\
\{Observation$_j$\} \\
\bigbreak
Please write a reasonable task instruction that is completed by the trajectory. \\
Wrap the instruction with \texttt{\textasciigrave\textasciigrave\textasciigrave}. \\
\midrule
\textbf{Prompt 2} \\
\midrule
Below is a trajectory to complete a task. \\
Observation: \\
\{Observation$_i$\} \\
Action: \\
\{Action$_{i+1}$\} \\
Observation: \\
\{Observation$_{i+1}$\} \\
Action: \\
\{Action$_{i+2}$\} \\
... \\
Action: \\
\{Action$_{j-1}$\} \\
Observation: \\
\{Observation$_j$\} \\
\bigbreak
Please summarize the trajectory about each observation and changes after each action.  \\
Wrap the summarization with \texttt{\textasciigrave\textasciigrave\textasciigrave}. \\
\bottomrule
\end{tabular}
\label{tab:sub_trajectory_prompt_1}
\end{table}

%% file: tables/prompts/filter.tex
\begin{table}[ht]
\caption{LLM prompts to filter low-quality data}
\centering
\begin{tabular}{p{13cm}}
\toprule
Task instruction: \\
\{instruction\} \\
Below is the trajectory to complete the task. \\
Observation: \\
\{Observation$_i$\} \\
Action: \\
\{Action$_{i+1}$\} \\
Observation: \\
\{Observation$_{i+1}$\} \\
Action: \\
\{Action$_{i+2}$\} \\
... \\
Action: \\
\{Action$_{j-1}$\} \\
Observation: \\
\{Observation$_j$\} \\
\bigbreak
Here are the criteria to indicate a good pair of the instruction and the trajectory: \\
1. The instruction and the trajectory are aligned, which means the trajectory successfully accomplishes the goal in the instruction. \\
2. The trajectory is coherent, indicating that each action is logical based on its previous observation and the actions do not contradict with each other based on the task instruction. \\
3. The trajectory is natural, meaning that the trajectory closely mimics real-world interactions and a human user would possibly perform it when engaging in the environment. \\
4. The trajectory is reasonable, indicating that the trajectory finishes the task instruction using a reasonable solution, e.g., not using an over-complicated method, not over-simply the problem, not going back and forth in states, etc. 
\bigbreak
Please answer yes if the task instruction and the trajectory satisfies all the criteria, otherwise, answer with no. \\
\bottomrule
\end{tabular}
\label{tab:filter}
\end{table}

%% file: tables/prompts/inference.tex
\begin{table}[ht]
\caption{Model inference prompts without external knowledge}
\centering
\begin{tabular}{p{13cm}}
\toprule
SYSTEM MESSAGE: \\
\{system message\} \\
OBJECTIVE: \\
\{task instruction\} \\
INTERACTION HISTORY: \\
\{interaction history\} \\
OBSERVATIONS: \\
\{observations\} \\
\bigbreak
Your REASONING and ACTION in the format: \\
REASON: \\
Your reason to choose a specific action. \\
ACTION: \\
Your action \\
\bottomrule
\end{tabular}
\label{tab:inference_prompt_1}
\end{table}

\begin{table}[ht]
\caption{Model inference prompts with external knowledge}
\centering
\begin{tabular}{p{13cm}}
\toprule
SYSTEM MESSAGE: \\
\{system message\} \\
ADDITIONAL INFORMATION FOR REFERENCE: \\
\{external knowledge\} \\
OBJECTIVE: \\
\{task instruction\} \\
INTERACTION HISTORY: \\
\{interaction history\} \\
OBSERVATIONS: \\
\{observations\} \\
\bigbreak
Your REASONING and ACTION in the format: \\
REASON: \\
Your reason to choose a specific action. \\
ACTION: \\
Your action \\
\bottomrule
\end{tabular}
\label{tab:inference_prompt_2}
\end{table}

\begin{table}[ht]
\caption{Expected model outputs}
\centering
\begin{tabular}{p{13cm}}
\toprule
REASON: \\
... \\
ACTION: \\
... \\
\bottomrule
\end{tabular}
\label{tab:expected_output}
\end{table}

%% file: tables/prompts/write_query.tex
\begin{table}[ht]
\caption{Model prompts to write query for retrieval}
\centering
\begin{tabular}{p{13cm}}
\toprule
SYSTEM MESSAGE: \\
\{system message\} \\
Here is the final goal we want to achieve: \\
\{task instruction\} \\
To achieve the goal, we have done the following: \\
\{interaction history\} \\
Now, we have observed: \\
\{observations\} \\
\bigbreak
To better finish the task, write a query to ask for useful information, e.g., what kind of examples or interaction history will be helpful to predict the next action. \\
\bottomrule
\end{tabular}
\label{tab:write_query}
\end{table}

%% file: tables/data_samples/osworld_filtered.tex
\begin{table}[ht]
\caption{\textbf{\textcolor{rebuttal}{OSWorld example (filtered)}}}
\centering
\begin{tabular}{p{13cm}}
\toprule
\textbf{\textcolor{rebuttal}{Instruction}} \\
\midrule
\textcolor{rebuttal}{Sum numbers in the first column.} \\
\midrule
\textbf{\textcolor{rebuttal}{Observation 0 (Interface of the software LibreOffice Calc)}} \\
\midrule
\includegraphics[width=0.8\textwidth]{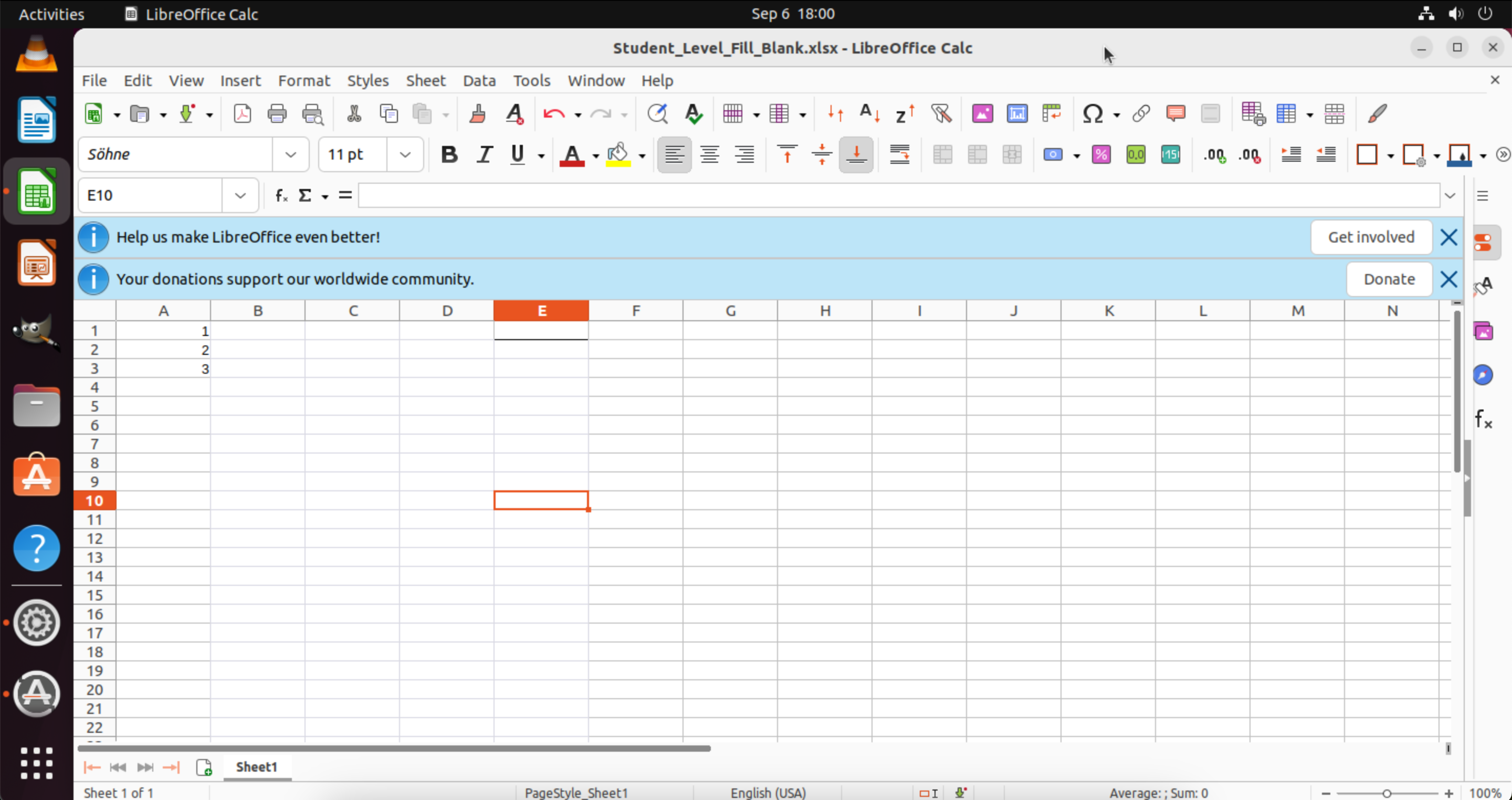} \\
\midrule
\textbf{\textcolor{rebuttal}{Action 1}} \\
\midrule
\textcolor{rebuttal}{import pyautogui} \\
\textcolor{rebuttal}{pyautogui.click(543,126) // click Tools.} \\
\midrule
\textbf{\textcolor{rebuttal}{Observation 1}} \\
\midrule
\includegraphics[width=0.8\textwidth]{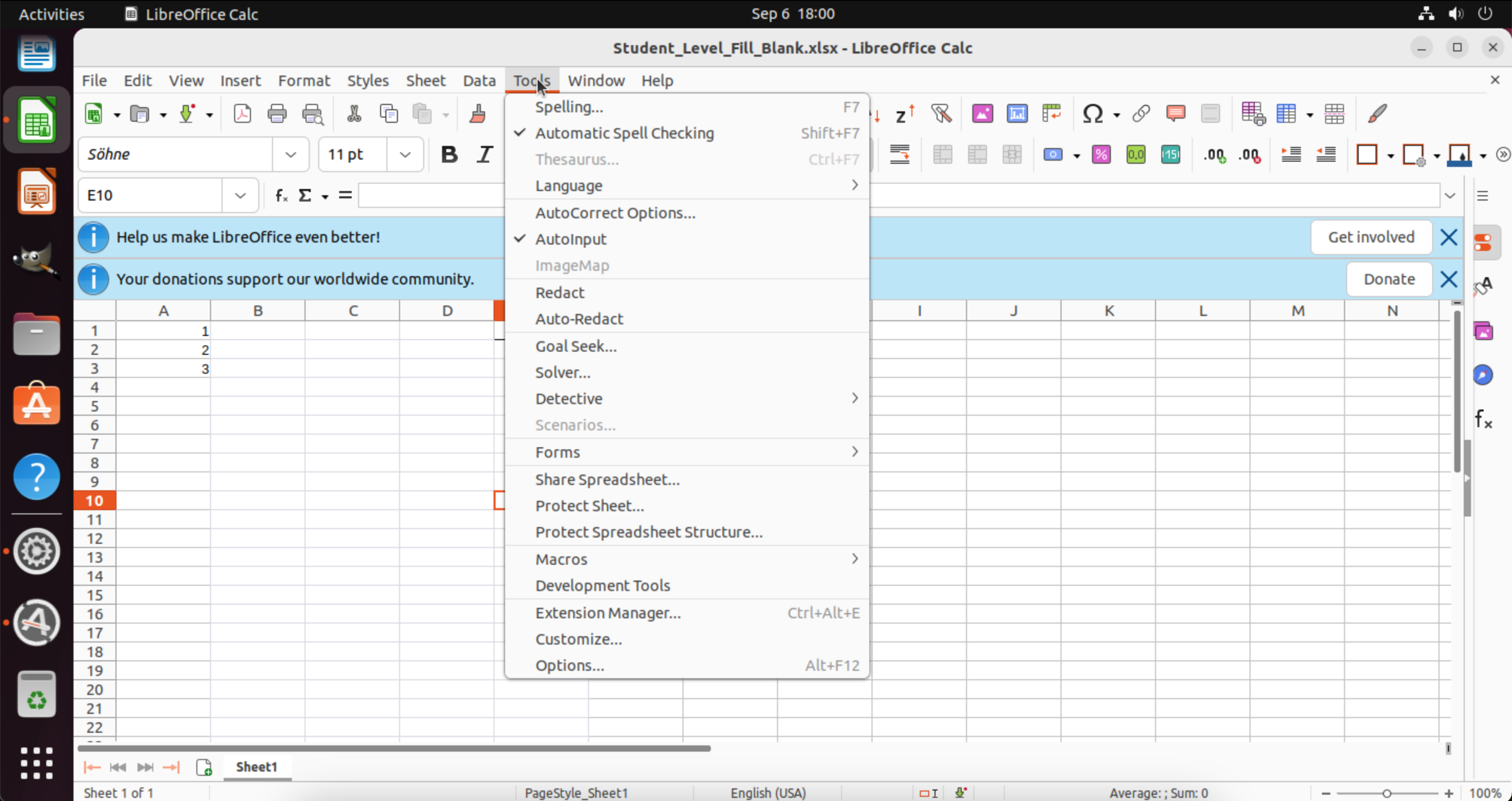} \\
\midrule
\bottomrule
\end{tabular}

\label{tab:osworld_filtered_1}
\end{table}

\begin{table}[ht]
\caption{\textbf{\textcolor{rebuttal}{OSWorld example (filtered) cont.}}}
\centering
\begin{tabular}{p{13cm}}
\toprule
\textbf{\textcolor{rebuttal}{Action 2}} \\
\midrule
\textcolor{rebuttal}{import pyautogui} \\
\textcolor{rebuttal}{pyautogui.click(543,580) // click Solver.} \\
\midrule
\textbf{\textcolor{rebuttal}{Observation 2}} \\
\midrule
\includegraphics[width=0.8\textwidth]{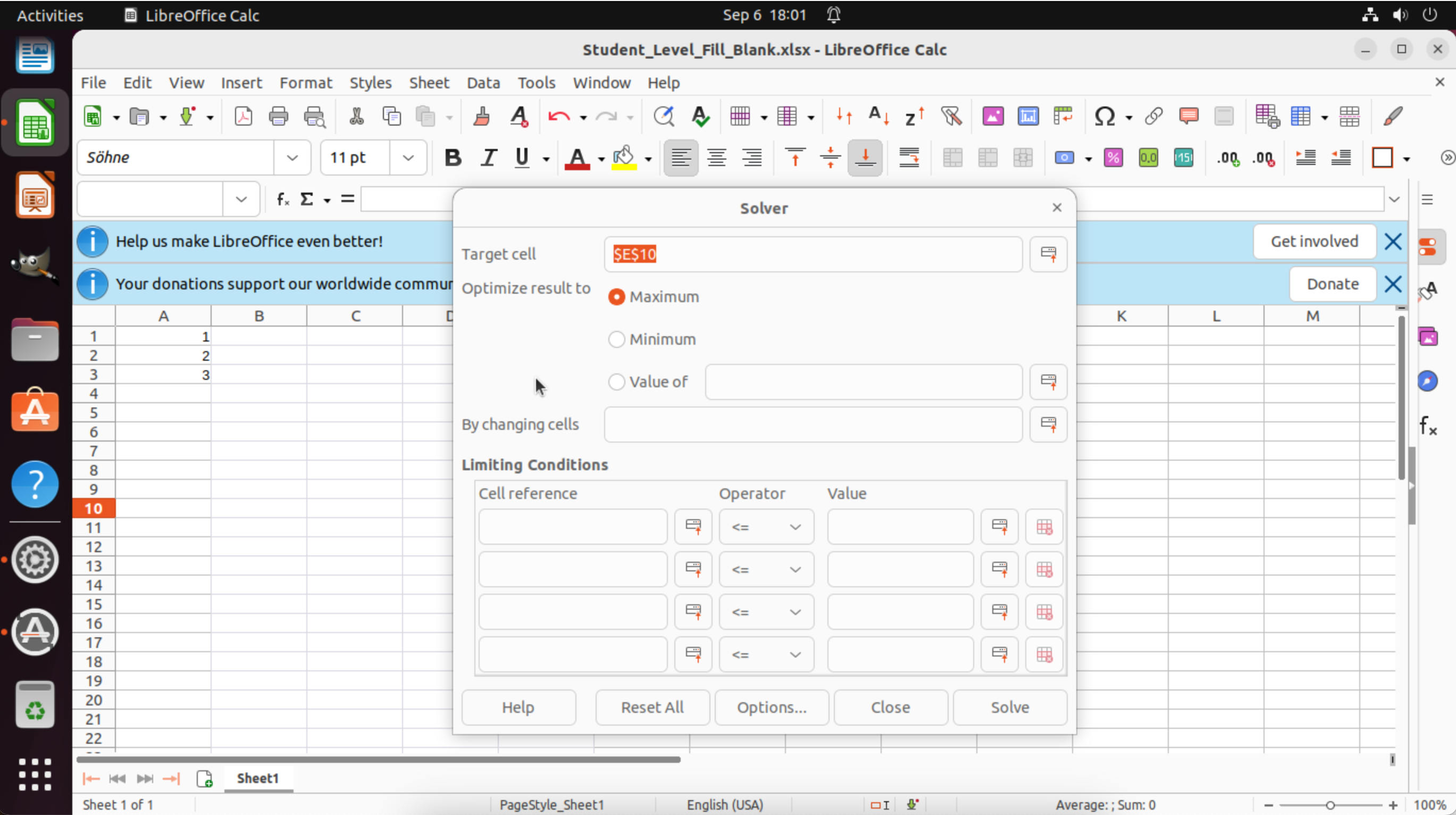} \\
\midrule
\textbf{\textcolor{rebuttal}{Action 3}} \\
\midrule
\textcolor{rebuttal}{import pyautogui} \\
\textcolor{rebuttal}{pyautogui.click(772,892) // click Close.} \\
\midrule
\textbf{\textcolor{rebuttal}{Observation 3}} \\
\midrule
\includegraphics[width=0.8\textwidth]{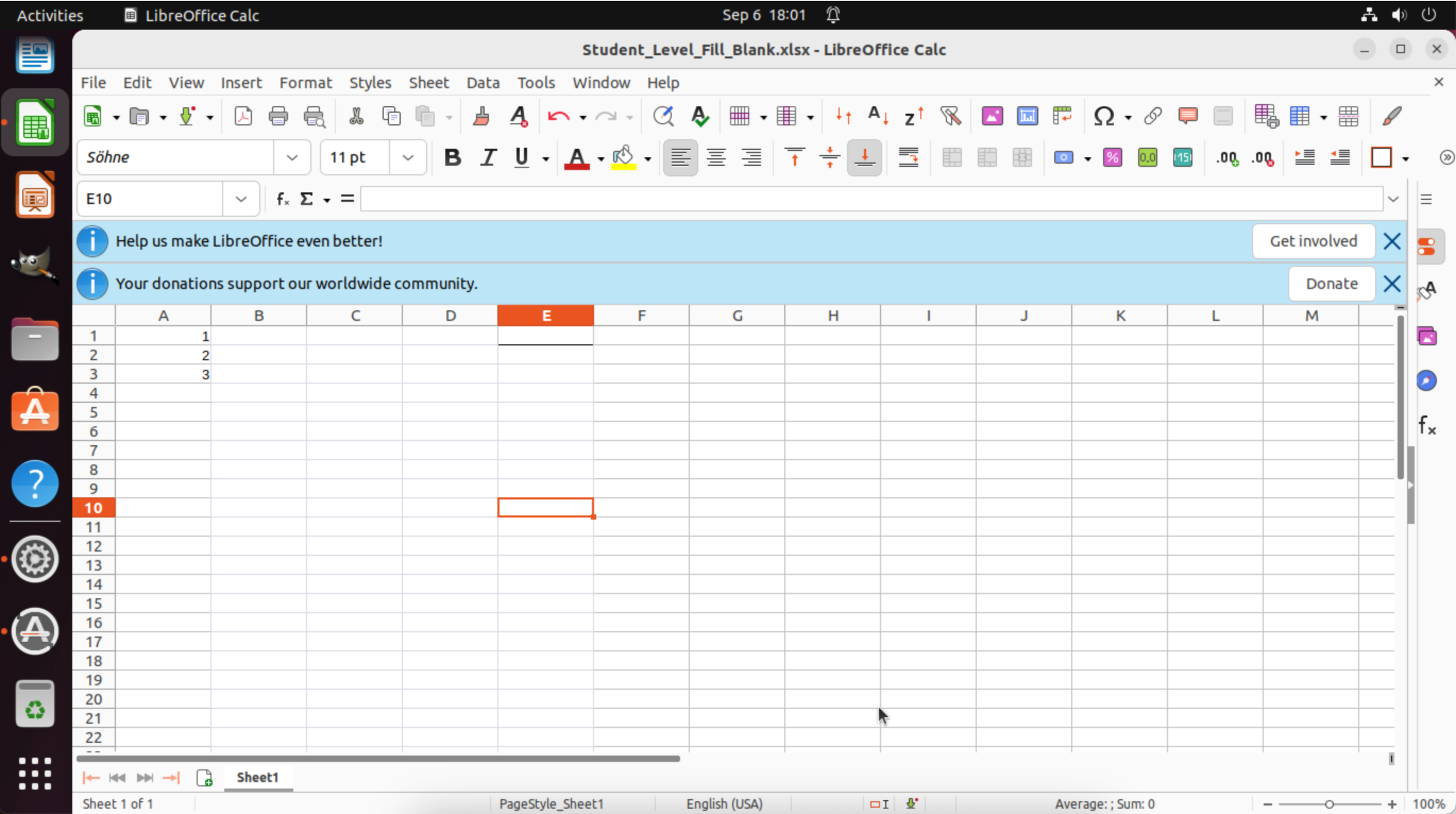} \\
\midrule
\bottomrule
\end{tabular}

\label{tab:osworld_filtered_2}
\end{table}

\begin{table}[ht]
\caption{\textbf{\textcolor{rebuttal}{OSWorld example (filtered) cont.}}}
\centering
\begin{tabular}{p{13cm}}
\toprule
\textbf{\textcolor{rebuttal}{Action 4}} \\
\midrule
\textcolor{rebuttal}{import pyautogui} \\
\textcolor{rebuttal}{pyautogui.click(520,126) // click Data.} \\
\midrule
\textbf{\textcolor{rebuttal}{Observation 4}} \\
\midrule
\includegraphics[width=0.8\textwidth]{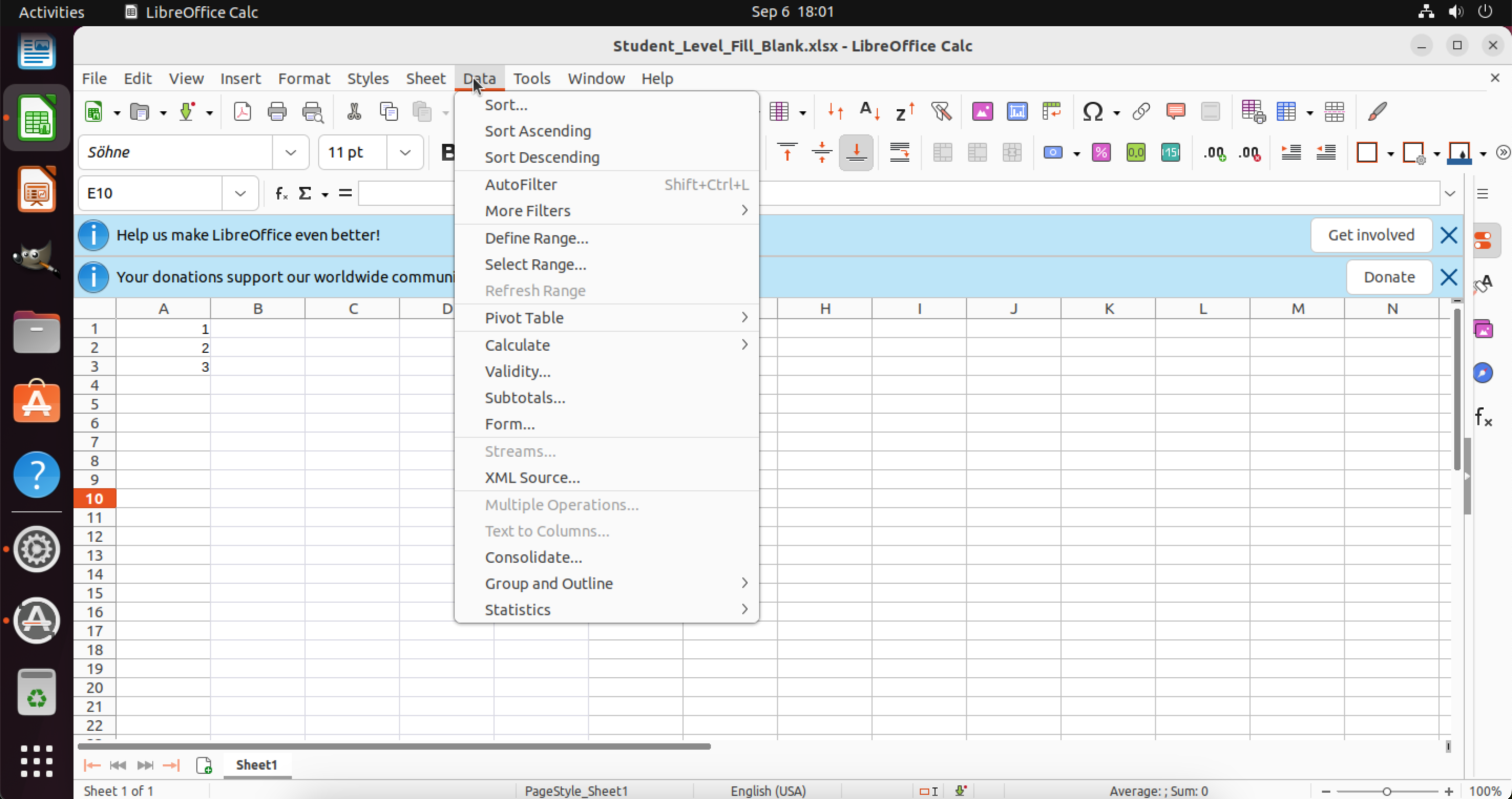} \\
\midrule
\textbf{\textcolor{rebuttal}{Action 5}} \\
\midrule
\textcolor{rebuttal}{import pyautogui} \\
\textcolor{rebuttal}{pyautogui.moveTo(520,562) // move to Calculate.} \\
\midrule
\textbf{\textcolor{rebuttal}{Observation 5}} \\
\midrule
\includegraphics[width=0.8\textwidth]{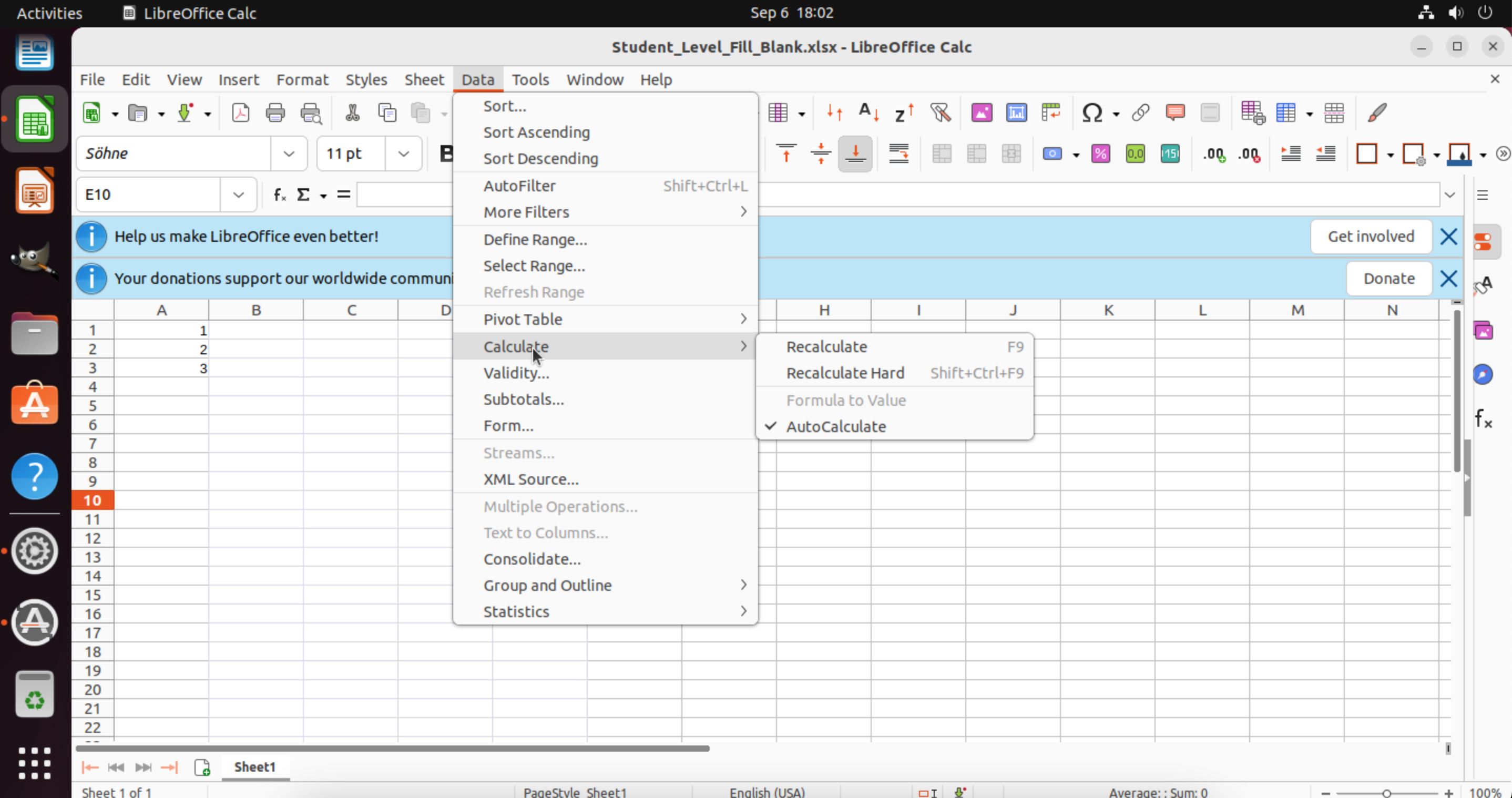} \\
\midrule
\bottomrule
\end{tabular}

\label{tab:osworld_filtered_3}
\end{table}

\begin{table}[ht]
\caption{\textbf{\textcolor{rebuttal}{OSWorld example (filtered) cont.}}}
\centering
\begin{tabular}{p{13cm}}
\toprule
\textbf{\textcolor{rebuttal}{Action 6}} \\
\midrule
\textcolor{rebuttal}{import pyautogui} \\
\textcolor{rebuttal}{pyautogui.click(498,126) // click Sheet.} \\
\midrule
\textbf{\textcolor{rebuttal}{Observation 6}} \\
\midrule
\includegraphics[width=0.8\textwidth]{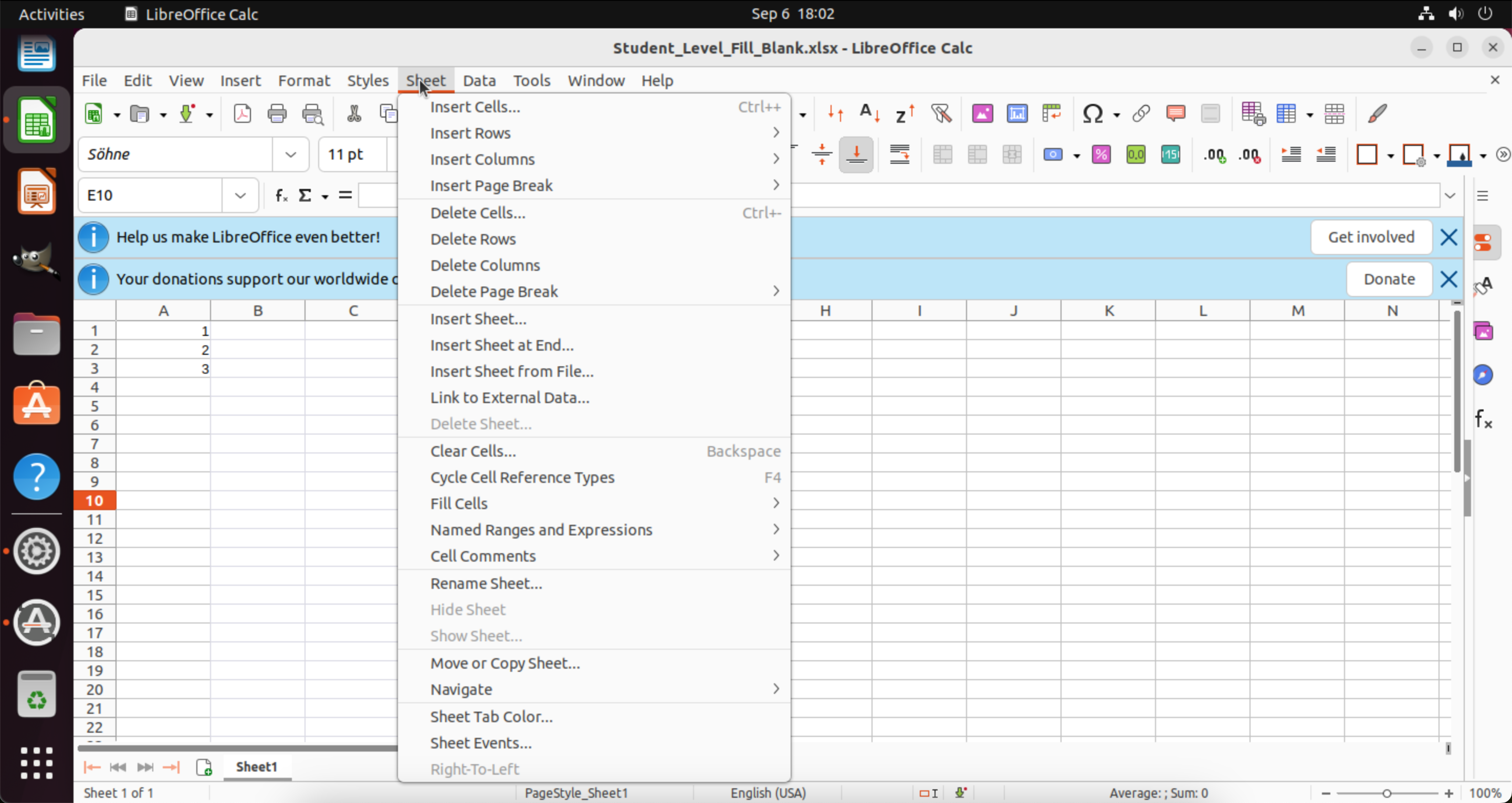} \\
\midrule
\textbf{\textcolor{rebuttal}{Action 7}} \\
\midrule
\textcolor{rebuttal}{import pyautogui} \\
\textcolor{rebuttal}{pyautogui.click(152,591) // click cell A4.} \\
\midrule
\textbf{\textcolor{rebuttal}{Observation 7}} \\
\midrule
\includegraphics[width=0.8\textwidth]{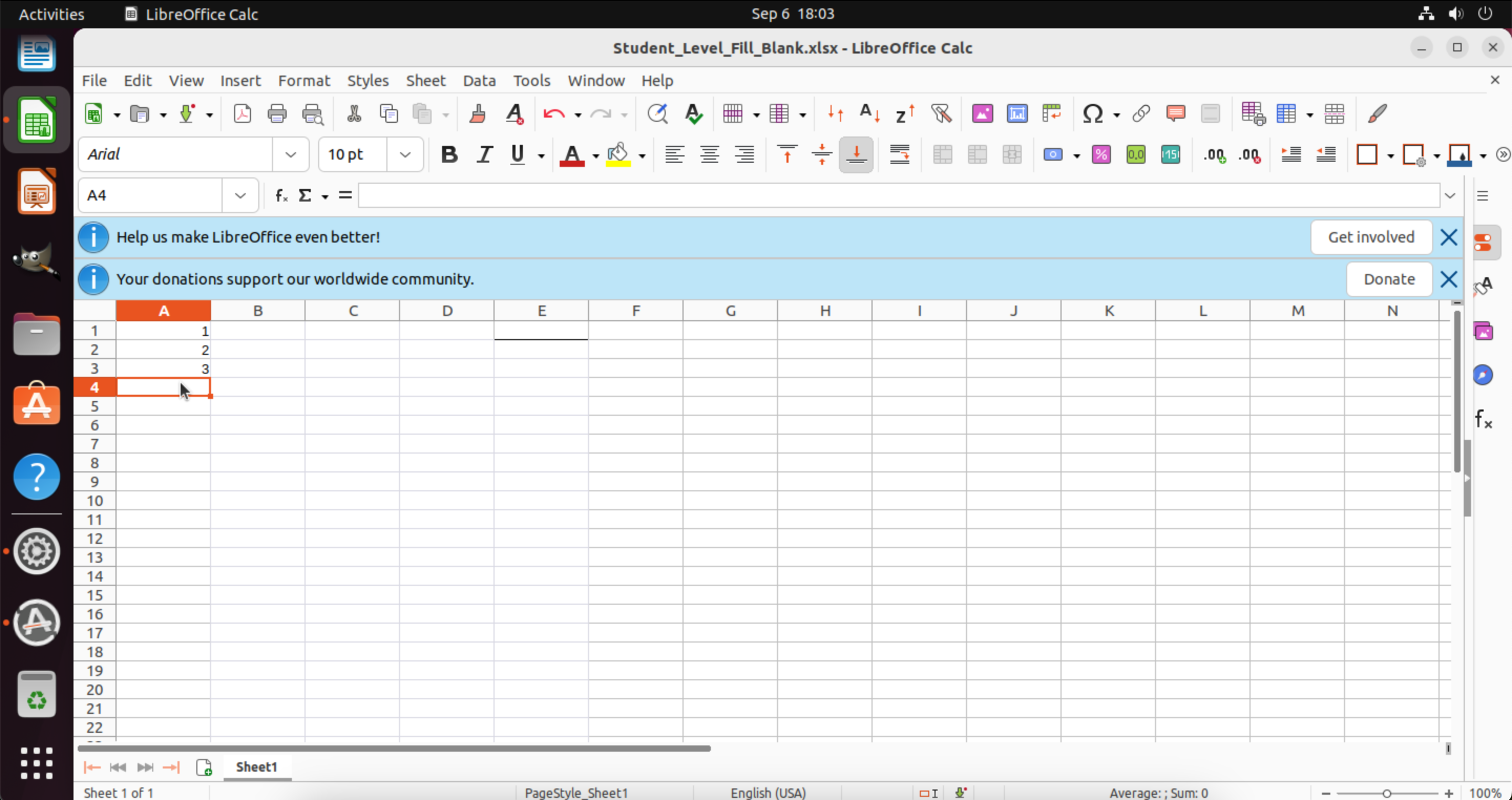} \\
\midrule
\bottomrule
\end{tabular}

\label{tab:osworld_filtered_4}
\end{table}

\begin{table}[ht]
\caption{\textbf{\textcolor{rebuttal}{OSWorld example (filtered) cont.}}}
\centering
\begin{tabular}{p{13cm}}
\toprule
\textbf{\textcolor{rebuttal}{Action 8}} \\
\midrule
\textcolor{rebuttal}{import pyautogui} \\
\textcolor{rebuttal}{pyautogui.click(480,302) // click the formula box.} \\
\midrule
\textbf{\textcolor{rebuttal}{Observation 8}} \\
\midrule
\includegraphics[width=0.8\textwidth]{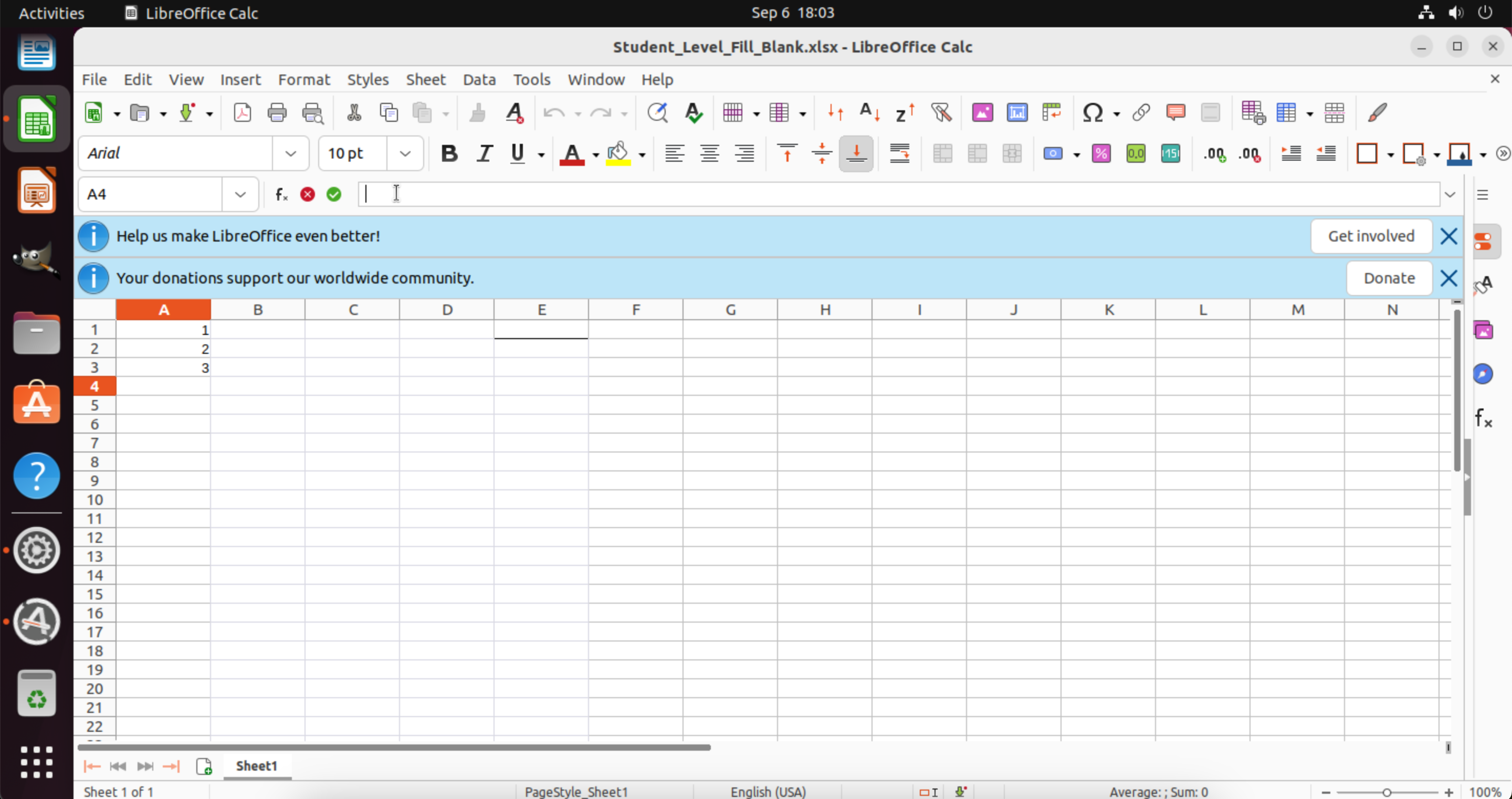} \\
\midrule
\textbf{\textcolor{rebuttal}{Action 9}} \\
\midrule
\textcolor{rebuttal}{import pyautogui} \\
\textcolor{rebuttal}{pyautogui.write("=SUM(A1:A3)") // type the formula.} \\
\midrule
\textbf{\textcolor{rebuttal}{Observation 9}} \\
\midrule
\includegraphics[width=0.8\textwidth]{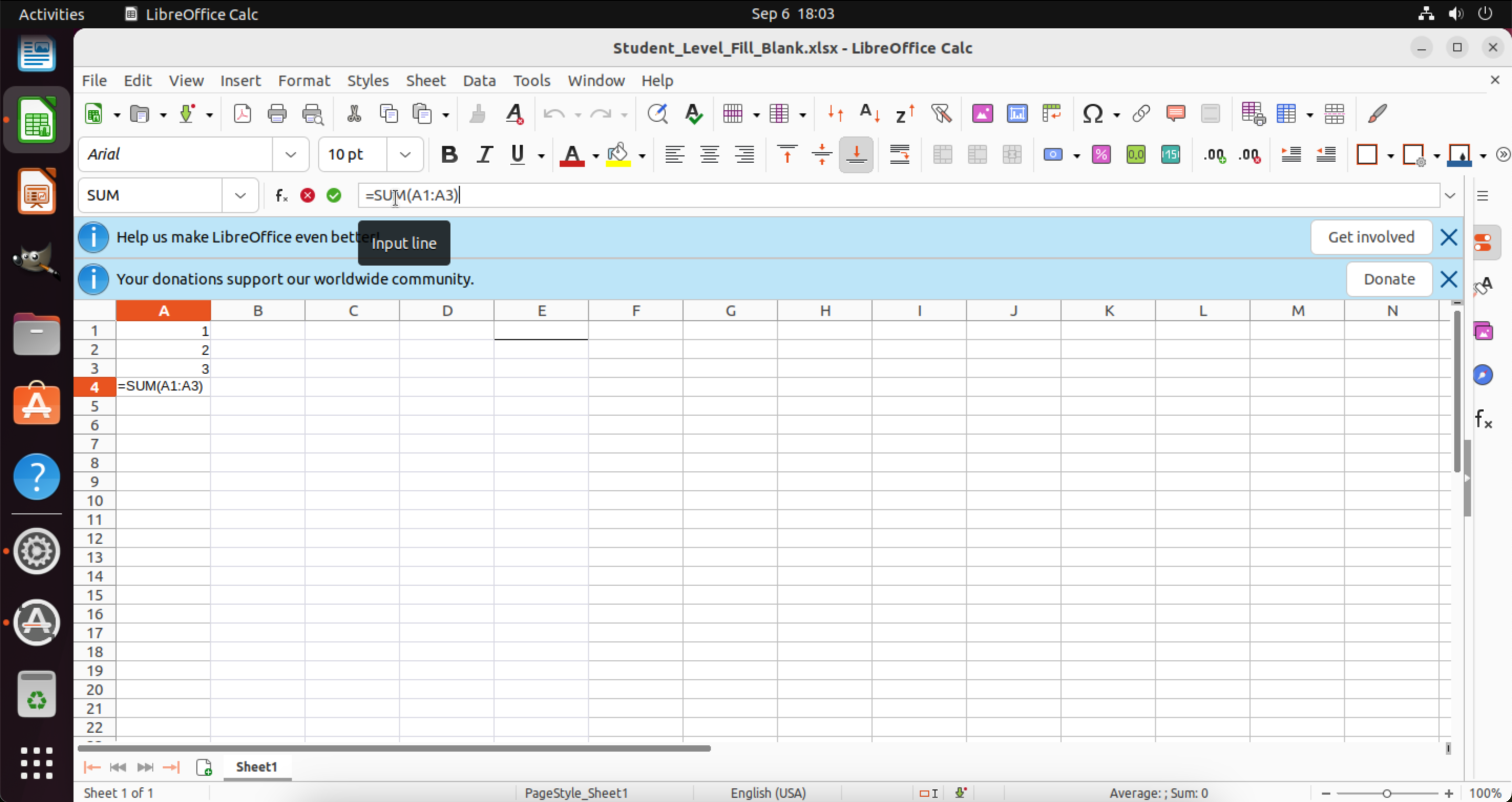} \\
\midrule
\bottomrule
\end{tabular}

\label{tab:osworld_filtered_5}
\end{table}

\begin{table}[ht]
\caption{\textbf{\textcolor{rebuttal}{OSWorld example (filtered) cont.}}}
\centering
\begin{tabular}{p{13cm}}
\toprule
\textbf{\textcolor{rebuttal}{Action 10}} \\
\midrule
\textcolor{rebuttal}{import pyautogui} \\
\textcolor{rebuttal}{pyautogui.press("enter")} \\
\midrule
\textbf{\textcolor{rebuttal}{Observation 10}} \\
\midrule
\includegraphics[width=0.8\textwidth]{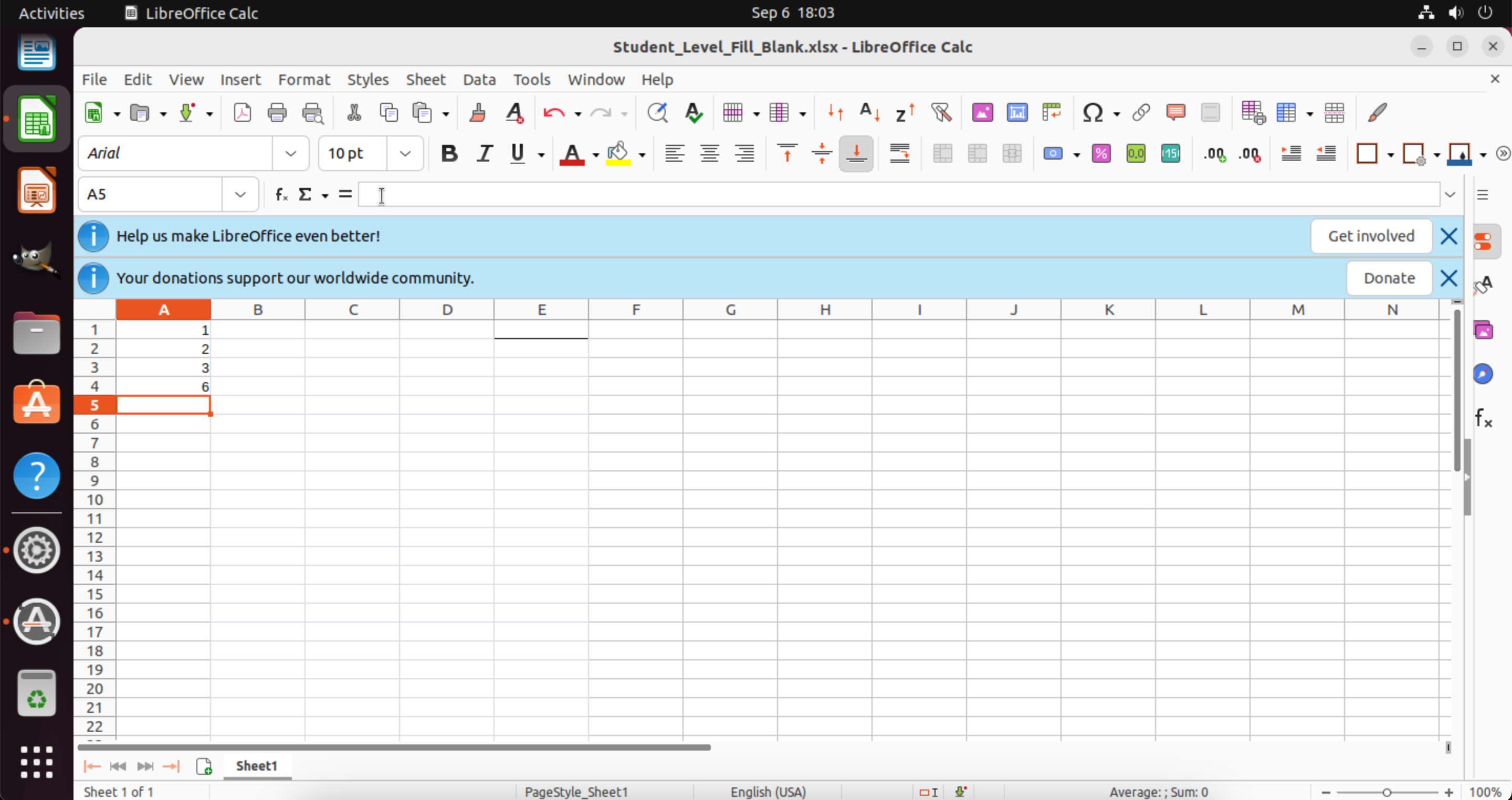} \\
\midrule
\bottomrule
\end{tabular}

\label{tab:osworld_filtered_6}
\end{table}

%% file: tables/data_samples/webarena_filtered.tex
\begin{table}[ht]
\caption{\textbf{\textcolor{rebuttal}{WebArena example (filtered)}}}
\centering
\begin{tabular}{p{13cm}}
\toprule
\textbf{\textcolor{rebuttal}{Instruction}} \\
\midrule
\textcolor{rebuttal}{What are items ordered in the latest cancelled order?} \\
\midrule
\textbf{\textcolor{rebuttal}{Observation 0}} \\
\midrule
\includegraphics[width=0.8\textwidth]{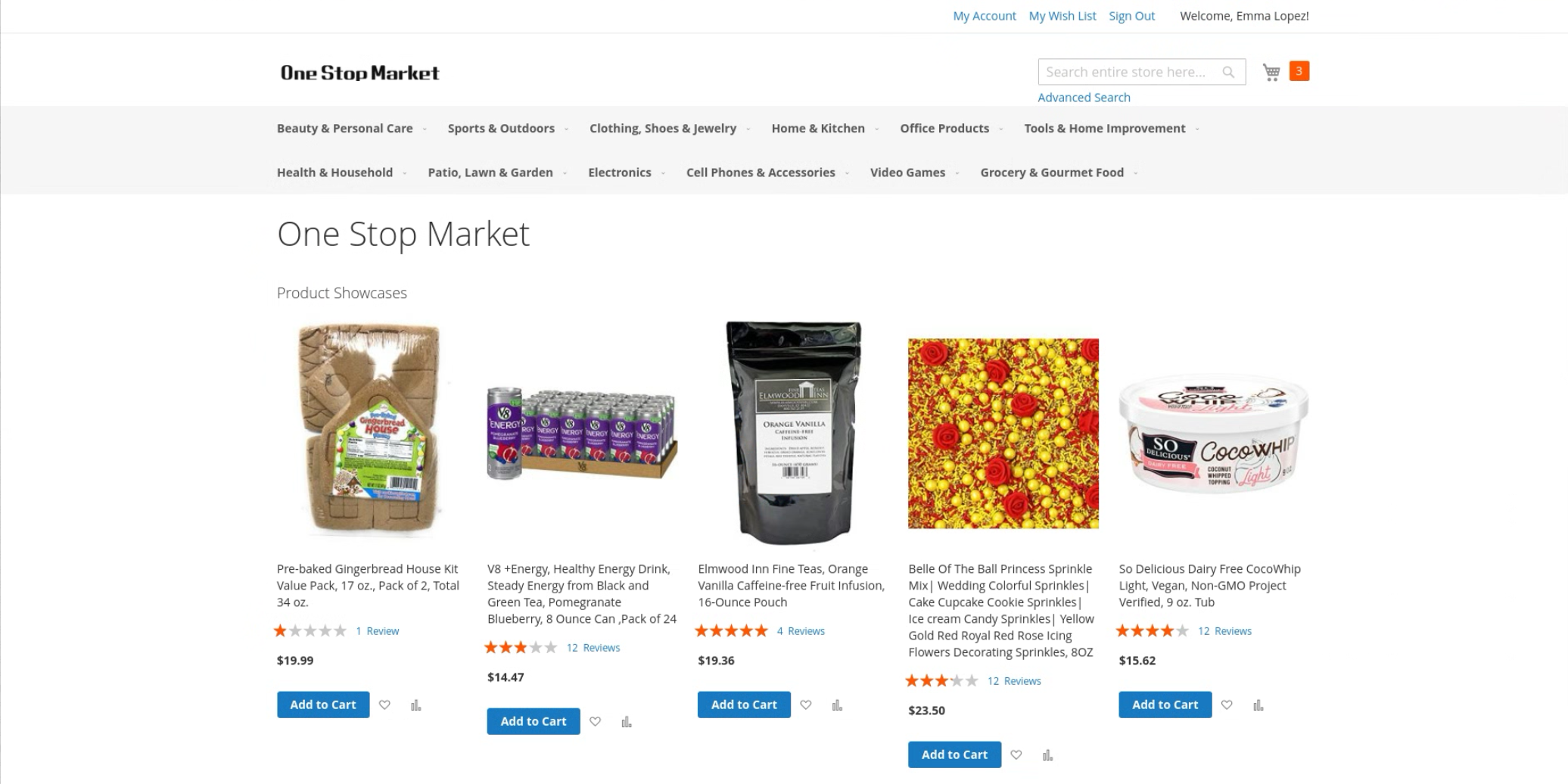} \\
\midrule
\textbf{\textcolor{rebuttal}{Action 1}} \\
\midrule
\textcolor{rebuttal}{click [4918] // click the button "My Account"} \\
\midrule
\textbf{\textcolor{rebuttal}{Observation 1}} \\
\midrule
\includegraphics[width=0.8\textwidth]{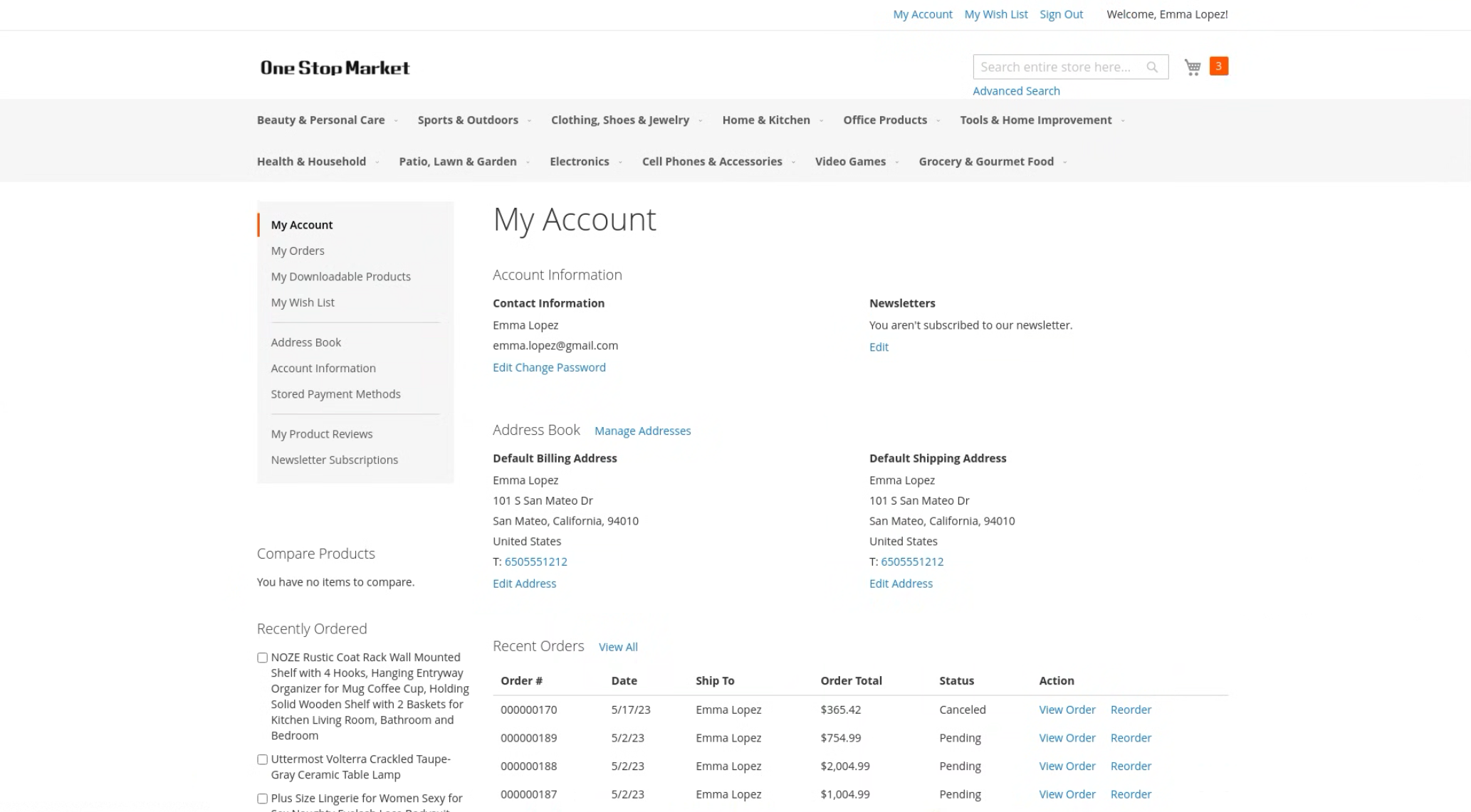} \\
\midrule
\bottomrule
\end{tabular}

\label{tab:webarena_filtered_1}
\end{table}

\begin{table}[ht]
\caption{\textbf{\textcolor{rebuttal}{WebArena example (filtered) cont.}}}
\centering
\begin{tabular}{p{13cm}}
\toprule
\textbf{\textcolor{rebuttal}{Action 2}} \\
\midrule
\textcolor{rebuttal}{click [4922] // click the button "My Orders".} \\
\midrule
\textbf{\textcolor{rebuttal}{Observation 2}} \\
\midrule
\includegraphics[width=0.8\textwidth]{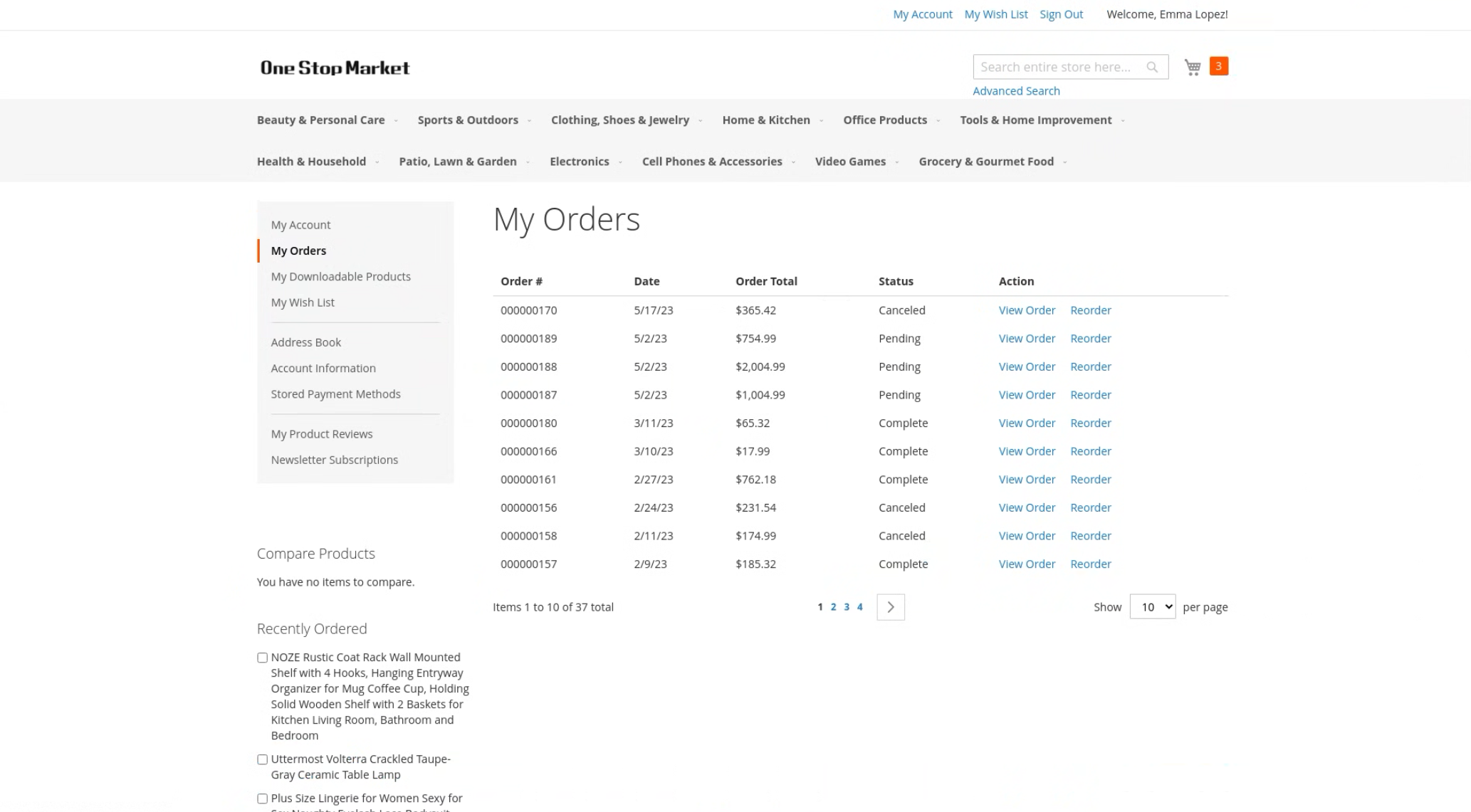} \\
\midrule
\textbf{\textcolor{rebuttal}{Action 3}} \\
\midrule
\textcolor{rebuttal}{click [6357] // click the button "View Order".} \\
\midrule
\textbf{\textcolor{rebuttal}{Observation 3}} \\
\midrule
\includegraphics[width=0.8\textwidth]{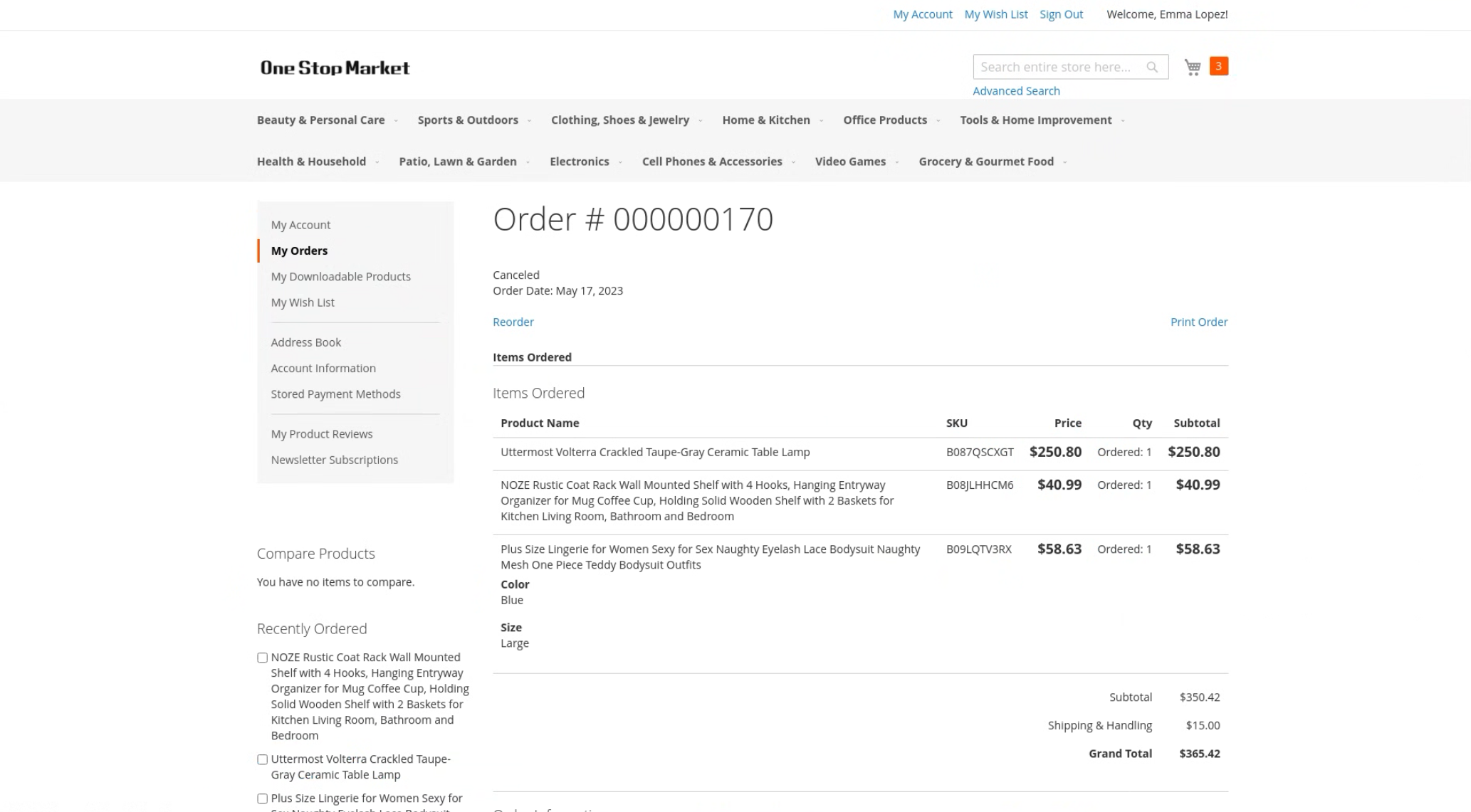} \\
\midrule
\bottomrule
\end{tabular}

\label{tab:webarena_filtered_2}
\end{table}

\begin{table}[ht]
\caption{\textbf{\textcolor{rebuttal}{WebArena example (filtered) cont.}}}
\centering
\begin{tabular}{p{13cm}}
\toprule
\textbf{\textcolor{rebuttal}{Action 4}} \\
\midrule
\textcolor{rebuttal}{click [4922] // click the button "My Orders".} \\
\midrule
\textbf{\textcolor{rebuttal}{Observation 4}} \\
\midrule
\includegraphics[width=0.8\textwidth]{images/data_examples/webarena_filtered/3.png} \\
\midrule
\textbf{\textcolor{rebuttal}{Action 5}} \\
\midrule
\textcolor{rebuttal}{click [6357] // click the button "View Order".} \\
\midrule
\textbf{\textcolor{rebuttal}{Observation 5}} \\
\midrule
\includegraphics[width=0.8\textwidth]{images/data_examples/webarena_filtered/4.png} \\
\midrule
\bottomrule
\end{tabular}

\label{tab:webarena_filtered_3}
\end{table}

\begin{table}[ht]
\caption{\textbf{\textcolor{rebuttal}{WebArena example (filtered) cont.}}}
\centering
\begin{tabular}{p{13cm}}
\toprule
\textbf{\textcolor{rebuttal}{Action 6}} \\
\midrule
\textcolor{rebuttal}{click [4922] // click the button "My Orders".} \\
\midrule
\textbf{\textcolor{rebuttal}{Observation 6}} \\
\midrule
\includegraphics[width=0.8\textwidth]{images/data_examples/webarena_filtered/3.png} \\
\midrule
\textbf{\textcolor{rebuttal}{Action 7}} \\
\midrule
\textcolor{rebuttal}{click [6357] // click the button "View Order".} \\
\midrule
\textbf{\textcolor{rebuttal}{Observation 7}} \\
\midrule
\includegraphics[width=0.8\textwidth]{images/data_examples/webarena_filtered/4.png} \\
\midrule
\bottomrule
\end{tabular}

\label{tab:webarena_filtered_4}
\end{table}

%% file: main.bbl
\begin{thebibliography}{52}
\providecommand{\natexlab}[1]{#1}
\providecommand{\url}[1]{\texttt{#1}}
\expandafter\ifx\csname urlstyle\endcsname\relax
  \providecommand{\doi}[1]{doi: #1}\else
  \providecommand{\doi}{doi: \begingroup \urlstyle{rm}\Url}\fi

\bibitem[Achiam et~al.(2023)Achiam, Adler, Agarwal, Ahmad, Akkaya, Aleman, Almeida, Altenschmidt, Altman, Anadkat, et~al.]{achiam2023gpt}
J.~Achiam, S.~Adler, S.~Agarwal, L.~Ahmad, I.~Akkaya, F.~L. Aleman, D.~Almeida, J.~Altenschmidt, S.~Altman, S.~Anadkat, et~al.
\newblock Gpt-4 technical report.
\newblock \emph{arXiv preprint arXiv:2303.08774}, 2023.

\bibitem[Aksitov et~al.(2023)Aksitov, Miryoosefi, Li, Li, Babayan, Kopparapu, Fisher, Guo, Prakash, Srinivasan, et~al.]{aksitov2023rest}
R.~Aksitov, S.~Miryoosefi, Z.~Li, D.~Li, S.~Babayan, K.~Kopparapu, Z.~Fisher, R.~Guo, S.~Prakash, P.~Srinivasan, et~al.
\newblock Rest meets react: Self-improvement for multi-step reasoning llm agent.
\newblock \emph{arXiv preprint arXiv:2312.10003}, 2023.

\bibitem[Anthropic(2024)]{claude35anthropic}
Anthropic.
\newblock Introducing claude 3.5 sonnet, 2024.
\newblock URL \url{https://www.anthropic.com/news/claude-3-5-sonnet}.

\bibitem[Ball et~al.(2023)Ball, Smith, Kostrikov, and Levine]{ball2023efficient}
P.~J. Ball, L.~Smith, I.~Kostrikov, and S.~Levine.
\newblock Efficient online reinforcement learning with offline data.
\newblock In \emph{International Conference on Machine Learning}, pages 1577--1594. PMLR, 2023.

\bibitem[Cao et~al.(2024)Cao, Lei, Wu, Chen, Fu, Gao, Xiong, Zhang, Mao, Hu, et~al.]{cao2024spider2}
R.~Cao, F.~Lei, H.~Wu, J.~Chen, Y.~Fu, H.~Gao, X.~Xiong, H.~Zhang, Y.~Mao, W.~Hu, et~al.
\newblock Spider2-v: How far are multimodal agents from automating data science and engineering workflows?
\newblock \emph{arXiv preprint arXiv:2407.10956}, 2024.

\bibitem[Chen et~al.(2023)Chen, Shu, Shareghi, Collier, Narasimhan, and Yao]{chen2023fireact}
B.~Chen, C.~Shu, E.~Shareghi, N.~Collier, K.~Narasimhan, and S.~Yao.
\newblock Fireact: Toward language agent fine-tuning.
\newblock \emph{arXiv preprint arXiv:2310.05915}, 2023.

\bibitem[Chen et~al.(2024{\natexlab{a}})Chen, Lin, Zeng, Zan, Wang, Cheshkov, Sun, Yu, Dong, Aliev, et~al.]{chen2024coder}
D.~Chen, S.~Lin, M.~Zeng, D.~Zan, J.-G. Wang, A.~Cheshkov, J.~Sun, H.~Yu, G.~Dong, A.~Aliev, et~al.
\newblock Coder: Issue resolving with multi-agent and task graphs.
\newblock \emph{arXiv preprint arXiv:2406.01304}, 2024{\natexlab{a}}.

\bibitem[Chen et~al.(2024{\natexlab{b}})Chen, Liu, Wang, Zhang, Liu, Lin, Chen, and Zhao]{chen2024agent}
Z.~Chen, K.~Liu, Q.~Wang, W.~Zhang, J.~Liu, D.~Lin, K.~Chen, and F.~Zhao.
\newblock Agent-flan: Designing data and methods of effective agent tuning for large language models.
\newblock \emph{arXiv preprint arXiv:2403.12881}, 2024{\natexlab{b}}.

\bibitem[Coulom(2006)]{coulom2006efficient}
R.~Coulom.
\newblock Efficient selectivity and backup operators in monte-carlo tree search.
\newblock In \emph{International conference on computers and games}, pages 72--83. Springer, 2006.

\bibitem[Deng et~al.(2024)Deng, Gu, Zheng, Chen, Stevens, Wang, Sun, and Su]{deng2024mind2web}
X.~Deng, Y.~Gu, B.~Zheng, S.~Chen, S.~Stevens, B.~Wang, H.~Sun, and Y.~Su.
\newblock Mind2web: Towards a generalist agent for the web.
\newblock \emph{Advances in Neural Information Processing Systems}, 36, 2024.

\bibitem[Drouin et~al.(2024)Drouin, Gasse, Caccia, Laradji, Del~Verme, Marty, Vazquez, Chapados, and Lacoste]{workarena2024}
A.~Drouin, M.~Gasse, M.~Caccia, I.~H. Laradji, M.~Del~Verme, T.~Marty, D.~Vazquez, N.~Chapados, and A.~Lacoste.
\newblock {W}ork{A}rena: How capable are web agents at solving common knowledge work tasks?
\newblock In R.~Salakhutdinov, Z.~Kolter, K.~Heller, A.~Weller, N.~Oliver, J.~Scarlett, and F.~Berkenkamp, editors, \emph{Proceedings of the 41st International Conference on Machine Learning}, volume 235 of \emph{Proceedings of Machine Learning Research}, pages 11642--11662. PMLR, 21--27 Jul 2024.
\newblock URL \url{https://proceedings.mlr.press/v235/drouin24a.html}.

\bibitem[Gulcehre et~al.(2023)Gulcehre, Paine, Srinivasan, Konyushkova, Weerts, Sharma, Siddhant, Ahern, Wang, Gu, et~al.]{gulcehre2023reinforced}
C.~Gulcehre, T.~L. Paine, S.~Srinivasan, K.~Konyushkova, L.~Weerts, A.~Sharma, A.~Siddhant, A.~Ahern, M.~Wang, C.~Gu, et~al.
\newblock Reinforced self-training (rest) for language modeling.
\newblock \emph{arXiv preprint arXiv:2308.08998}, 2023.

\bibitem[Gur et~al.(2023)Gur, Furuta, Huang, Safdari, Matsuo, Eck, and Faust]{gur2023real}
I.~Gur, H.~Furuta, A.~Huang, M.~Safdari, Y.~Matsuo, D.~Eck, and A.~Faust.
\newblock A real-world webagent with planning, long context understanding, and program synthesis.
\newblock \emph{arXiv preprint arXiv:2307.12856}, 2023.

\bibitem[Hu et~al.(2021)Hu, Shen, Wallis, Allen-Zhu, Li, Wang, Wang, and Chen]{hu2021lora}
E.~J. Hu, Y.~Shen, P.~Wallis, Z.~Allen-Zhu, Y.~Li, S.~Wang, L.~Wang, and W.~Chen.
\newblock Lora: Low-rank adaptation of large language models.
\newblock \emph{arXiv preprint arXiv:2106.09685}, 2021.

\bibitem[Hu et~al.(2024)Hu, Zhao, Xu, Sun, Lou, Lin, Luo, Rajmohan, and Zhang]{hu2024agentgen}
M.~Hu, P.~Zhao, C.~Xu, Q.~Sun, J.~Lou, Q.~Lin, P.~Luo, S.~Rajmohan, and D.~Zhang.
\newblock Agentgen: Enhancing planning abilities for large language model based agent via environment and task generation.
\newblock \emph{arXiv preprint arXiv:2408.00764}, 2024.

\bibitem[Huang et~al.(2022)Huang, Abbeel, Pathak, and Mordatch]{huang2022language}
W.~Huang, P.~Abbeel, D.~Pathak, and I.~Mordatch.
\newblock Language models as zero-shot planners: Extracting actionable knowledge for embodied agents.
\newblock In \emph{International conference on machine learning}, pages 9118--9147. PMLR, 2022.

\bibitem[Jimenez et~al.(2023)Jimenez, Yang, Wettig, Yao, Pei, Press, and Narasimhan]{jimenez2023swe}
C.~E. Jimenez, J.~Yang, A.~Wettig, S.~Yao, K.~Pei, O.~Press, and K.~Narasimhan.
\newblock Swe-bench: Can language models resolve real-world github issues?
\newblock \emph{arXiv preprint arXiv:2310.06770}, 2023.

\bibitem[Keipour(2022)]{keipour2022physical}
A.~Keipour.
\newblock Physical interaction and manipulation of the environment using aerial robots.
\newblock \emph{arXiv preprint arXiv:2207.02856}, 2022.

\bibitem[Kocsis and Szepesv{\'a}ri(2006)]{kocsis2006bandit}
L.~Kocsis and C.~Szepesv{\'a}ri.
\newblock Bandit based monte-carlo planning.
\newblock In \emph{European conference on machine learning}, pages 282--293. Springer, 2006.

\bibitem[Koh et~al.(2024)Koh, Lo, Jang, Duvvur, Lim, Huang, Neubig, Zhou, Salakhutdinov, and Fried]{koh2024visualwebarena}
J.~Y. Koh, R.~Lo, L.~Jang, V.~Duvvur, M.~C. Lim, P.-Y. Huang, G.~Neubig, S.~Zhou, R.~Salakhutdinov, and D.~Fried.
\newblock Visualwebarena: Evaluating multimodal agents on realistic visual web tasks.
\newblock \emph{arXiv e-prints}, pages arXiv--2401, 2024.

\bibitem[Li et~al.(2020)Li, He, Zhou, Zhang, and Baldridge]{li2020mapping}
Y.~Li, J.~He, X.~Zhou, Y.~Zhang, and J.~Baldridge.
\newblock Mapping natural language instructions to mobile ui action sequences.
\newblock \emph{arXiv preprint arXiv:2005.03776}, 2020.

\bibitem[Liu et~al.(2023)Liu, Shi, He, Ye, Fabbri, Liu, Radev, and Cohan]{liu2023learning}
Y.~Liu, K.~Shi, K.~S. He, L.~Ye, A.~R. Fabbri, P.~Liu, D.~Radev, and A.~Cohan.
\newblock On learning to summarize with large language models as references.
\newblock \emph{arXiv preprint arXiv:2305.14239}, 2023.

\bibitem[Madaan et~al.(2024)Madaan, Tandon, Gupta, Hallinan, Gao, Wiegreffe, Alon, Dziri, Prabhumoye, Yang, et~al.]{madaan2024self}
A.~Madaan, N.~Tandon, P.~Gupta, S.~Hallinan, L.~Gao, S.~Wiegreffe, U.~Alon, N.~Dziri, S.~Prabhumoye, Y.~Yang, et~al.
\newblock Self-refine: Iterative refinement with self-feedback.
\newblock \emph{Advances in Neural Information Processing Systems}, 36, 2024.

\bibitem[Nachum et~al.(2018)Nachum, Gu, Lee, and Levine]{nachum2018data}
O.~Nachum, S.~S. Gu, H.~Lee, and S.~Levine.
\newblock Data-efficient hierarchical reinforcement learning.
\newblock \emph{Advances in neural information processing systems}, 31, 2018.

\bibitem[Pu et~al.(2023)Pu, Gao, and Wan]{pu2023summarization}
X.~Pu, M.~Gao, and X.~Wan.
\newblock Summarization is (almost) dead.
\newblock \emph{arXiv preprint arXiv:2309.09558}, 2023.

\bibitem[Reid et~al.(2024)Reid, Savinov, Teplyashin, Lepikhin, Lillicrap, Alayrac, Soricut, Lazaridou, Firat, Schrittwieser, et~al.]{reid2024gemini}
M.~Reid, N.~Savinov, D.~Teplyashin, D.~Lepikhin, T.~Lillicrap, J.-b. Alayrac, R.~Soricut, A.~Lazaridou, O.~Firat, J.~Schrittwieser, et~al.
\newblock Gemini 1.5: Unlocking multimodal understanding across millions of tokens of context.
\newblock \emph{arXiv preprint arXiv:2403.05530}, 2024.

\bibitem[Schwarzer et~al.(2020)Schwarzer, Anand, Goel, Hjelm, Courville, and Bachman]{schwarzer2020data}
M.~Schwarzer, A.~Anand, R.~Goel, R.~D. Hjelm, A.~Courville, and P.~Bachman.
\newblock Data-efficient reinforcement learning with self-predictive representations.
\newblock \emph{arXiv preprint arXiv:2007.05929}, 2020.

\bibitem[Schwarzer et~al.(2021)Schwarzer, Rajkumar, Noukhovitch, Anand, Charlin, Hjelm, Bachman, and Courville]{schwarzer2021pretraining}
M.~Schwarzer, N.~Rajkumar, M.~Noukhovitch, A.~Anand, L.~Charlin, R.~D. Hjelm, P.~Bachman, and A.~C. Courville.
\newblock Pretraining representations for data-efficient reinforcement learning.
\newblock \emph{Advances in Neural Information Processing Systems}, 34:\penalty0 12686--12699, 2021.

\bibitem[Shinn et~al.(2024)Shinn, Cassano, Gopinath, Narasimhan, and Yao]{shinn2024reflexion}
N.~Shinn, F.~Cassano, A.~Gopinath, K.~Narasimhan, and S.~Yao.
\newblock Reflexion: Language agents with verbal reinforcement learning.
\newblock \emph{Advances in Neural Information Processing Systems}, 36, 2024.

\bibitem[Su et~al.(2022)Su, Kasai, Wu, Shi, Wang, Xin, Zhang, Ostendorf, Zettlemoyer, Smith, et~al.]{su2022selective}
H.~Su, J.~Kasai, C.~H. Wu, W.~Shi, T.~Wang, J.~Xin, R.~Zhang, M.~Ostendorf, L.~Zettlemoyer, N.~A. Smith, et~al.
\newblock Selective annotation makes language models better few-shot learners.
\newblock \emph{arXiv preprint arXiv:2209.01975}, 2022.

\bibitem[Team(2024{\natexlab{a}})]{team2024codegemma}
C.~Team.
\newblock Codegemma: Open code models based on gemma.
\newblock \emph{arXiv preprint arXiv:2406.11409}, 2024{\natexlab{a}}.

\bibitem[Team(2024{\natexlab{b}})]{codestralblog}
T.~M.~A. Team.
\newblock Codestral: Hello, world!, 2024{\natexlab{b}}.
\newblock URL \url{https://mistral.ai/news/codestral/}.

\bibitem[Thomas and Brunskill(2016)]{thomas2016data}
P.~Thomas and E.~Brunskill.
\newblock Data-efficient off-policy policy evaluation for reinforcement learning.
\newblock In \emph{International Conference on Machine Learning}, pages 2139--2148. PMLR, 2016.

\bibitem[Wang et~al.(2023{\natexlab{a}})Wang, Xie, Jiang, Mandlekar, Xiao, Zhu, Fan, and Anandkumar]{wang2023voyager}
G.~Wang, Y.~Xie, Y.~Jiang, A.~Mandlekar, C.~Xiao, Y.~Zhu, L.~Fan, and A.~Anandkumar.
\newblock Voyager: An open-ended embodied agent with large language models.
\newblock \emph{arXiv preprint arXiv:2305.16291}, 2023{\natexlab{a}}.

\bibitem[Wang et~al.(2022{\natexlab{a}})Wang, Jansen, C{\^o}t{\'e}, and Ammanabrolu]{wang2022scienceworld}
R.~Wang, P.~Jansen, M.-A. C{\^o}t{\'e}, and P.~Ammanabrolu.
\newblock Scienceworld: Is your agent smarter than a 5th grader?
\newblock \emph{arXiv e-prints}, pages arXiv--2203, 2022{\natexlab{a}}.

\bibitem[Wang et~al.(2024{\natexlab{a}})Wang, Chen, Yuan, Zhang, Li, Peng, and Ji]{wang2024executable}
X.~Wang, Y.~Chen, L.~Yuan, Y.~Zhang, Y.~Li, H.~Peng, and H.~Ji.
\newblock Executable code actions elicit better llm agents.
\newblock \emph{arXiv preprint arXiv:2402.01030}, 2024{\natexlab{a}}.

\bibitem[Wang et~al.(2024{\natexlab{b}})Wang, Li, Song, Xu, Tang, Zhuge, Pan, Song, Li, Singh, Tran, Li, Ma, Zheng, Qian, Shao, Muennighoff, Zhang, Hui, Lin, Brennan, Peng, Ji, and Neubig]{openhands}
X.~Wang, B.~Li, Y.~Song, F.~F. Xu, X.~Tang, M.~Zhuge, J.~Pan, Y.~Song, B.~Li, J.~Singh, H.~H. Tran, F.~Li, R.~Ma, M.~Zheng, B.~Qian, Y.~Shao, N.~Muennighoff, Y.~Zhang, B.~Hui, J.~Lin, R.~Brennan, H.~Peng, H.~Ji, and G.~Neubig.
\newblock {OpenHands: An Open Platform for AI Software Developers as Generalist Agents}, 2024{\natexlab{b}}.
\newblock URL \url{https://arxiv.org/abs/2407.16741}.

\bibitem[Wang et~al.(2022{\natexlab{b}})Wang, Kordi, Mishra, Liu, Smith, Khashabi, and Hajishirzi]{wang2022self}
Y.~Wang, Y.~Kordi, S.~Mishra, A.~Liu, N.~A. Smith, D.~Khashabi, and H.~Hajishirzi.
\newblock Self-instruct: Aligning language models with self-generated instructions.
\newblock \emph{arXiv preprint arXiv:2212.10560}, 2022{\natexlab{b}}.

\bibitem[Wang et~al.(2023{\natexlab{b}})Wang, Cai, Chen, Liu, Ma, and Liang]{wang2023describe}
Z.~Wang, S.~Cai, G.~Chen, A.~Liu, X.~Ma, and Y.~Liang.
\newblock Describe, explain, plan and select: Interactive planning with large language models enables open-world multi-task agents.
\newblock \emph{arXiv preprint arXiv:2302.01560}, 2023{\natexlab{b}}.

\bibitem[Xie et~al.(2023)Xie, Zhou, Cheng, Shi, Weng, Liu, Hua, Zhao, Liu, Liu, et~al.]{xie2023openagents}
T.~Xie, F.~Zhou, Z.~Cheng, P.~Shi, L.~Weng, Y.~Liu, T.~J. Hua, J.~Zhao, Q.~Liu, C.~Liu, et~al.
\newblock Openagents: An open platform for language agents in the wild.
\newblock \emph{arXiv preprint arXiv:2310.10634}, 2023.

\bibitem[Xie et~al.(2024)Xie, Zhang, Chen, Li, Zhao, Cao, Hua, Cheng, Shin, Lei, et~al.]{xie2024osworld}
T.~Xie, D.~Zhang, J.~Chen, X.~Li, S.~Zhao, R.~Cao, T.~J. Hua, Z.~Cheng, D.~Shin, F.~Lei, et~al.
\newblock Osworld: Benchmarking multimodal agents for open-ended tasks in real computer environments.
\newblock \emph{arXiv preprint arXiv:2404.07972}, 2024.

\bibitem[Yang et~al.(2024)Yang, Jimenez, Wettig, Lieret, Yao, Narasimhan, and Press]{yang2024swe}
J.~Yang, C.~E. Jimenez, A.~Wettig, K.~Lieret, S.~Yao, K.~Narasimhan, and O.~Press.
\newblock Swe-agent: Agent-computer interfaces enable automated software engineering.
\newblock \emph{arXiv preprint arXiv:2405.15793}, 2024.

\bibitem[Yang et~al.(2023)Yang, Liu, Han, Chen, Huang, Fu, and Yu]{yang2023appagent}
Z.~Yang, J.~Liu, Y.~Han, X.~Chen, Z.~Huang, B.~Fu, and G.~Yu.
\newblock Appagent: Multimodal agents as smartphone users.
\newblock \emph{arXiv preprint arXiv:2312.13771}, 2023.

\bibitem[Yao et~al.(2022{\natexlab{a}})Yao, Chen, Yang, and Narasimhan]{yao2022webshop}
S.~Yao, H.~Chen, J.~Yang, and K.~Narasimhan.
\newblock Webshop: Towards scalable real-world web interaction with grounded language agents.
\newblock \emph{Advances in Neural Information Processing Systems}, 35:\penalty0 20744--20757, 2022{\natexlab{a}}.

\bibitem[Yao et~al.(2022{\natexlab{b}})Yao, Zhao, Yu, Du, Shafran, Narasimhan, and Cao]{yao2022react}
S.~Yao, J.~Zhao, D.~Yu, N.~Du, I.~Shafran, K.~Narasimhan, and Y.~Cao.
\newblock React: Synergizing reasoning and acting in language models.
\newblock \emph{arXiv preprint arXiv:2210.03629}, 2022{\natexlab{b}}.

\bibitem[Yin et~al.(2023)Yin, Brahman, Ravichander, Chandu, Chang, Choi, and Lin]{yin2023lumos}
D.~Yin, F.~Brahman, A.~Ravichander, K.~Chandu, K.-W. Chang, Y.~Choi, and B.~Y. Lin.
\newblock Lumos: Learning agents with unified data, modular design, and open-source llms.
\newblock \emph{arXiv preprint arXiv:2311.05657}, 2023.

\bibitem[Zeng et~al.(2023)Zeng, Liu, Lu, Wang, Liu, Dong, and Tang]{zeng2023agenttuning}
A.~Zeng, M.~Liu, R.~Lu, B.~Wang, X.~Liu, Y.~Dong, and J.~Tang.
\newblock Agenttuning: Enabling generalized agent abilities for llms.
\newblock \emph{arXiv preprint arXiv:2310.12823}, 2023.

\bibitem[Zhan and Zhang(2023)]{zhan2023you}
Z.~Zhan and A.~Zhang.
\newblock You only look at screens: Multimodal chain-of-action agents.
\newblock \emph{arXiv preprint arXiv:2309.11436}, 2023.

\bibitem[Zhang et~al.(2024)Zhang, Yu, Liao, Li, Wu, and Wei]{zhang2024ui}
J.~Zhang, Y.~Yu, M.~Liao, W.~Li, J.~Wu, and Z.~Wei.
\newblock Ui-hawk: Unleashing the screen stream understanding for gui agents.
\newblock \emph{arXiv preprint}, 2024.

\bibitem[Zhao et~al.(2024)Zhao, Ma, Chai, Wang, Chen, Guo, Zhang, Wang, and Wang]{zhao2024we}
Z.~Zhao, K.~Ma, W.~Chai, X.~Wang, K.~Chen, D.~Guo, Y.~Zhang, H.~Wang, and G.~Wang.
\newblock Do we really need a complex agent system? distill embodied agent into a single model.
\newblock \emph{arXiv preprint arXiv:2404.04619}, 2024.

\bibitem[Zhou et~al.(2023{\natexlab{a}})Zhou, Yan, Shlapentokh-Rothman, Wang, and Wang]{zhou2023language}
A.~Zhou, K.~Yan, M.~Shlapentokh-Rothman, H.~Wang, and Y.-X. Wang.
\newblock Language agent tree search unifies reasoning acting and planning in language models.
\newblock \emph{arXiv preprint arXiv:2310.04406}, 2023{\natexlab{a}}.

\bibitem[Zhou et~al.(2023{\natexlab{b}})Zhou, Xu, Zhu, Zhou, Lo, Sridhar, Cheng, Ou, Bisk, Fried, et~al.]{zhou2023webarena}
S.~Zhou, F.~F. Xu, H.~Zhu, X.~Zhou, R.~Lo, A.~Sridhar, X.~Cheng, T.~Ou, Y.~Bisk, D.~Fried, et~al.
\newblock Webarena: A realistic web environment for building autonomous agents.
\newblock \emph{arXiv preprint arXiv:2307.13854}, 2023{\natexlab{b}}.

\end{thebibliography}
